%% file: main.tex
\documentclass{article} 
\usepackage{iclr2021_conference,times}

\usepackage{amsmath,amsfonts,bm,amssymb}

\usepackage{xcolor}
\usepackage{color}
\usepackage{graphicx}

\usepackage{algorithm}
\usepackage{algorithmic}

\usepackage{caption}
\usepackage{subcaption}

\usepackage{booktabs}

\usepackage{tcolorbox}

\usepackage{wrapfig}
\usepackage{textcomp}

\usepackage{tikz}
\usepackage{tikz-3dplot}

\usepackage{inputenc}  

\UseRawInputEncoding 
\usetikzlibrary{positioning}
\usetikzlibrary{plotmarks}
\usetikzlibrary{arrows.meta}

\newcommand{\diag}{\mathsf{diag}}

\newcommand{\dft}{\mathrm{DFT}}
\newcommand{\idft}{\mathrm{IDFT}}

\renewcommand{\vec}{\mathsf{vec}}
\renewcommand{\mod}{~\mathsf{mod}~}
\newcommand{\trans}{\mathsf{trans}}
\newcommand{\circm}{\mathsf{circ}}

\newcommand{\Appendix}[1]{the full version for}

\newtheorem{theorem}{Theorem}[section]

\newtheorem{proposition}[theorem]{Proposition}

\newtheorem{remark}{Remark}

\newtheorem{fact}{Fact}

\newcommand{\p}{\bm{p}}

\renewcommand{\u}{\bm{u}}
\newcommand{\bv}{\bm{v}}

\newcommand{\x}{\bm{x}}
\newcommand{\y}{\bm{y}}
\newcommand{\z}{\bm{z}}
\newcommand{\q}{\bm{q}}

\newcommand{\C}{\bm{C}}
\newcommand{\D}{\bm{D}}
\newcommand{\E}{\bm{E}}
\newcommand{\F}{\bm{F}}

\newcommand{\I}{\bm{I}}

\renewcommand{\P}{\mathbf{P}}

\newcommand{\R}{\mathbb{R}}
\renewcommand{\Re}{\mathbb{R}}
\newcommand{\Co}{\mathbb{C}}

\newcommand{\U}{\bm{U}}
\newcommand{\V}{\bm{V}}
\newcommand{\W}{\bm{W}}
\newcommand{\X}{\bm{X}}

\newcommand{\Z}{\bm{Z}}

\newcommand{\0}{\mathbf{0}}

\newcommand{\cC}{\mathcal{C}}

\newcommand{\cE}{\mathcal{E}}

\DeclareMathOperator*{\argmin}{argmin}

\newenvironment{proof}{\paragraph{Proof:}}{\hfill$\square$}

\usepackage{hyperref}
\usepackage{url}
\usepackage{hyperref}
\hypersetup{
    colorlinks,
    linkcolor={blue!80!black},
    citecolor={blue!80!black},    
}
\colorlet{linkequation}{blue}

\newcommand\blfootnote[1]{%
	\begingroup
	\renewcommand\thefootnote{}\footnote{#1}%
	\addtocounter{footnote}{-1}%
	\endgroup
}

\newcommand{\todo}[1]{\textcolor{red}{[TODO]}}

\title{\Large Deep Networks from the Principle of Rate Reduction}

\author{
Kwan Ho Ryan Chan$^\dagger$\,
Yaodong Yu$^\dagger$\,
Chong You$^\dagger$\,
Haozhi Qi$^\dagger$\,
John Wright$^{\ddagger \diamond}$\,
Yi Ma$^\dagger$
\\
\vspace*{-0.05in} 
\\
$^\dagger$Department of EECS, University of California, Berkeley\\
$^\ddagger$Department of Electrical Engineering and Data Science Institute, Columbia University\\
$^\diamond$Department of Applied Physics and Applied Mathematics, Columbia University\\
}

\preprint
\begin{document}

\maketitle
\blfootnote{\hspace*{-1.4mm}$^*$The first three authors contributed equally to this work.}

\begin{abstract}
This work attempts to interpret modern deep (convolutional) networks from the principles of rate reduction and (shift) invariant classification.
We show that the basic iterative gradient ascent scheme for optimizing the rate reduction of learned features naturally leads to a multi-layer deep network, one iteration per layer. The layered architectures, linear and nonlinear operators, and even parameters of the network are all explicitly constructed layer-by-layer in a forward propagation fashion by emulating the gradient scheme. All components of this ``white box'' network have precise optimization, statistical, and geometric interpretation. This principled framework also reveals and justifies the role of multi-channel lifting and sparse coding in early stage of deep networks. Moreover, all linear operators of the so-derived network naturally become multi-channel convolutions when we enforce classification to be rigorously shift-invariant. The derivation also indicates that such a convolutional network is significantly more efficient to construct and learn in the spectral domain. Our preliminary simulations and experiments indicate that so constructed deep network can already learn a good discriminative representation even without any back propagation training.
\end{abstract}

\input{intro}
\input{approch}

\input{experiments}

\section{Conclusions and Future Work}
Following the recently proposed maximal coding rate reduction framework of  \cite{yu2020learning}, this work offers a principled interpretation of modern deep (convolutional) networks by construction from first principles and with minimal assumptions. As we see, the rate reduction principle provides a rigorous explanation for the deep architecture and components from the perspective of optimizing the rate reduction of final representations. In particular, most key characteristics of modern deep neural networks, including their layered architectures, linear (convolutional) operators (for shift invariance), and nonlinear operators (for classifying and sparsifying), can all be derived as necessary operations for optimizing this objective. Somewhat unexpected, our analysis shows that the nominal architecture and parameters of such a deep (convolution) network can all be explicitly constructed layer-by-layer in a forward propagation fashion without the need of back propagation training. The analysis further reveals that such a deep convolutional network is computationally more efficient to construct and learn in the spectral domain. Preliminary simulations and experiments on basic data sets clearly verify the so-constructed ReduNet achieves the desired functionality and objective. 

Although in this work the ReduNet is purely forward constructed as the nominal optimization path for rate reduction, one may study how to effectively fine tune it via back propagation. In this work the rate distortion function primarily treats each class as low-dimensional subspace or (degenerate) Gaussian. As already pointed out by \cite{wright2008classification}, if the data have more sophisticated nonlinear structures, to better classify such data, the rate distortion function $R(\Z)$ can readily incorporate an arbitrary kernel function $k(\z, \z')$ by replacing the inner product in $\log\det \Big(\I + {\alpha} \Z \Z^{*} \Big) = \log\det \Big(\I + {\alpha} \Z^{*}\Z \Big).$ Hence this framework is naturally amenable to further analysis and development with kernel functions. Furthermore, the lossy compression and rate distortion framework was originally developed for seeking optimal data clustering $\bm \Pi$ via minimizing the second term $R_c$ of $\Delta R$ \citep{ma2007segmentation,GPCA}. Hence we believe the maximal rate reduction framework can be  naturally extended to settings of {online}, self-supervised, or {\em unsupervised} learning if the class information $\bm \Pi$ is partially or entirely unknown and is to be optimized jointly with the representation $\bm Z$.

All in all, we strongly believe that maximizing rate reduction provides a principled framework for designing new networks with interpretable architectures and operators that could scale up to real-world datasets and problems, with better performance guarantees. 

\section*{Acknowledgements}
Yi acknowledges support from ONR grant N00014-20-1-2002 and the joint Simons Foundation-NSF DMS grant \#2031899, as well as support from Berkeley FHL Vive Center for Enhanced Reality and Berkeley Center for Augmented Cognition. Chong and Yi acknowledge support from Tsinghua-Berkeley Shenzhen Institute (TBSI) Research Fund. Yaodong, Haozhi, and Yi acknowledge support from Berkeley AI Research (BAIR). John acknowledges support from NSF grants 1838061, 1740833, and 1733857. 

\bibliography{reference,iclr2021_conference}
\bibliographystyle{iclr2021_conference}

\newpage
\appendix

\input{appendix}
\input{appendix-experiment}

\end{document}

%% file: intro.tex
\section{Introduction and Motivation}
In recent years, various deep (convolution) network architectures such as AlexNet \citep{krizhevsky2012imagenet}, VGG \citep{simonyan2014very}, ResNet \citep{he2016deep}, DenseNet \citep{dense-net},  Recurrent CNN, LSTM \citep{LSTM}, Capsule Networks  \citep{Hinton2011TransformingA}, etc.,  have demonstrated very good performance in classification tasks of real-world datasets such as speeches or images. Nevertheless, almost all such networks are developed through years of empirical {\em trial and error}, including both their architectures/operators and the ways they are to be effectively trained. Some recent practices even take to the extreme by searching for effective network structures and training strategies through extensive random search techniques, such as Neural Architecture Search \citep{NAS-1,Baker2017DesigningNN}, AutoML \citep{automl}, and Learning to Learn \citep{andrychowicz2016learning}. 

Despite tremendous empirical advances, there is still a lack of rigorous theoretical justification of the need or reasons for ``deep'' network architectures and a lack of fundamental understanding of the associated operators (e.g. multi-channel convolution and nonlinear activation) in each layer. As a result, deep networks are often designed and trained heuristically and then used as a ``black box.'' There have been a severe lack of guiding principles for each of the stages: For a given task, how wide or deep the network should be? What are the roles and relationships among the multiple (convolution) channels? Which parts of the networks need to be learned and trained and which can be determined in advance? How to evaluate the optimality of the resulting network? As a consequence, besides empirical evaluation, it is usually impossible to offer any rigorous guarantees for certain performance of a trained network, such as invariance to transformation \citep{azulay2018deep,engstrom2017rotation} or overfitting noisy or even arbitrary labels \citep{ZhangBeHaReVi17}. 

In this paper, we do not intend to address all these questions but we would attempt to offer a plausible interpretation of deep (convolution) neural networks by deriving a class of deep networks {\em from first principles}. We contend that all key features and structures of modern deep (convolution) neural networks can be naturally derived from optimizing a principled objective, namely the {\em rate reduction} recently proposed by \cite{yu2020learning}, that seeks a compact discriminative (invariant) representation of the data. More specifically, the basic iterative {\em gradient ascent} scheme for optimizing the  objective naturally takes the form of a deep neural network, one layer per iteration.

This principled approach brings a couple of nice surprises: First, architectures, operators, and parameters of the network can be constructed explicitly layer-by-layer in a {\em forward propagation} fashion, and all inherit precise optimization, statistical and geometric interpretation. As result, the so constructed ``white box'' deep network already gives a good discriminative representation (and achieves good classification performance) {\em without  any back propagation} for training the deep network. Second, in the case of seeking a representation {\em rigorously} {invariant to shift or  translation}, the network naturally lends itself to a multi-channel convolutional network. Moreover, the derivation indicates such a convolutional network is computationally more efficient to learn and construct in the {\em spectral (Fourier) domain}, analogous to how neurons in the visual cortex encode and transit information with their spiking frequencies \citep{spiking-neuron-book,Belitski5696}.

%% file: approch.tex
\section{Technical Approach}
Consider a basic classification task: given a set of $m$ samples $\X \doteq [\x^1,\ldots, \x^m] \in \R^{n\times m}$ and their associated memberships $\bm \pi(\x^i) \in [k]$ in $k$ different classes, a deep network is typically used to model a direct mapping from the input data $\x \in\Re^n$ to its class label $f(\x, \bm \theta): \x \mapsto \y \in \Re^k$, where $\y$ is typically a ``one-hot'' vector encoding the membership information $\bm \pi(\x)$: the $j$-th entry of $\y$ is 1 iff $\bm \pi(\x) = j$. The parameters $\bm \theta$ of the network is typically learned to minimize certain prediction loss, say the cross entropy loss, via gradient-descent type back propagation. Although this popular approach provides people a direct and effective way to train a network that predicts the class information, the so learned representation is however implicit and lacks clear interpretation. 

\subsection{Principle of Rate Reduction and Group Invariance}
\paragraph{The Principle of Maximal Coding Rate Reduction.}
To help better understand features learned in a deep network, the recent work of \cite{yu2020learning} has argued that the goal of (deep) learning is to learn a compact discriminative and diverse feature representation\footnote{To simplify the presentation, we assume for now that the feature $\z$ and $\x$ have the same dimension $n$. But in general they can be different as we will soon see, say in the case $\z$ is multi-channel extracted from $\x$.} $\z = f(\x) \in \Re^n$ of the data $\x$ before any subsequent tasks such as classification:
$
\x  \xrightarrow{\hspace{0.5mm} f(\x)\hspace{0.5mm}} \z \xrightarrow{\hspace{0.5mm} h(\z) \hspace{0.5mm}} \y.
$
To be more precise, instead of directly fitting the class label $\y$, a principled objective is to learn a feature map $f(\x): \x \mapsto \z$ which transforms the data $\x$ onto a set of most discriminative low-dimensional linear subspaces $\{\mathcal S^j \}_{j=1}^k \subset \Re^n$, one subspace $\mathcal S^j$ per class $j \in [k]$. 

Let $\Z \doteq [\z^1,\ldots, \z^m] = [f(\x^1), \ldots, f(\x^m)]$ be the features of the given samples $\X$. WLOG, we may assume all features $\z^i$ are normalized to be of unit norm: $\z^i \in \mathbb{S}^{n-1}$. For convenience, let $\bm \Pi^j \in \Re^{m\times m}$ be a diagonal matrix whose diagonal entries encode the membership of samples/features belong to the $j$-th class: $\bm \Pi^j(i,i) = \bm \pi(\x^i) = \bm \pi(\z^i)$. Then based on principles from lossy data compression \citep{ma2007segmentation}, \cite{yu2020learning} has suggested that the optimal representation $\Z_\star \subset \mathbb{S}^{n-1}$ should maximize the following coding rate reduction objective, known as the MCR$^2$ principle:
\begin{equation}\label{eq:mcr2-formulation}
\mbox{\em Rate Reduction:} \quad \Delta R(\Z ) \doteq \underbrace{\frac{1}{2}\log\det \Big(\I + {\alpha} \Z \Z^{*} \Big)}_{R(\Z)} - \underbrace{\sum_{j=1}^{k}\frac{\gamma_j}{2} \log\det\Big(\I + {\alpha_j} \Z \bm{\Pi}^{j} \Z^{*} \Big)}_{R_c(\Z,\bm \Pi)},
\end{equation}
where $\alpha=n/(m\epsilon^2)$, $\alpha_j=n/(\textsf{tr}(\bm{\Pi}^{j})\epsilon^2)$, $\gamma_j=\textsf{tr}(\bm{\Pi}^{j})/m$ for $j = 1,\ldots, k$.
Given a prescribed quantization error $\epsilon$, the first term $R$ of $\Delta R(\Z )$ measures the total coding length for all the features $\Z$ and the second term $R_c$ is the sum of coding lengths for features in each of the $k$ classes.

In \cite{yu2020learning}, the authors have shown the optimal representation $\Z_\star$ that maximizes the above objective indeed has desirable nice properties. Nevertheless, they adopted a conventional deep network (e.g. ResNet) as a black box to model and parameterize the feature mapping: $\z = f(\x, \bm \theta)$. It has empirically shown that with such a choice, one can effectively optimize the MCR$^2$ objective and obtain discriminative and diverse representations for classifying real image data. 

However, there remain several unanswered problems. Although the resulting feature representation is more interpretable, the network itself is still not. It is {\em not} clear why any chosen network is able to optimize the desired MCR$^2$ objective: Would there be any potential limitations? The good empirical results (say with a ResNet) do not necessarily justify the particular choice in architectures and operators of the network: Why is a layered model necessary, how wide and deep is adequate, and is there any rigorous justification for the use of convolutions and nonlinear operators used? In Section \ref{sec:vector}, we show that using gradient ascent to maximize the rate reduction $\Delta R(\Z )$ naturally leads to a ``white box'' deep network that represents such a mapping. All linear/nonlinear operators and parameters of the network are {\em explicitly constructed in a purely forward propagation fashion}. 

\paragraph{Group Invariant Rate Reduction.}
So far, we have considered the data and features as vectors. In many applications, such as serial data or imagery data, the semantic meaning (labels) of the data and their features are  {\em invariant} to certain  transformations $\mathfrak{g} \in \mathbb{G}$ (for some group $\mathbb{G}$) \citep{CohenW16}. For example, the meaning of an audio signal is invariant to shift in time; and the identity of an object in an image is invariant to translation in the image plane. Hence, we prefer the feature mapping $f(\x,\bm \theta)$ is rigorously invariant to such transformations:
\begin{equation}
 \mbox{\em Group Invariance:}\quad    f(\x\circ \mathfrak{g}, \bm \theta) \sim f(\x,\bm \theta), \quad \forall \mathfrak{g} \in \mathbb{G},
\end{equation}
where ``$\sim$'' indicates two features belonging to the same equivalent class. The recent work of \cite{deep-sets-NIPS2017,Maron2020OnLS} characterize properties of networks and operators for set permutation groups. Nevertheless, it remains challenging to learn features via a deep network that are {\em guaranteed} to be invariant even to simple transformations such as translation and rotation  \citep{azulay2018deep,engstrom2017rotation}.  In Section \ref{sec:invariance}, we show that the MCR$^2$ principle is compatible with invariance in a very natural and precise way: we only need to assign all transformed versions $\{\x\circ \mathfrak{g} \mid \mathfrak{g} \in \mathbb G\}$  into the same class as $\x$ and map them all to the same subspace $\mathcal S$.\footnote{Hence, any subsequent classifiers defined on the resulting set of subspaces will be automatically invariant to such transformations.} We will rigorously show (in the Appendices) that, when the group $\mathbb G$ is (discrete) circular 1D shifting or 2D translation, the resulting deep network naturally becomes a {\em multi-channel convolution network}!

\subsection{Deep Networks from Optimizing Rate Reduction}\label{sec:vector}

\paragraph{Gradient Ascent for Rate Reduction on the Training Samples.} First let us directly try to optimize the objective $\Delta R(\Z )$ as a function in the training samples $\Z \subset \mathbb{S}^{n-1}$. To this end, we may adopt a (projected) {\em gradient ascent} scheme, for some step size $\eta >0$: 
\begin{equation}
\bm Z_{\ell+1}   \; \propto \; \bm Z_{\ell} + \eta \cdot \frac{\partial \Delta R}{\partial \bm Z}\bigg|_{\Z_\ell}
 \quad \mbox{subject to} \quad \Z_{\ell+1} \subset \mathbb{S}^{n-1}.
\label{eqn:gradient-descent}
\end{equation}
This scheme can be interpreted as how one should incrementally adjust locations of the current features $\Z_\ell$ in order for the resulting $\Z_{\ell +1}$ to improve the rate reduction $\Delta R(\Z)$. Simple calculation shows that the gradient $\frac{\partial \Delta R}{\partial \bm Z}$ entails evaluating the following derivatives of the terms in \eqref{eq:mcr2-formulation}:
\begin{small}
\begin{eqnarray}
    \frac{1}{2}\frac{\partial \log \det (\I + \alpha \Z \Z^{*} )}{\partial \bm Z}\bigg|_{\Z_\ell} &=&  \underbrace{\alpha(\I + \alpha\Z_\ell \Z_\ell^{*})^{-1}}_{\E_{\ell} \; \in \Re^{n\times n}}\Z_\ell \quad \in \Re^{n \times m}, \label{eqn:expand-directions} \\
    \frac{1}{2}\frac{\partial \left( \gamma_j  \log \det (\I + \alpha_j \Z \bm \Pi^j \Z^{*} )  \right)}{\partial \bm Z}\bigg|_{\Z_\ell} &=& \gamma_j  \underbrace{ \alpha_j  (\I +  \alpha_j \Z_\ell \bm \Pi^j \Z_\ell^{*})^{-1}}_{\bm C_{\ell}^j \; \in \Re^{n\times n}} \Z_{\ell} \bm \Pi^j \quad \in \Re^{n \times m}.\label{eqn:compress-directions}
\end{eqnarray}
\end{small}Notice that in the above, the matrix $\bm E_{\ell}$  only depends on $\Z_{\ell}$ and it aims to {\em expand} all the features to increase the overall coding rate; the matrix $\bm C_{\ell}^{j}$ depends on features from each class and aims to {\em compress} them to reduce the coding rate of each class. 
We provide the geometric and statistic meaning of $\bm E_\ell$ and $\bm C^j_\ell$ in Remark \ref{rem:regression-interpretation} below. 
Then the complete gradient $\frac{\partial \Delta R}{\partial \bm Z}\big|_{\Z_\ell}$ is of the following form:
\begin{equation}
\frac{\partial \Delta R}{\partial \bm Z}\bigg|_{\Z_\ell}  = \underbrace{\bm E_{\ell}}_{\text{Expansion}} \Z_{\ell} - \sum_{j=1}^k \gamma_j \underbrace{\bm C^{j}_{\ell}}_{\text{Compression}}  \Z_{\ell} \bm{\Pi}^j \quad \in \Re^{n\times m}.
\label{eqn:DR-gradient}
\end{equation}

\begin{remark}[Interpretation of $\bm E_\ell$ and $\bm C^j_\ell$ as Linear Operators]\label{rem:regression-interpretation} 
For any $\z_\ell \in \mathbb{R}^n$, we have
\begin{gather}
    \bm E_\ell \z_\ell = \alpha(\z_\ell - \Z_\ell \q_\ell^*) \quad
    \mbox{where}\quad \q_\ell^* \doteq \argmin_{\q_\ell} \alpha \|\z_\ell - \Z_\ell \q_\ell\|_2^2 + \|\q_\ell\|_2^2.
\end{gather}
Notice that $\q_\ell^*$ is exactly the solution to the ridge regression by all the data points $\Z_\ell$ concerned. Therefore, $\E_\ell$ (similarly for $\bm C_\ell^j$) is approximately (i.e. when $m$ is large enough) the projection onto the orthogonal complement of the subspace spanned by columns of $\Z_\ell$.
Another way to interpret the matrix $\E_\ell$ is through eigenvalue decomposition of the covariance matrix $\Z_\ell \Z_\ell^*$. Assuming that $\Z_\ell \Z_\ell^* \doteq \U_\ell \bm \Lambda_\ell \U_\ell^*$ where $\bm \Lambda_\ell \doteq \diag\{\sigma_1, \ldots, \sigma_d \}$, we have 
\begin{equation}\E_\ell = \alpha\, \U_\ell\, \diag\left\{\frac{1}{1+\alpha\sigma_1}, \ldots, \frac{1}{1+\alpha\sigma_d}\right\} \U_\ell^*.
\end{equation}

\begin{figure}[t]
    \centering
    \includegraphics[width=0.65\linewidth]{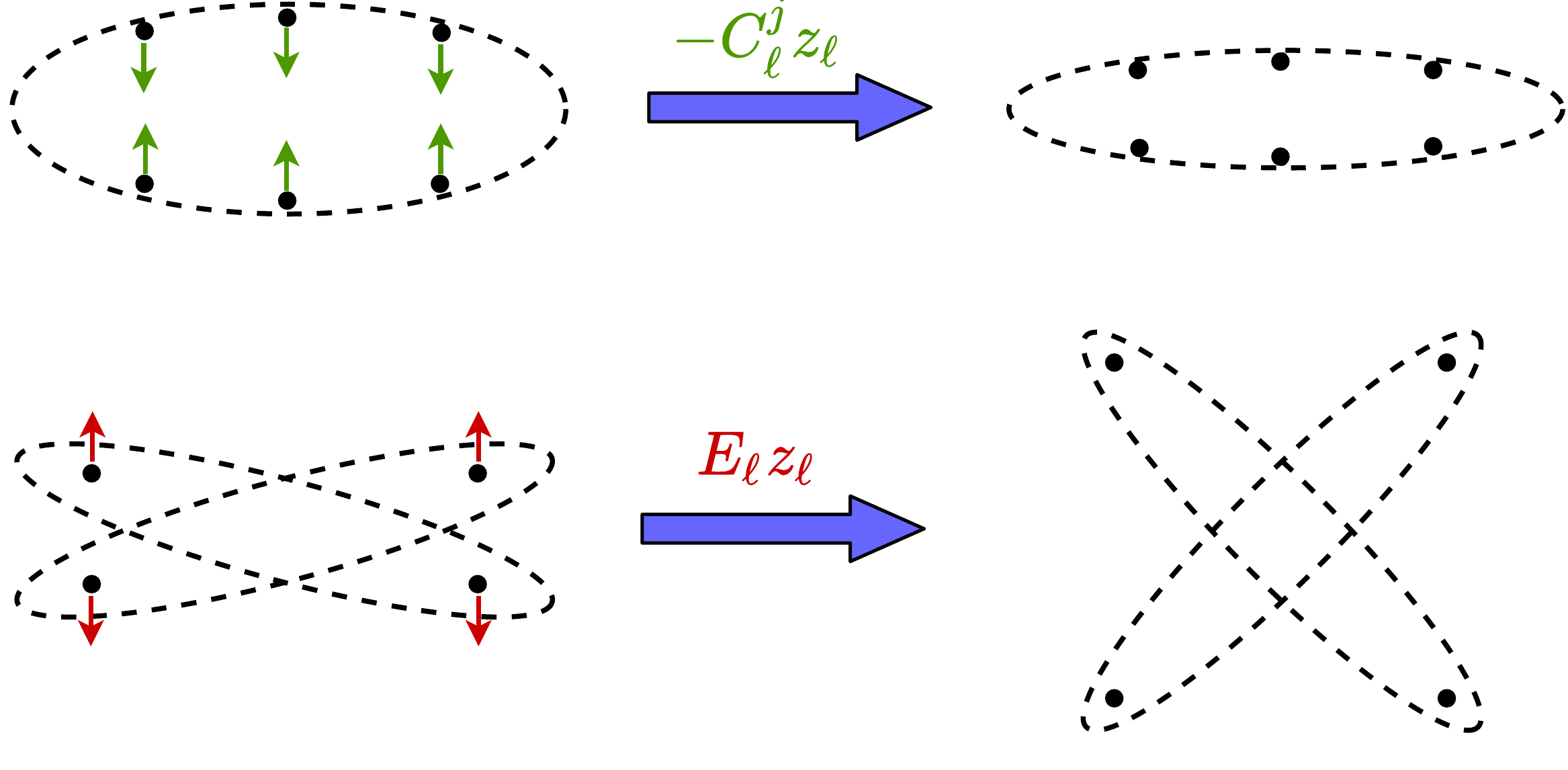}
    \caption{\small Interpretation of $\bm E_\ell$ and $\bm C_\ell^j$: $\bm E_\ell$ expands all features by contrasting and repelling features across different classes; $\bm C_\ell^j$ compresses each class by contracting the features to a low-dimensional subspace.}
    \label{fig:regression-interpretation}
    \vspace{-0.1in}
\end{figure}

Therefore, the matrix $\E_\ell$ operates on a vector $\z_\ell$ by stretching in a way that directions of large variance are shrunk while directions of vanishing variance are kept. These are exactly the directions \eqref{eqn:expand-directions} in which we move the features so that the overall volume expands and the coding rate will increase, hence the positive sign. To the opposite effect, the directions associated with \eqref{eqn:compress-directions} are ``residuals'' of features of each class deviate from the subspace to which they are supposed to belong. These are exactly the directions in which the features need to be compressed back onto their respective subspace, hence the negative sign (see Figure~\ref{fig:regression-interpretation}). 

Essentially, the linear operations $\bm E_\ell$ and $\bm C^j_\ell$ in gradient ascend for rate reduction are determined by training data conducting ``auto-regressions". The recent renewed understanding about ridge regression in an over-parameterized setting \citep{yang2020rethinking,Wu2020OnTO} indicates that using seemingly redundantly sampled data (from each subspaces) as regressors do not lead to overfitting. 

\end{remark}

\paragraph{Gradient-Guided Feature Map Increment.}
Notice that in the above, the gradient ascent considers all the features $\Z_{\ell} = [\z_{\ell}^{1}, \dots, \z_{\ell}^{m}]$ as free variables. The increment $\Z_{\ell+1} - \Z_{\ell} = \eta \frac{\partial \Delta R}{\partial \bm Z}\big|_{\Z_\ell}$ does not yet give a transform on the entire feature domain $\z_\ell \in \Re^n$. Hence, in order to find the optimal $f(\x,\bm  \theta)$ explicitly, we may consider constructing a small increment transform $g(\cdot, \bm{\theta}_{\ell})$ on the $\ell$-th layer feature $\z_\ell$ to emulate the above (projected) gradient scheme:
\begin{equation}
\z_{\ell + 1}   \; \propto \; \z_{\ell} + \eta\cdot  g(\z_{\ell}, \bm{\theta}_{\ell}) \quad \mbox{subject to} \quad \z_{\ell +1} \in \mathbb{S}^{n-1}
\label{eqn:gradient-descent-transform}
\end{equation}
such that: $\big[g(\z^1_{\ell}, \bm \theta_{\ell}), \ldots, g(\z^m_{\ell}, \bm \theta_{\ell}) \big] \approx \frac{\partial \Delta R}{\partial \bm Z}\big|_{\Z_\ell}.$ That is, we need to approximate the gradient flow $\frac{\partial \Delta R}{\partial \bm Z}$ that locally deforms each (training) feature $\{\z^i_\ell\}_{i=1}^m$ with a continuous mapping $g(\z)$ defined on the entire feature space $\z_\ell \in \Re^n$.

\begin{remark}[Connection and Difference from Neural ODE]\label{rem:ODE}
One may interpret the increment \eqref{eqn:gradient-descent-transform} as a discretized version of a continuous ordinary differential equation (ODE):
\begin{equation}
\dot{\z} = g(\z, \theta).
\end{equation}
Hence the (deep) network so constructed can be interpreted as certain neural ODE \citep{chen2018neural}. Nevertheless, unlike neural ODE where the flow $g$ is chosen to be some generic structures whose parameters are trained later, here our $g(\z, \theta)$ is to emulate the gradient flow of the rate reduction on the feature set \begin{equation*}
    \dot{\Z} = \eta \cdot \frac{\partial \Delta R}{\partial \bm Z} \quad \Longrightarrow \quad \bm Z_{\ell+1}   \; \propto \; \bm Z_{\ell} + \eta \cdot \frac{\partial \Delta R}{\partial \bm Z}\bigg|_{\Z_\ell},
\end{equation*} 
and its structure and parameters are entirely derived and fully determined from this objective, without any other priors, heuristics or post training . 
\end{remark}

By inspecting the structure of the gradient \eqref{eqn:DR-gradient}, it suggests that a natural candidate for the increment transform $g(\z_\ell, \bm \theta_\ell)$ is of the form:
\begin{equation}
    g(\z_\ell, \bm \theta_\ell) \; \doteq \; \E_\ell \z_\ell - \sum_{j=1}^k \gamma_j \bm C^{j}_{\ell}  \z_{\ell} \bm \pi^j(\z_\ell) \quad \in \Re^n,
    \label{eqn:DR-gradient-transform}
\end{equation}
where  $\bm \pi^j(\z_\ell) \in [0,1]$ {indicates the probability of $\z_{\ell}$ belonging to the $j$-th class.}\footnote{Notice that on the training samples $\Z_\ell$, for which the memberships $\bm \Pi^j$ are known,  the so defined $g(\z_\ell, \bm \theta)$ gives exactly the values for the gradient $\frac{\partial \Delta R}{\partial \bm Z}\big|_{\Z_\ell}$.} Notice that the increment depends on  1). A  linear map represented by $\bm E_{\ell}$ that depends only on statistics of all features from the preceding layer; 2). A set of linear maps $ \{ \bm C_{\ell}^{j}\}_{j=1}^{k}$ and memberships $\{ \bm \pi^j(\z_\ell)\}_{j=1}^k$ of the  features.

\begin{figure}[t]
    \centering
    \begin{subfigure}[b]{0.4\textwidth}
        \includegraphics[width=\textwidth]{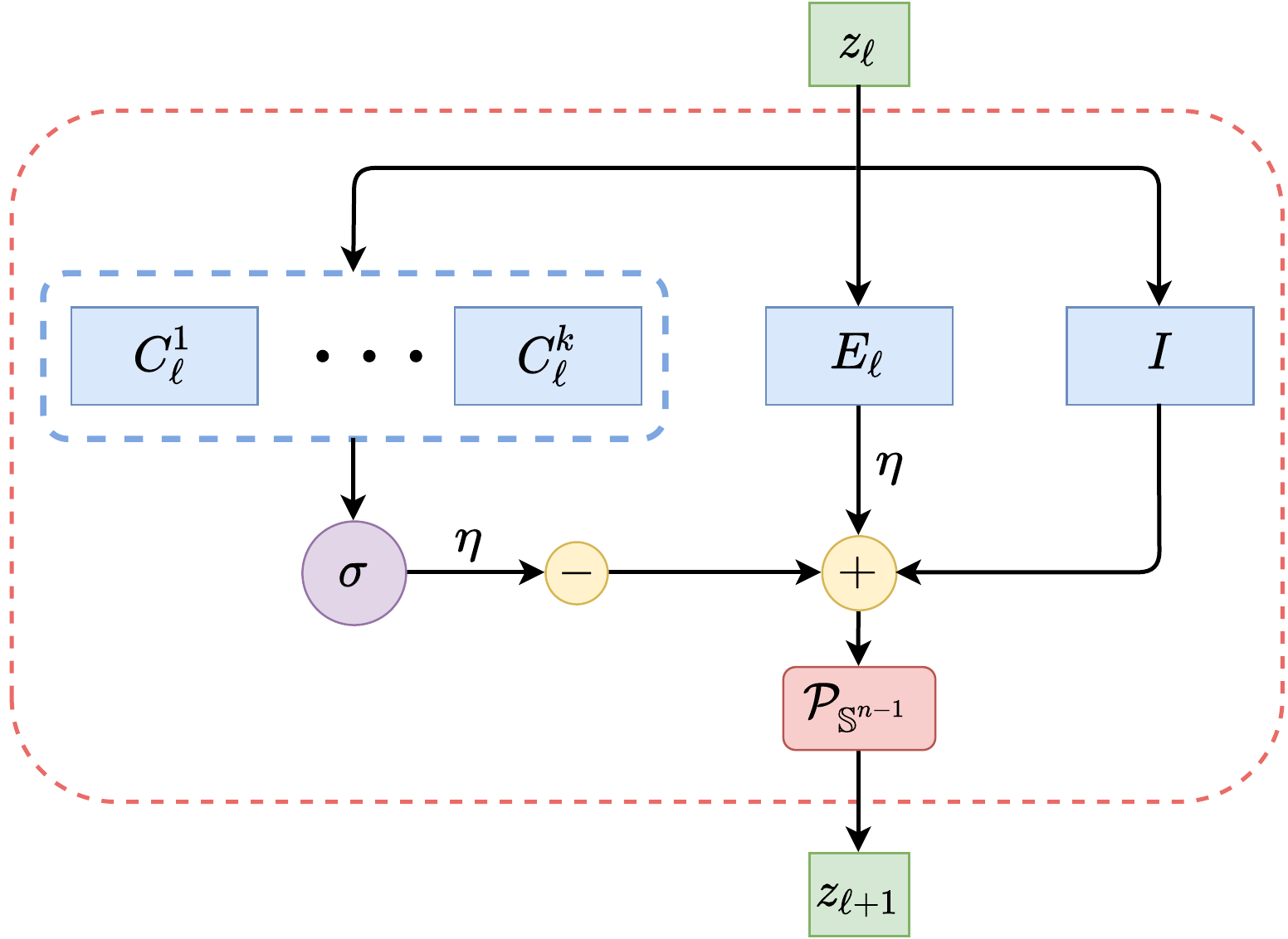}
        \caption{\textbf{ReduNet}.}
    \end{subfigure}
    \hspace{3mm}
    \begin{subfigure}[b]{0.55\textwidth}
        \includegraphics[width=\textwidth]{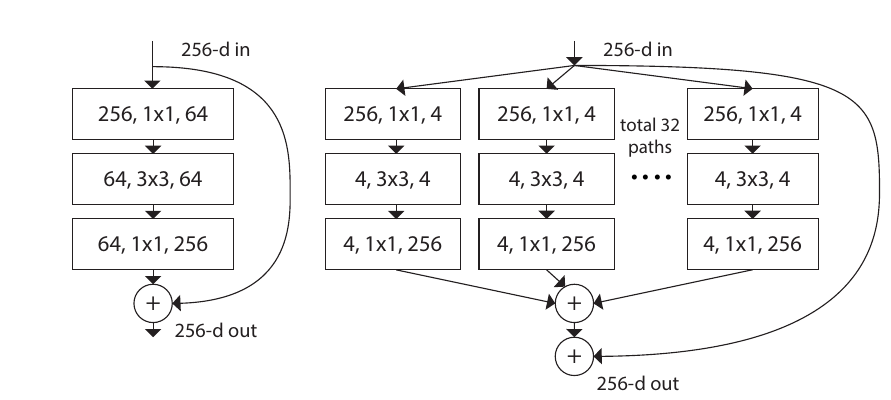}
        \caption{\textbf{ResNet} and \textbf{ResNeXt}.}
    \end{subfigure}
    \caption{\small Comparison of Network Architectures. \textbf{(a)}: Layer structure of the \textbf{ReduNet} derived from one iteration of gradient ascent for optimizing rate reduction. \textbf{(b)} (left): A layer of ResNet~\citep{he2016deep}; and \textbf{(b)} (right): A layer of ResNeXt~\citep{ResNEXT}. As we will see in the next section, the linear operators $\bm E_\ell$ and $\bm{C}^j_\ell$ of the ReduNet naturally become (multi-channel) convolutions when shift-invariance is imposed.}
    \label{fig:arch}
    \vspace{-0.1in}
\end{figure}

Since we only have the membership $\bm \pi^j$ for the training samples, the function $g$ defined in \eqref{eqn:DR-gradient-transform} can only be evaluated on the training samples. To extrapolate the function $g$ to the entire feature space, we need to estimate $\bm \pi^j(\z_\ell)$ in its second term. In the conventional deep learning, this map is typically modeled as a deep network and learned from the training data, say via {\em back propagation}. Nevertheless, our goal here is not to learn a precise classifier $\bm \pi^{j}(\z_\ell)$ already. Instead, we only need a good enough estimate of the class information in order for $g$ to approximate the gradient $\frac{\partial \Delta R}{\partial \bm Z}$ well.

From the geometric interpretation of the linear maps $\bm E_\ell$ and $\bm C^j_\ell$ given by Remark \ref{rem:regression-interpretation}, the term $\p^{j}_{\ell} \doteq \bm C_{\ell}^j \z_{\ell}$ can be viewed as projection of $\z_{\ell}$ onto the orthogonal complement of each class $j$. 
Therefore, $\|\p^{j}_{\ell}\|_2$ is small if $\z_\ell$ is in class $j$ and large otherwise. This motivates us to estimate its membership based on the following softmax function:
$
\widehat{\bm \pi}^j(\z_\ell) \doteq \frac{\exp{( -\lambda \|\bm C_{\ell}^j  \z_{\ell} \|)} }{ \sum_{j=1}^{k} \exp{(-\lambda \|\bm C_{\ell}^j  \z_{\ell}  \|)}} \in [0,1].
$
Hence the second term of \eqref{eqn:DR-gradient-transform} can be approximated by this estimated membership:\footnote{The choice of the softmax is mostly for its simplicity as it is widely used in other (forward components of) deep networks for purposes such as selection, gating \citep{MoE} and routing \citep{capsule-net}. In principle, this term can be approximated by other operators, say using ReLU that is more amenable to training with back propagation, see Remark \ref{rem:ReLU} in Appendix \ref{app:remarks}. }
\begin{equation}
\sum_{j=1}^k \gamma_j \bm C^{j}_{\ell}  \z_{\ell} \bm \pi^j(\z_\ell) \approx  \sum_{j=1}^k \gamma_j  \bm{C}_{\ell}^j  \z_{\ell} \cdot \widehat{\bm \pi}^j(\z_\ell) \;\; \doteq\; \bm \sigma\Big([\bm{C}_{\ell}^{1} \z_{\ell}, \dots, \bm{C}_{\ell}^{k} \z_{\ell}]\Big) \quad \in \Re^n,
\label{eqn:soft-residual}
\end{equation}
which is denoted as a nonlinear operator $\bm \sigma(\cdot)$ on outputs of the feature $\z_\ell$ through $k$ banks of filters: $[\bm{C}_{\ell}^{1}, \dots, \bm{C}_{\ell}^{k}]$. Notice that the nonlinearality arises due to a ``soft'' assignment of class membership based on the feature responses from those filters.  
Overall, combining \eqref{eqn:gradient-descent-transform},  \eqref{eqn:DR-gradient-transform}, and \eqref{eqn:soft-residual}, 
the increment feature transform from $\z_{\ell}$ to $\z_{\ell+1}$ now becomes:
\begin{equation}
\z_{\ell+1}  \; \propto \;  \z_\ell +  \eta \cdot  \bm E_{\ell} \z_{\ell} - \eta\cdot  \bm \sigma\Big([\bm{C}_{\ell}^{1} \z_{\ell}, \dots, \bm{C}_{\ell}^{k} \z_{\ell}]\Big) \quad \mbox{subject to} \quad \z_{\ell +1} \in \mathbb{S}^{n-1},
\label{eqn:layer-approximate}
\end{equation}
with the nonlinear function $\bm \sigma(\cdot)$ defined above and $\bm \theta_\ell$ collecting all the layer-wise parameters including $\E_\ell, \bm{C}_{\ell}^{j}, \gamma_j$ and $\lambda$, and with features at each layer always ``normalized'' onto a sphere $\mathbb S^{n-1}$, denoted as $\mathcal P_{\mathbb S^{n-1}}$. The form of increment in \eqref{eqn:layer-approximate} can be illustrated by a diagram in  Figure~\ref{fig:arch}.

\paragraph{Deep Network from Rate Reduction.} Notice that the increment is constructed to emulate the gradient ascent for the rate reduction $\Delta R$. Hence by transforming the features iteratively via the above process, we expect the rate reduction to increase, as we will see in the experimental section. This iterative process, once converged say after $L$ iterations, gives the desired feature map $f(\x, \bm \theta)$ on the input $\z_0 = \x$, precisely in the form of a {\em deep network}, in which each layer has the structure shown in Figure~\ref{fig:arch}:
\begin{equation}
f(\x, \bm \theta) =  \phi^L \circ \phi^{L-1} \circ \cdots \circ \phi^0(\x), \quad\mbox{with} \quad \phi^\ell(\z_\ell, \bm \theta_\ell) \; \doteq \; \mathcal{P}_{\mathbb{S}^{n-1}}[\z_{\ell} + \eta\cdot g(\z_{\ell}, \bm \theta_{\ell})].
\label{eqn:ReduNet}
\end{equation}
As this deep network is derived from maximizing the rate \textbf{redu}ced, we call it the \textbf{ReduNet}. Notice that all parameters of the network are explicitly constructed layer by layer in a {\em forward propagation} fashion. Once constructed, there is no need of any additional supervised learning, say via back propagation. As suggested in \cite{yu2020learning}, the so learned features can be directly used for classification via a nearest subspace classifier. 
\vspace{-0.1in}

\paragraph{Comparison with Other Approaches and Architectures.}
Structural similarities between deep networks and iterative optimization schemes, especially those for solving sparse coding, have been long noticed. In particular, \cite{gregor2010learning} has argued that algorithms for sparse coding, such as the FISTA algorithm \citep{BeckA2009}, can be viewed as a deep network and be trained for better coding performance, known as LISTA. Later \cite{monga2019algorithm,deep-sparse} have proposed similar interpretation of deep networks as unrolling algorithms for sparse coding.  Like all networks that are inspired by unfolding certain iterative optimization schemes, the structure of the ReduNet naturally contains a skip connection between adjacent layers as in the ResNet \citep{he2016deep}. Remark \ref{rem:acceleration} in Appendix \ref{app:remarks} discusses possible improvement to the basic gradient scheme that may introduce additional skip connections beyond adjacent layers. As illustrated in Figure \ref{fig:arch}, the remaining $k+1$ parallel channels $\bm E, \bm C^j$ of the ReduNet actually draw resemblance to the parallel structures that people later  found empirically beneficial for deep networks, e.g. ResNEXT \citep{ResNEXT} or the mixture of experts (MoE) module adopted in \cite{MoE}. But a major difference here is that all components (layers, channels, and operators) of the ReduNet are by explicit construction from 
first principles and they all have precise optimization, statistical and geometric interpretation. Furthermore, there is no need to learn them from back-propagation, although in principle one still could if further fine-tuning of the network is needed (see Remark \ref{rem:ReLU} of Appendix \ref{app:remarks} for more discussions).

\subsection{Deep Convolution Networks from Shift-Invariant Rate Reduction}\label{sec:shift-invariant}
We next examine ReduNet from the perspective of invariance to transformation. Using the basic and important case of shift/translation invariance as an example, we will show that for data which are compatible with an invariant classifier, the ReduNet construction automatically takes the form of \textit{a (multi-channel) convolutional neural network}, rather than heuristically imposed upon. 

\paragraph{1D Serial Data and  Shift Invariance.}\label{sec:invariance}
For one-dimensional data $\x = [x(0), x(1), \ldots, x(n-1)] \in \Re^n$ under shift symmetry, we take $\mathbb{G}$ to be the group of circular shifts. Each observation $\x^i$ generates a family $\{ \x^i \circ \mathfrak{g} \, | \, \mathfrak{g} \in \mathbb G \}$ of shifted copies, which are the columns of the circulant matrix $\circm(\x^i) \in \Re^{n \times n}$ given by
\begin{equation*}
\circm(\x) \quad\doteq\quad \left[ \begin{array}{ccccc} x(0) & x(n-1) & \dots & x(2) & x(1) \\ x(1) & x(0) & x(n-1) & \cdots & x(2) \\ \vdots & x(1) & x(0) &\ddots & \vdots \\ x(n-2) &  \vdots & \ddots & \ddots & x(n-1) \\ x(n-1) & x(n-2) & \dots & x(1) & x(0)   \end{array} \right] \quad \in \Re^{n \times n}.
\end{equation*}
We refer the reader to Appendix \ref{ap:circulant} or \cite{Kra2012OnCM} for properties of circulant matrices.

What happens if we construct the ReduNet from these families $\bm Z_1 = [ \circm(\x^1), \dots, \circm(\x^m) ]$? The data covariance matrix: 
$$
\bm Z_1 \bm Z_1^* = \left[ \circm(\x^1), \dots, \circm(\x^m) \right] \left[ \circm(\x^1), \dots, \circm (\x^m) \right]^* = \sum_{i =1}^m \circm(\x^i) \circm(\x^i)^* \;\in \Re^{n\times n}
$$
associated with this family of samples is {\em automatically} a (symmetric) circulant matrix. Moreover, because the circulant property is preserved under sums, inverses, and products, the matrices $\bm E_1$ and $\bm C_1^j$ are also automatically circulant matrices, whose application to a feature vector $\bm z \in \Re^n$ can be implemented using circular convolution ``$\circledast$''.
Specifically, we have the following proposition. 

\begin{proposition}[Convolution structures of $\bm E_1$ and $\bm C_1^j$]
The matrix $\E_1 = \alpha\big(\bm I + \alpha \bm Z_1 \bm Z_1^*\big)^{-1}$
is a circulant matrix and represents a circular convolution: 
$$\E_1 \z = \bm e_1 \circledast \z,$$ 
where $\bm e_1 \in \Re^n$ is the first column vector of $\E_1$ and ``$\circledast$'' is circular convolution defined as
\begin{equation*}
    (\bm e_1 \circledast \bm z)_{i} \doteq \sum_{j=0}^{n-1} e_1(j) x(i+ n-j \,\, \textsf{mod} \,\,n).
\end{equation*}
Similarly, the matrices $\bm C_1^j$ associated with any subsets of $\bm Z_1$ are also circular convolutions. 
\label{prop:circular-conv-1}
\end{proposition}

From Proposition~\ref{prop:circular-conv-1}, we have
\begin{equation}
\z_{2} \propto \z_{1} +\eta\cdot g(\z_{1}, \bm \theta_{1}) =  \z_1 + \eta \cdot \bm e_{1} \circledast \z_{1} -  \eta \cdot \bm \sigma\Big([\bm{c}_{1}^{1} \circledast \z_{1}, \dots, \bm{c}_{1}^{k} \circledast \z_{1}]\Big).
\label{eqn:approximate-convolution}
\end{equation}
Because $g( \cdot, \bm \theta_1 )$ consists only of operations that co-vary with cyclic shifts, the features $\bm Z_2$ at the next level again consist of families of shifts: 
$\bm Z_2 = \big[ \circm( \bm x^1 + \eta g( \bm x^1, \bm \theta_1)), \dots,$ $\circm( \bm x^m + \eta g(\bm x^m, \bm \theta_m)) \big].$ 
Continuing inductively, we see that all matrices $\bm E_\ell$ and $\bm C_\ell^j$ based on such $\bm Z_\ell$ are circulant. By virtue of the properties of the data, ReduNet has taken the form of a convolutional network, {\em with no need to explicitly choose this structure!}

\paragraph{The Role of Multiple Channel Lifting and Sparsity.}
There is one problem though: In general, the set of all circular permutations of a vector $\z$ give a full-rank matrix. That is, the $n$ ``augmented'' features associated with each sample (hence each class) typically already span the entire space $\Re^n$. The MCR$^2$ objective \eqref{eq:mcr2-formulation} will not be able to distinguish classes as different subspaces.\footnote{All shifted versions  delta function $\delta(n)$ can generate any other signal as their (dense) weighted sum.}

One natural remedy is to improve the separability of the data by ``lifting'' the original signal to a higher dimensional space, e.g., by taking their responses to multiple, filters $\bm k_1, \ldots, \bm k_C \in \Re^n$: 
\begin{equation}
\bm z[c] = \bm k_c \circledast \bm x  =  \circm(\bm k_c) \bm x \quad \in \Re^n, \quad c = 1, \ldots, C.
\label{eqn:lift-1d}
\end{equation} 
The filers can be pre-designed invariance-promoting filters,\footnote{For 1D signals like audio, one may consider the conventional short time Fourier transform (STFT); for 2D images, one may consider 2D wavelets as in the ScatteringNet \citep{scattering-net}.} or adaptively learned from the data,\footnote{For learned filters, one can learn filters as the principal components of samples as in the PCANet \citep{chan2015pcanet} or from convolution dictionary learning \citep{li2019multichannel,qu2019nonconvex}.} or randomly selected as we do in our experiments. This operation lifts each original signal $\x \in \Re^n$ to a $C$-channel feature, denoted as $\bar{\z}  \doteq [\z[1], \ldots, \z[C]]^* \in \Re^{C\times n}$. 
Then, we may construct the ReduNet on vector representations of $\bar{\z}$, denoted as
$\vec(\bar\z) \doteq [\z[1]^*, \ldots, \z[C]^*] \in \Re^{nC}$. 
The associated circulant version $ \circm(\bar{\z})$ and its data covariance matrix, denoted as $\bar{\bm \Sigma}$, for all its shifted versions are given as:
\begin{equation}
 \circm(\bar{\z}) \doteq \left[\begin{smallmatrix}
    \circm(\z[1])  \\ \vdots \\ \circm(\z[C]) \end{smallmatrix} \right] \; \in \Re^{nC\times n},\quad  \bar{\bm \Sigma} \doteq 
    \left[\begin{smallmatrix}
    \circm(\z[1]) \\ \vdots \\ \circm(\z[C]) \end{smallmatrix} \right]
    \left[\begin{smallmatrix}\circm(\z[1])^*,\ldots, \circm(\z[C])^*\end{smallmatrix} \right]\; \in \Re^{nC\times nC},
    \label{eqn:W-multichannel}
\end{equation}
where $\circm(\z[c]) \in \Re^{n\times n}$ with $c \in [C]$ is the circulant version of the $c$-th channel of the feature $\bar \z$. Then the columns of $\circm(\bar\z)$  will only span at most an $n$-dimensional proper subspace in $\Re^{nC}$. 

However, this simple (linear) lifting operation is not sufficient to render the classes separable yet -- features associated with other classes will span the {\em same} $n$-dimensional subspace. This reflects a fundamental conflict between linear (subspace) modeling and invariance.

One way of resolving this conflict is to leverage additional structure within each class, in the form of {\em sparsity}: Signals within each class are not generated as arbitrary linear combinations of some base atoms (or motifs), but only {\em sparse} combinations of them and their shifted versions. Let $\mathcal{D}_j$ denote a collection of atoms associated for class $j$, also known as a dictionary, then each signal $\x$ in this class is sparsely generated as:
\begin{equation*}
    \x = \circm(\mathcal{D}_j)\z
\end{equation*}
for some sparse vector $\z$. Signals in different classes are then generated by different dictionaries whose atoms (or motifs) are incoherent from one another. Due to incoherence, signals in one class are unlikely to be sparsely represented by atoms in any other class. Hence all signals in the $k$ class can be represented as
\begin{equation*}
\x = [\circm(\mathcal{D}_1), \circm(\mathcal{D}_2), \ldots, \circm(\mathcal{D}_k)] \bar \z
\end{equation*}
where $\bar \z$ is sparse.\footnote{Notice that similar sparse representation models have long been proposed and used for classification purposes in applications such a face recognition, demonstrating excellent effectiveness  \citep{Wright:2009,wagner2012toward}. } There is a vast literature on how to learn the most compact and optimal sparsifying dictionaries from sample data, e.g.  \citep{li2019multichannel,qu2019nonconvex} and subsequently solve the inverse problem and compute the associated sparse code $\z$ or $\bar \z$. 

Nevertheless, here we are not interested in the optimal dictionary and the precise sparse code for each individual signal. We are only interested if the set of sparse codes for each class are collectively separable from those of other classes. Under the assumption of the sparse generative model, if the convolution kernels $\{\bm k_c\}$ match well with the ``transpose'' or ``inverse'' of the above sparsifying dictionaries, also known as the {\em analysis filters} \citep{Cosparse-Nam,Analysis-Filter}, signals in one class will only have high responses to a small subset of those filters and low responses to others (due to the incoherence assumption).  Nevertheless, in practice, often a sufficient number of random filters suffice the purpose of ensuring features of different classes have different response patterns to different filters hence make different classes separable \citep{chan2015pcanet}. As optimal sparse coding is not the focus of this paper, we will use the simple random filter design in our experiments, which is adequate to verify the concept.\footnote{Although better sparse coding schemes may surely lead to better classification performance, at a higher computational cost.} 

\begin{figure}[t]
    \centering
    \includegraphics[width=0.9\linewidth]{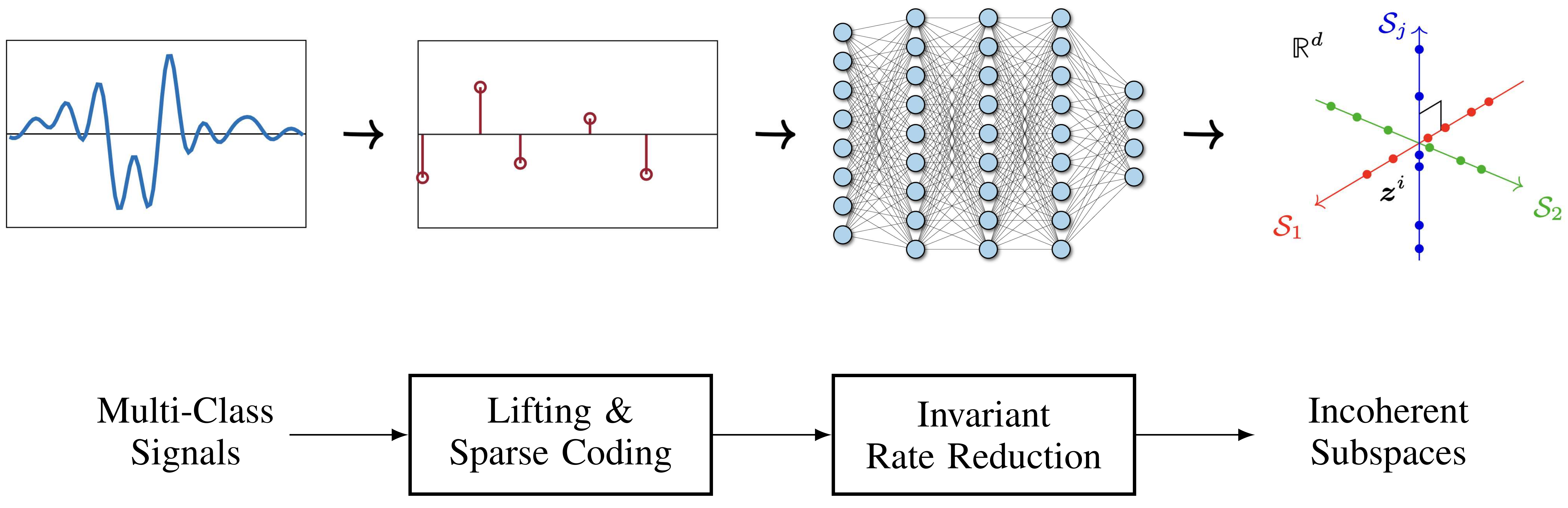}
    \caption{Overview of the process for classifying multi-class signals with shift invariance: Multi-channel lifting and sparse coding followed by a (convolutional) ReduNet for invariant rate reduction. These operations are {\em necessary} to map shift-invariant multi-class signals to incoherent (linear) subspaces. Note that most modern deep neural networks resemble this process.}
    \label{fig:learn-to-classify-diagram}
    \vspace{-0.1in}
\end{figure}

Hence the multi-channel responses $\bar \z$ should be sparse. So to approximate the sparse code $\bar \z$, we may take an entry-wise {\em sparsity-promoting nonlinear thresholding}, say $\bm \tau(\cdot)$, on the filter outputs by setting low (say absolute value below $\epsilon$) or negative  responses to be zero:
\begin{equation*}
\bar \z = \bm \tau \big[\circm(\bm k_1) \x, \ldots, \circm(\bm k_C) \x \big] \quad \in \Re^{n \times C}.
\end{equation*}
One may refer to \citep{Analysis-Filter} for a more systematical study on the design of the sparsifying thresholding operator. Nevertheless, here we are not so interested in obtaining the best sparse codes as long as the codes are sufficiently separable. Hence the nonlinear operator $\bm \tau$ can be simply chosen to be a soft thresholding or a ReLU. 
These presumably sparse features $\bar \z$ can be assumed to lie on a lower-dimensional (nonlinear) submanifold of $\mathbb{R}^{n\times C}$, which can be linearized and separated from the other classes by subsequent ReduNet layers, as illustrated in Figure \ref{fig:learn-to-classify-diagram}. 

The ReduNet constructed from circulant version of these multi-channel features $\bar \z$, i.e., $\circm(\bar\Z) \doteq [ \circm(\bar\z^1), \dots, \circm(\bar\z^m)]$, retains the good invariance properties described above: the linear operators, now denoted as $\bar{\bm E}$ and $\bar{\bm C}^j$, remain block circulant, and represent {\em multi-channel 1D circular convolutions. }
Specifically, we have the following result (see Appendix \ref{ap:multichannel-circulant} for a proof).
\begin{proposition}[Multi-channel convolution structures of $\bar{\bm E}$ and $\bar{\bm C}^j$]
The matrix 
\begin{equation}
\label{eq:def-E-bar}
\bar\E \doteq \alpha\left(\bm I + \alpha \circm(\bar\Z) \circm(\bar\Z)^* \right) ^{-1}   
\end{equation}
is block circulant, i.e.,
\begin{equation*}
    \bar{\bm E} = 
    \left[\begin{smallmatrix}
        \bar{\bm E}_{1, 1} & \cdots & \bar{\bm E}_{1, C}\\
        \vdots & \ddots & \vdots \\
        \bar{\bm E}_{C, 1} & \cdots & \bar{\bm E}_{C, C}\\
    \end{smallmatrix}\right] \in \Re^{nC \times nC},
\end{equation*}
where each $\bar{\bm E}_{c, c'}\in \Re^{n \times n}$ is a circulant matrix. Moreover, $\bar{\bm E}$ represents a multi-channel circular convolution, i.e., for any multi-channel signal $\bar\z \in \Re^{C \times n}$ we have 
$$\bar\E \cdot \vec(\bar\z) = \vec( \bar{\bm e} \circledast \bar\z).$$ 
In above, $\bar{\bm e} \in \Re^{C \times C \times n}$ is a multi-channel convolutional kernel with $\bar{\bm e}[c, c'] \in \Re^{n}$ being the first column vector of $\bar{\bm E}_{c, c'}$, and $\bar{\bm e} \circledast \bar\z \in \Re^{C \times n}$ is the multi-channel circular convolution defined as
\begin{equation*}
    (\bar{\bm e} \circledast \bar\z)[c] \doteq \sum_{c'=1}^C \bar{\bm e}[c, c'] \circledast \bar{\z}[c'], \quad \forall c = 1, \ldots, C.
\end{equation*}
Similarly, the matrices $\bar{\bm C}^j$ associated with any subsets of $\bar{\bm Z}$ are also multi-channel circular convolutions. 
\label{prop:multichannel-circular-conv-1}
\end{proposition}
From Proposition \ref{prop:multichannel-circular-conv-1}, ReduNet is a deep convolutional network for multi-channel 1D signals by construction.\footnote{Unlike Xception nets~\citep{Xception}, these multi-channel convolutions in general are {\em not} depthwise separable. It remains open what additional structures on the data would lead to depthwise separable convolutions.} Figure \ref{fig:learn-to-classify-diagram} illustrates the whole process of rate reduction with such sparse and invariant features. 

\paragraph{Fast Computation in the Spectral Domain.} 
The calculation of $\bar\E$ in \eqref{eq:def-E-bar} requires inverting a matrix of size $nC \times nC$, which has complexity $O(n^3C^3)$. 
By using the relationship between circulant matrix and Discrete Fourier Transform (DFT) of a 1D signal, this complexity can be significantly reduced. 
Specifically, let $\F \in \Co^{n \times n}$ be the DFT matrix, and $\dft(\z) \doteq \F \z \in \Co^{n \times n}$ be the DFT of $\z \in \Re^n$, where $\Co$ denotes the set of complex numbers. 
We have 
\begin{equation}
\label{eq:circ-dft-main}
    \circm(\z) = \F^* \diag(\dft(\z)) \F. 
\end{equation}  
We refer the reader to Appendix \ref{ap:1D-shift} for properties of circulant matrices and DFT. 
By using the relation in \eqref{eq:circ-dft-main}, $\bar{\E}$ can be computed as
\begin{equation*}
    \bar\E = 
    \left[\begin{smallmatrix}
    \F^* & \bm 0 & \bm 0  \\
    \bm 0 & \footnotesize{\ddots} & \bm 0 \\
    \bm 0 & \bm 0 & \F^*
    \end{smallmatrix}\right]
    \cdot \alpha \left(\I + \alpha
    \left[\begin{smallmatrix}
    \D_{11} & \cdots & \D_{1C} \\
    {\footnotesize \vdots} & {\footnotesize \ddots} & {\footnotesize \vdots} \\
    \D_{C1} & \cdots & \D_{CC}
    \end{smallmatrix}\right]
    \right)^{-1} \cdot
    \left[\begin{smallmatrix}
    \F & \bm 0 & \bm 0  \\
    \bm 0 & \footnotesize{\ddots} & \bm 0 \\
    \bm 0 & \bm 0 & \F
    \end{smallmatrix}\right],
\end{equation*}
where $\D_{cc'} \doteq \sum_{i=1}^m \diag(\dft(\z^i[c])) \cdot \diag(\dft(\z^i[c']))^* \in \Co^{n \times n}$ is a diagonal matrix. 
The matrix in the inverse operator is a block diagonal matrix after a permutation of rows and columns. 
Hence, to compute $\bar \E$ and $\bar{\C}^j \in \Re^{nC \times nC}$, we only need to compute in the frequency domain the inverse of $C\times C$ blocks for $n$ times and the overall complexity is $O(nC^3)$. 

The benefit of computation with DFT motivates us to construct the ReduNet in the spectral domain. 
Let $\{\bar\z^i \in \Re^{n \times C}\}_{i=1}^m$ be a collection of multi-channel 1D signals and represent it as a matrix $\bar\Z  \in \Re^{C \times n \times m}$. 
Then, we define the \emph{shift invariant coding rate reduction} for $\bar\Z$ as
\begin{multline*}
    \Delta R_\circm(\bar\Z, \bm{\Pi}) \doteq \frac{1}{n}\Delta R(\circm(\bar\Z), \bar{\bm{\Pi}}) \\= 
    \frac{1}{2n}\log\det \Bigg(\I + \alpha  \circm(\bar\Z)  \circm(\bar\Z)^{*} \Bigg) 
    - \sum_{j=1}^{k}\frac{\gamma_j}{2n}\log\det\Bigg(\I + \alpha_j  \circm(\bar\Z) \bar{\bm{\Pi}}^{j} \circm(\bar\Z)^{*} \Bigg),
\end{multline*}
where $\alpha = \frac{Cn}{mn\epsilon^{2}} = \frac{C}{m\epsilon^{2}}$, $\alpha_j = \frac{Cn}{\textsf{tr}\left(\bm{\Pi}^{j}\right)n\epsilon^{2}} = \frac{C}{\textsf{tr}\left(\bm{\Pi}^{j}\right)\epsilon^{2}}$, $\gamma_j = \frac{\textsf{tr}\left(\bm{\Pi}^{j}\right)}{m}$, and $\bar{\bm \Pi}^j$ is augmented membership matrix in an obvious way.
The normalization factor $n$ is introduce because the circulant matrix $\circm(\bar\Z)$ contains $n$ (shifted) copies of each signal.

Let $\dft(\bar\Z) \in \Co^{C \times n \times m}$ be data in spectral domain obtained by taking DFT on each channel of each signal $\bar\z^i$ and denote $\dft(\bar\Z)(p) \in \Co^{C \times m}$ the $p$-th slice of $\dft(\bar\Z)$ on the second dimension. 
Then, the gradient of $\Delta R_\circm(\bar\Z, \bm{\Pi})$ w.r.t. $\bar\Z$ can be computed from the expansion $\bar\cE \in \Co^{C \times C \times n}$ and compression $\bar\cC^j \in \Co^{C \times C \times n}$ operators in the spectral domain, defined as
\begin{eqnarray*}
    \bar\cE(p) &\doteq& \alpha \cdot \left[\I + \alpha \cdot \dft(\bar\Z)(p) \cdot \dft(\bar\Z)(p)^* \right]^{-1} \quad \in \Co^{C\times C}, \\
    \bar\cC^j(p) &\doteq& \alpha_j \cdot\left[\I + \alpha_j \cdot \dft(\bar\Z)(p) \cdot \bm{\Pi}_j \cdot \dft(\bar\Z)(p)^*\right]^{-1} \quad \in \Co^{C\times C}.
\end{eqnarray*}
In above, $\bar\cE(p)$ (resp., $\bar\cC^j(p)$) is the $p$-th slice of $\bar\cE$ (resp., $\bar\cC^j$) on the last dimension. 
Specifically, we have the following result (see Appendix \ref{ap:1D-shift} for a proof).

\begin{theorem} [Computing multi-channel convolutions $\bar{\bm E}$ and $\bar{\bm C}^j$]
\label{thm:1D-convolution}
Let $\bar\U \in \Co^{C \times n \times m}$ and $\bar\W^{j} \in \Co^{C \times n \times m}, j=1,\ldots, k$ be given by 
\begin{eqnarray*}
    \bar\U(p) &\doteq& \bar\cE(p) \cdot \dft(\bar\Z)(p), \\
    \bar\W^{j}(p) &\doteq& \bar\cC^j(p) \cdot \dft(\bar\Z)(p), \quad j=1,\ldots, k,
\end{eqnarray*}
for each  $p \in \{0, \ldots, n-1\}$. Then, we have
\begin{eqnarray*}
    \frac{1}{2n}\frac{\partial \log \det (\I + \alpha \cdot \circm(\bar\Z) \circm(\bar\Z)^{*} )}{\partial \bar\Z} &=& \idft(\bar\U), 
    \\
    \frac{\gamma_j}{2n}\frac{\partial  \log\det (\I + \alpha_j \cdot \circm(\bar\Z) \bm \bar{\bm \Pi}^j \circm(\bar\Z)^{*})}{\partial \bar\Z}&=&
    \gamma_j \cdot \idft(\bar\W^{j} \bm \Pi^j).
\end{eqnarray*}
In above, $\idft(\bar\U)$ is the time domain signal obtained by taking inverse DFT on each channel of each signal in $\bar\U$. 
\end{theorem}

By this result, the gradient ascent update in \eqref{eqn:gradient-descent} (when applied to $\Delta R_\circm(\bar\Z, \bm{\Pi})$) can be equivalently expressed as an update in spectral domain on $\bar\V_\ell \doteq \dft(\bar\Z_\ell)$ as
\begin{equation*}
    \bar\V_{\ell+1}(p) \; \propto \; \bar\V_{\ell}(p) + \eta \; \Big(\bar\cE_\ell(p) \cdot \bar\V_\ell(p) - \sum_{j=1}^k \gamma_j \bar\cC_\ell^j(p) \cdot \bar\V_\ell(p)\Pi^j \Big), \quad p = 0, \ldots, n-1,
\end{equation*}
and a ReduNet can be constructed in a similar fashion as before. For implementation details, we refer the reader to Algorithm~\ref{alg:training-1D} of Appendix~\ref{ap:1D-shift}.

\paragraph{Connections to Recurrent and Convolutional Sparse Coding.}
The sparse coding perspective of \cite{gregor2010learning} has later been extended to recurrent and convolutional networks for serial data, e.g. \cite{Wisdom2016InterpretableRN,sparse-land,sulam2018multilayer,monga2019algorithm}. Although both sparsity and convolution have long been advocated as desired characteristics for deep networks, their necessity and precise role have never been clearly and rigorously justified, at least not directly from the objective of the network, say classification. In our framework, we see how  multi-channel convolutions ($\bar{\bm E}, \bar{\bm C}^j$), different nonlinear activations ($\widehat{\bm \pi}^j, \bm \tau$), and the sparsity requirement are derived {\em from}, rather than heuristically proposed for, the objective of maximizing rate reduction of the features while enforcing shift invariance. 

\paragraph{2D Images and  Translation Invariance.}
In the case of classifying images invariant to arbitrary 2D translation, for simplicity we may view the image (feature) $\z \in \Re^{(W\times H)\times C}$ as a function defined on a torus $\mathcal{T}^2$ (discretized as a  $W\times H$ grid) and consider $\mathbb{G}$ to be the (Abelian) group of all 2D (circular) translations on the torus (see Figure \ref{fig:mnist-translation-visualize} in the experiment section for visualization). As we will show in the Appendix \ref{ap:2D-translation}, the associated linear operators $\bar \E$ and $\bar{\C}^j$'s act on the image feature $\z$ as {\em multi-channel 2D circular convolutions}. The resulting network will be a deep convolutional network that shares the same multi-channel convolution structures as conventional CNNs for 2D images \citep{LeNet,krizhevsky2012imagenet}. The difference is that, again, the architectures and parameters of our network are derived from the rate reduction objective, and so are the nonlinear activation $\widehat{\bm \pi}^j$ and $\bm \tau$. Again, our derivation in Appendix \ref{ap:2D-translation} shows that this multi-channel 2D convolutional network can be constructed more efficiently in the spectral domain (see Theorem \ref{thm:2D-convolution} of Appendix \ref{ap:2D-translation} for a rigorous statement and justification).

\paragraph{Sparse Coding and Spectral Computing in Nature.} Interestingly, there have been strong scientific evidences that neurons in the visual cortex encode and transmit information in the rate of spiking, hence the so-called ``spiking neurons'' \citep{spking-neuron-1993,spiking-neuron-book,Belitski5696}. Notice that sparse coding  is also a main characteristic of the visual cortex \citep{olshausen1996emergence}. So remarkably, nature might have already ``learned'' to exploit benefits of the above mathematical principles, in particular the computational efficiency in sparse coding and in the spectral domain for achieving invariant (visual) recognition!

%% file: experiments.tex
\section{Simulations and Experiments}\label{sec:experiments}
\begin{figure}
    \centering
    \begin{subfigure}[b]{\textwidth}
        \centering
        \begin{subfigure}[b]{0.32\textwidth}
            \centering
            \vspace{-0.05in}
            \includegraphics[width=\textwidth]{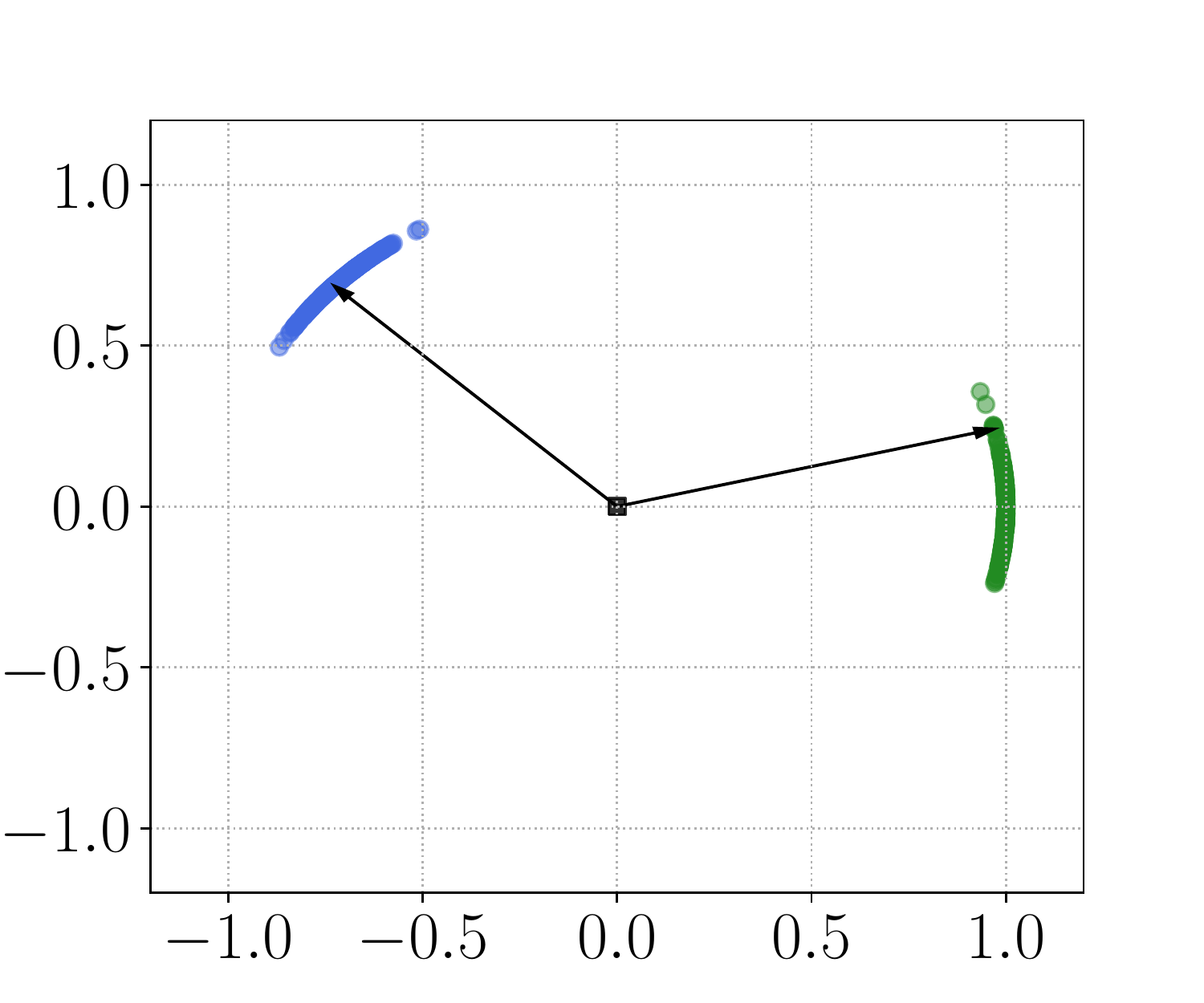}
            \caption{$\X_{\text{train}}$ ($2D$)}
        \label{fig:gaussian2d3d-scatter-heatmap-a}
        \end{subfigure}
        \begin{subfigure}[b]{0.32\textwidth}
            \centering
            \vspace{-0.05in}
            \includegraphics[width=\textwidth]{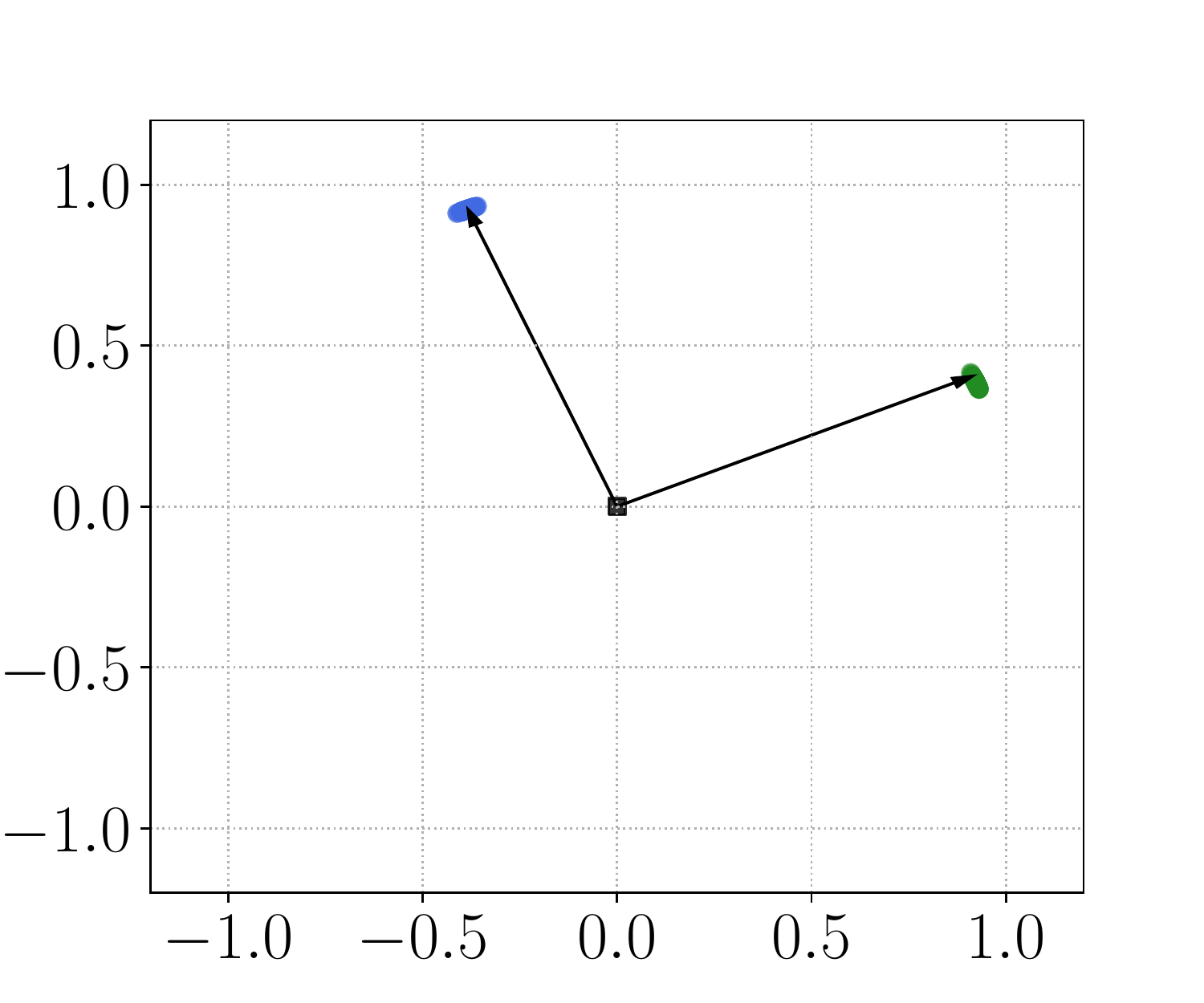}
            \caption{$\Z_{\text{train}}$ ($2D$)}
        \label{fig:gaussian2d3d-scatter-heatmap-b}
        \end{subfigure}
        \begin{subfigure}[b]{0.34\textwidth}
            \centering
            \vspace{-0.05in}
            \includegraphics[width=\textwidth]{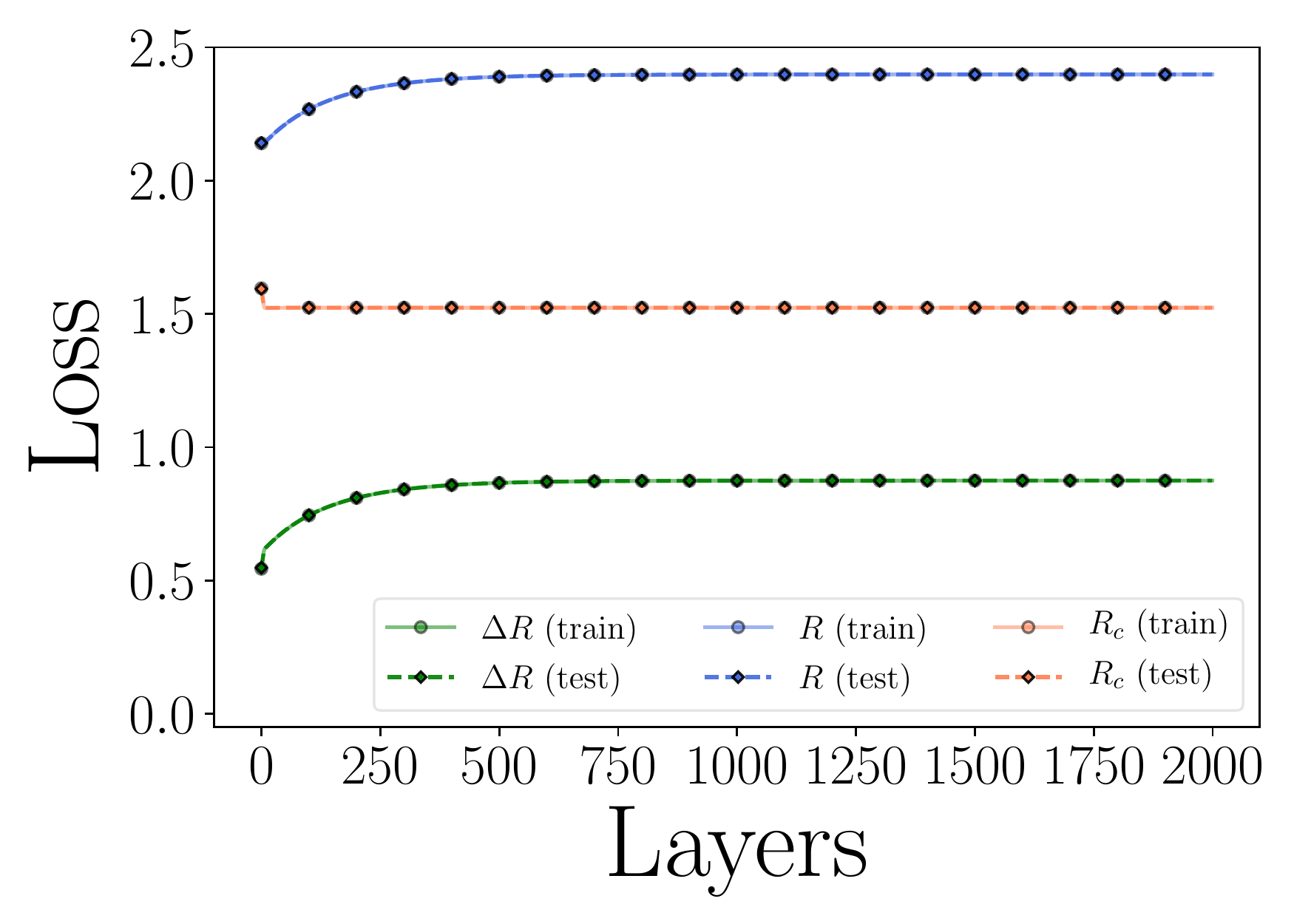}
            \caption{Loss ($2D$)}
        \label{fig:gaussian2d3d-scatter-heatmap-c}
        \end{subfigure}
        \label{fig:gaussian2d3d-1}
        \begin{subfigure}[b]{0.32\textwidth}
            \centering
            \vspace{-0.05in}
            \includegraphics[width=\textwidth]{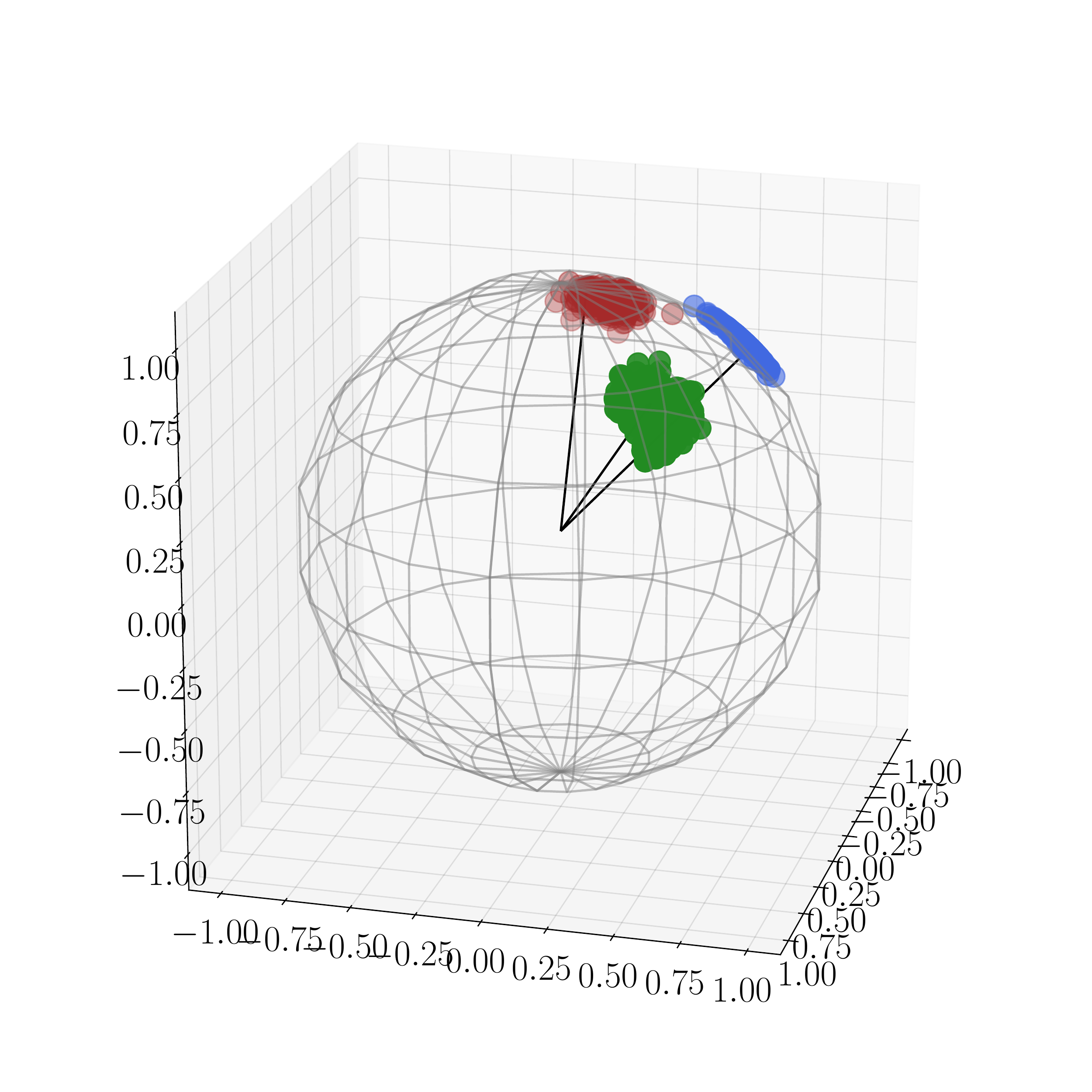}
            \caption{$\X_{\text{train}}$ ($3D$)}
        \label{fig:gaussian2d3d-scatter-heatmap-d}
        \end{subfigure}
        \begin{subfigure}[b]{0.32\textwidth}
            \centering
            \vspace{-0.05in}
            \includegraphics[width=\textwidth]{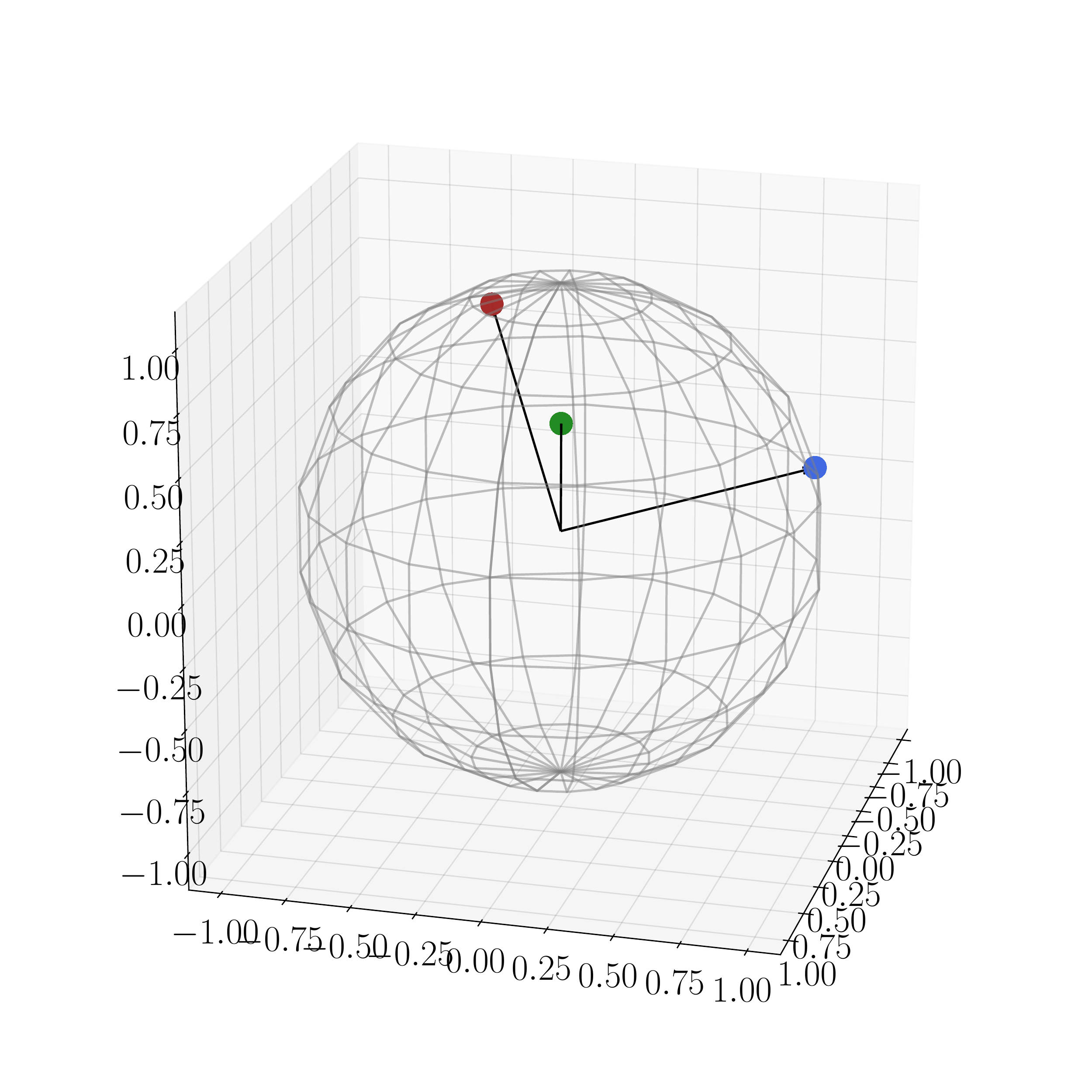}
            \caption{$\Z_{\text{train}}$ ($3D$)}
        \label{fig:gaussian2d3d-scatter-heatmap-e}
        \end{subfigure}
        \begin{subfigure}[b]{0.34\textwidth}
            \centering
            \vspace{-0.05in}
            \includegraphics[width=\textwidth]{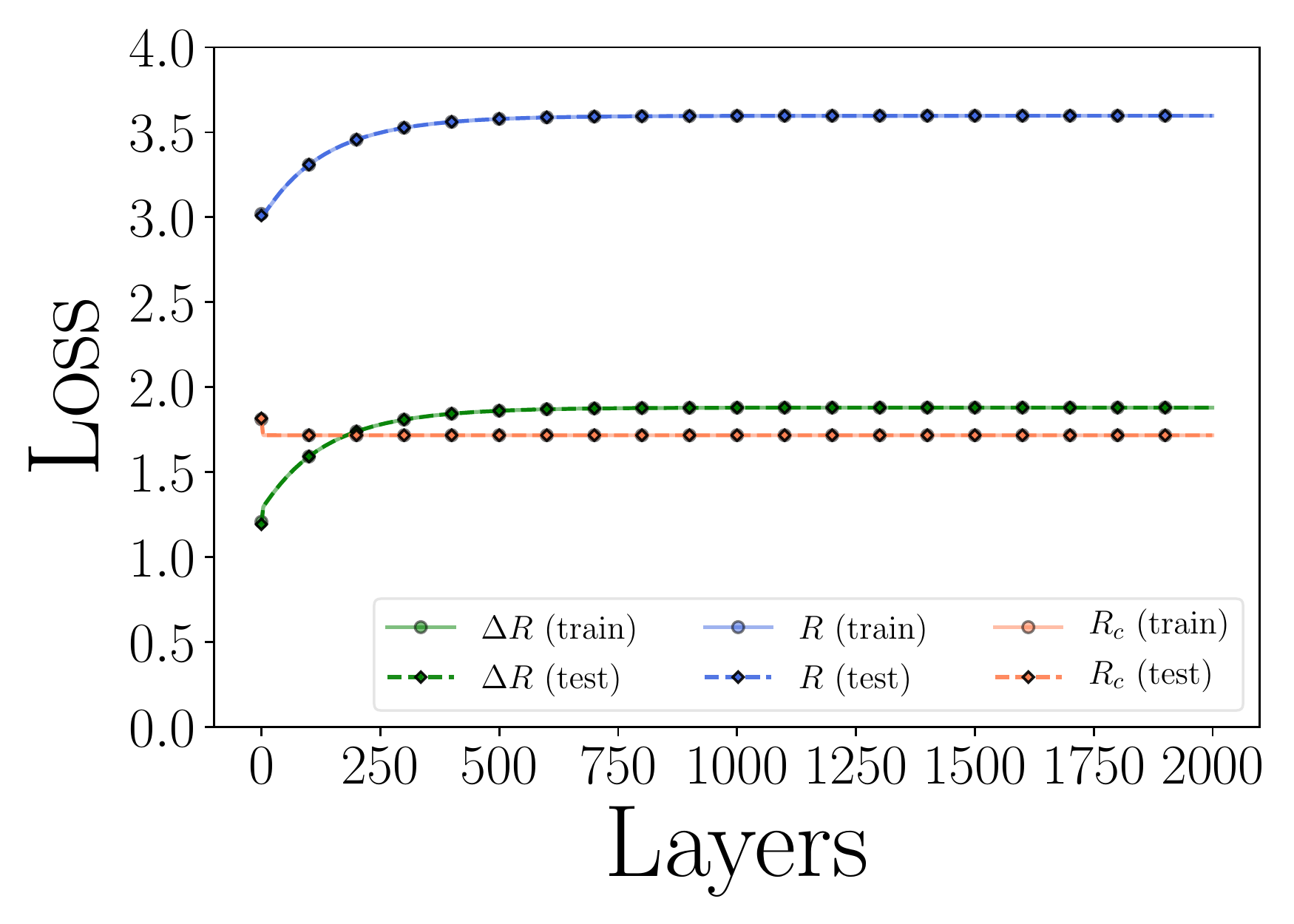}
            \caption{Loss ($3D$)}
        \label{fig:gaussian2d3d-scatter-heatmap-f}
        \end{subfigure}
    \end{subfigure}
    \label{fig:gaussian2d3d-2}
    \vspace{-0.1in}
    \caption{\small Original samples and learned representations for 2D and 3D Mixture of Gaussians. We visualize data points $\X$ (before mapping) and features $\Z$ (after mapping) by scatter plot. In each scatter plot, each color represents one class of samples. We also show the plots for the progression of values of the objective functions.}
    \label{fig:gaussian2d3d-scatter-heatmap}
    \vspace{-0.1in}
\end{figure} 
We now \textit{verify} whether the so constructed ReduNet achieves its design objectives through experiments on synthetic data and real images. The datasets and experiments are chosen to clearly demonstrate the behaviors of the network obtained by our algorithm, in terms of learning the correct discriminative representation and truly achieving invariance. It is not the purpose of this work to push the state of the art on any real datasets with highly engineered networks and systems, although we believe this framework has this potential in the future.  All code is implemented in Python mainly using NumPy. All our experiments are conducted in a computer with 2.8 GHz Intel i7 CPU and 16GB of memory. Implementation details and more experiments and  can be found in Appendix~\ref{sec:appendix-exp}.

\textbf{Learning Mixture of Gaussians in $\mathbb{S}^1$ and $\mathbb{S}^2$.}
Consider a mixture of two Gaussian distributions in $\R^{2}$ that is projected onto $\mathbb{S}^1$. We first generate data points from these two distributions,  $\X_{1}=[\x^1_{1}, \ldots, \x^{m}_{1}] \in \R^{2\times m}$, $\x^{i}_{1} \sim \mathcal{N}(\bm{\mu}_{1}, \sigma_{1} \I)$, and $\bm{\pi}(\x^{i}_{1}) = 1$; 
$\X_{2}=[\x^1_{2}, \ldots, \x^{m}_{2}] \in \R^{2\times m}$, $\x^{i}_{2} \sim \mathcal{N}(\bm{\mu}_{2}, \sigma_{2} \I)$, and $\bm{\pi}(\x^{i}_{2}) = 2$. We set $m=500, \sigma_1=\sigma_2=0.1$ and $\bm{\mu}_{1}, \bm{\mu}_{2} \in \mathbb{S}^1$. Then we project all the data points onto $\mathbb{S}^{1}$, i.e., $\x^i_{j}/\|\x^i_{j}\|_{2}$.
To construct the network (computing $\E, \C^{j}$ for each layer), we set the \# of iterations/layers $L=2,000$,\footnote{It is remarkable to see how easily our framework leads to working deep networks with thousands of layers! But this also indicates the efficiency of the layers is not so high. Remark \ref{rem:acceleration} provides possible ways to improve.} step size $\eta=0.5$, and precision $\epsilon=0.1$. As shown in Figure~\ref{fig:gaussian2d3d-scatter-heatmap-a}-\ref{fig:gaussian2d3d-scatter-heatmap-b},
we can observe that after the mapping $f(\cdot, \bm{\theta})$,
samples from the same class converge to a single cluster and the angle between two different clusters is approximately $\pi/2$, which is well aligned with the optimal solution $\Z_{\star}$ of the MCR$^2$ loss in $\mathbb{S}^1$.
MCR$^2$ loss of features on different layers can be found in Figure~\ref{fig:gaussian2d3d-scatter-heatmap-c}.
Empirically, we find that our constructed network is able to maximize MCR$^2$ loss and converges stably.
Similarly, we consider mixture of three Gaussian distributions in $\R^{3}$ with means 
$\bm{\mu}_{1}, \bm{\mu}_{2}, \bm{\mu}_{3}$ uniformly in $\mathbb{S}^2$,  and variance $\sigma_{1}=\sigma_{2}=\sigma_{3}=0.1$, and all data points are projected onto $\mathbb{S}^2$ (See Figure~\ref{fig:gaussian2d3d-scatter-heatmap-d}-\ref{fig:gaussian2d3d-scatter-heatmap-f}\,). 
We can observe similar behavior as in $\mathbb{S}^{2}$, i.e., samples from the same class converge to one cluster and different clusters are orthogonal to each other. Moreover, we sample new data points from the same distributions for both cases and find that new samples form the same class consistently converge to the same cluster as the training samples. More examples and details can be found in Appendix~\ref{sec:appendix-exp}.

\begin{figure}
    \centering
    \begin{subfigure}[b]{0.2\textwidth}
        \centering
        \includegraphics[width=\textwidth]{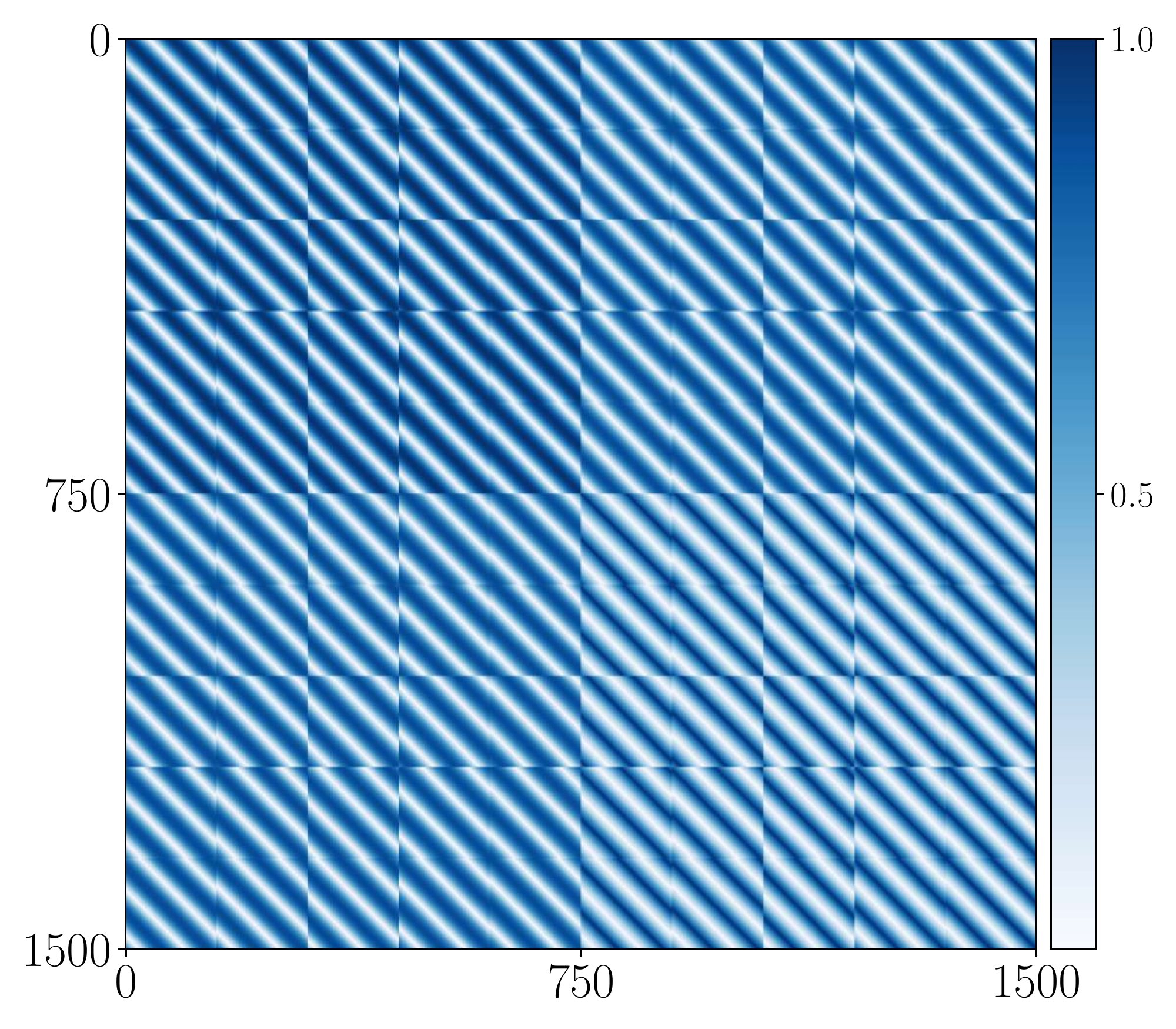}\vspace{-0.05in}
        \caption{\small $\X_{\text{shift}}$ ($1D$)}
    \label{fig:1d-invariance-plots-a}
    \end{subfigure}
    \begin{subfigure}[b]{0.2\textwidth}
        \centering
        \includegraphics[width=\textwidth]{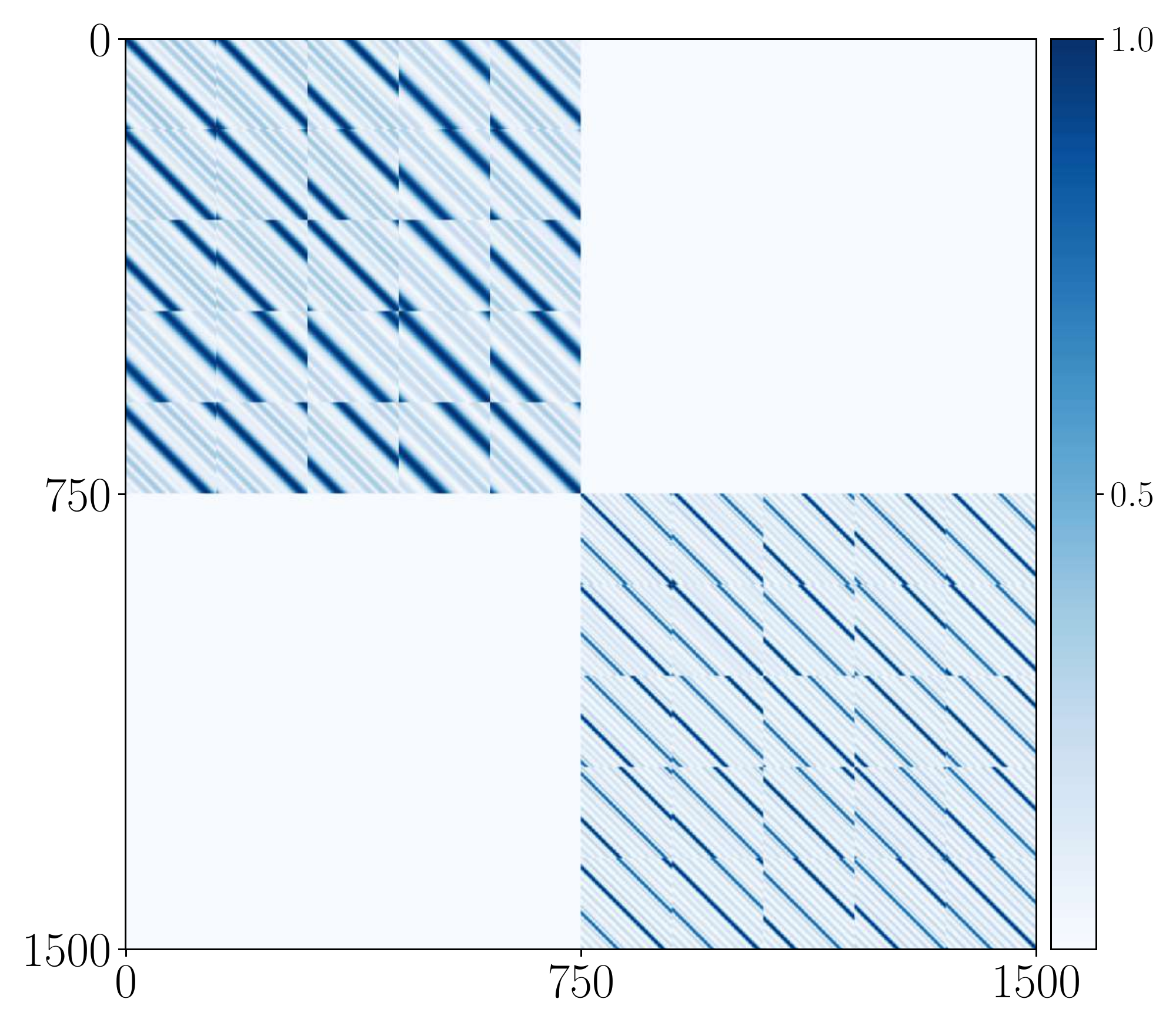}\vspace{-0.05in}
        \caption{\small $\bar{\Z}_{\text{shift}}$ ($1D$)}
    \label{fig:1d-invariance-plots-b}
    \end{subfigure}
    \begin{subfigure}[b]{0.25\textwidth}
        \centering
        \includegraphics[width=\textwidth]{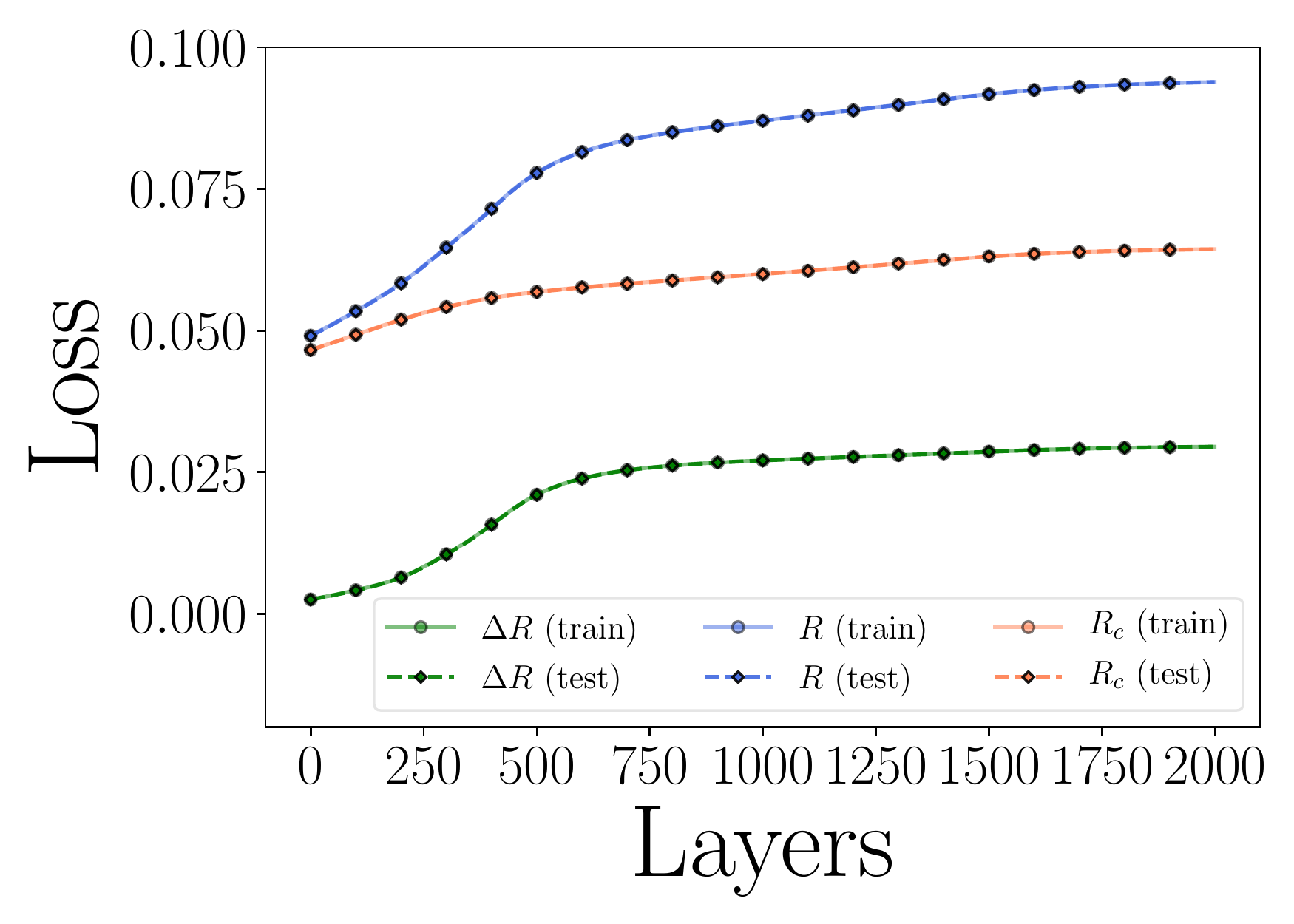}\vspace{-0.05in}
        \caption{\small Loss ($1D$)}
    \label{fig:1d-invariance-plots-c}
    \end{subfigure}
    \begin{subfigure}[b]{0.25\textwidth}
        \centering
        \includegraphics[width=\textwidth]{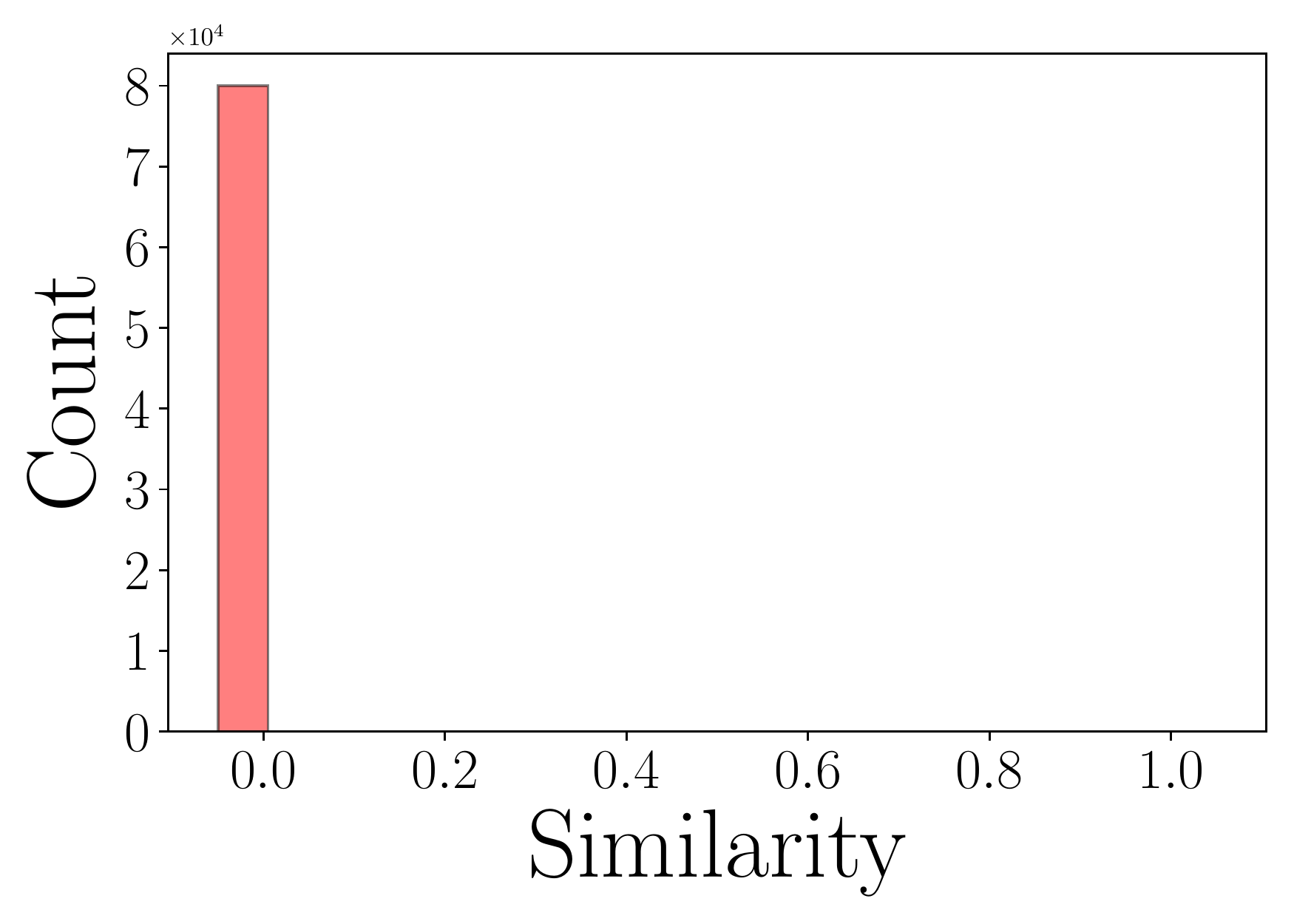}\vspace{-0.05in}
        \caption{\small Similarity ($1D$)}
    \label{fig:1d-invariance-plots-d}
    \end{subfigure}
        \begin{subfigure}[b]{0.2\textwidth}
        \centering
        \includegraphics[width=\textwidth]{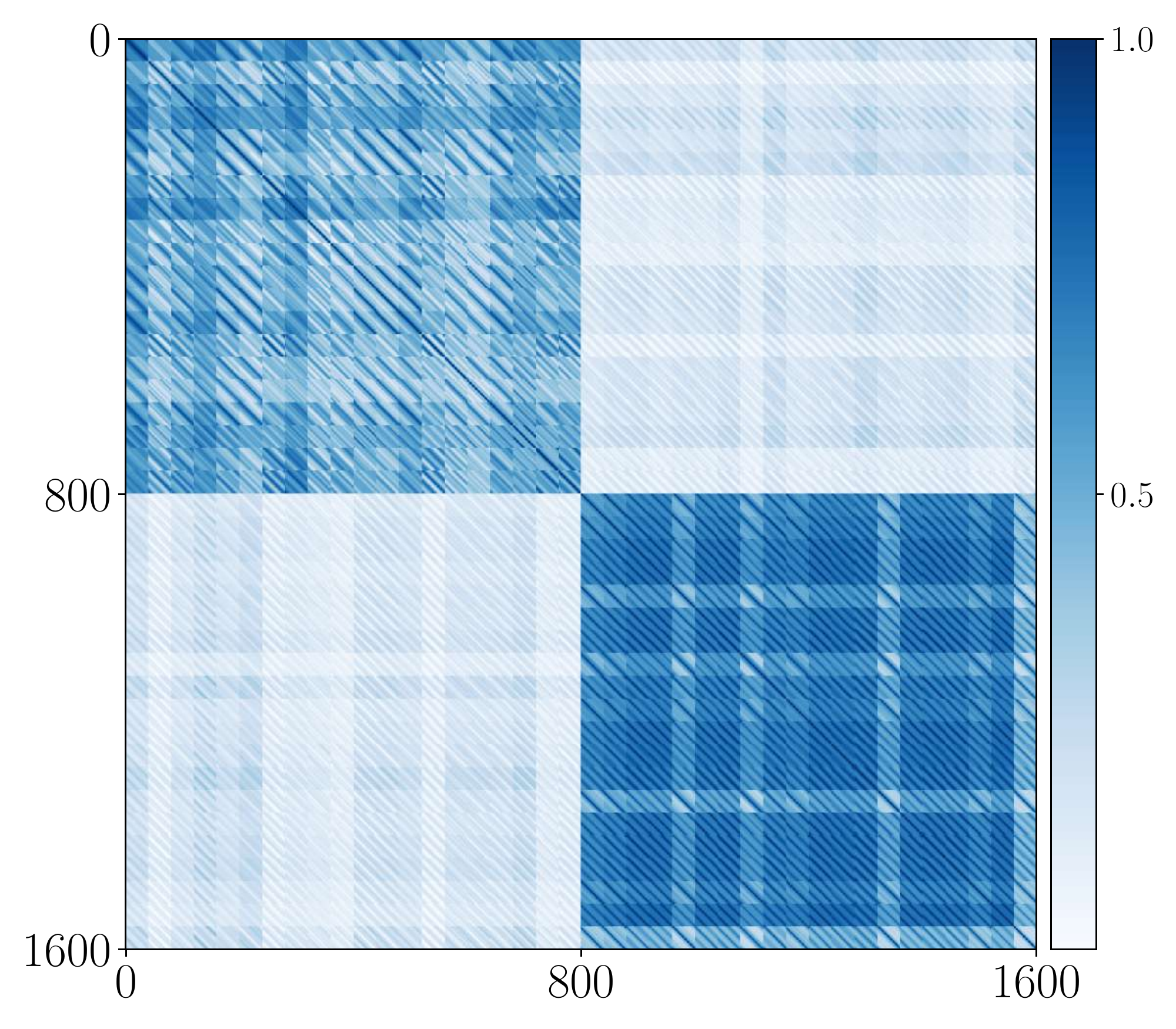}\vspace{-0.05in}
        \caption{\small $\X_{\text{shift}}$ (MNIST)}
    \label{fig:1d-invariance-plots-e}
    \end{subfigure}
    \begin{subfigure}[b]{0.2\textwidth}
        \centering
        \includegraphics[width=\textwidth]{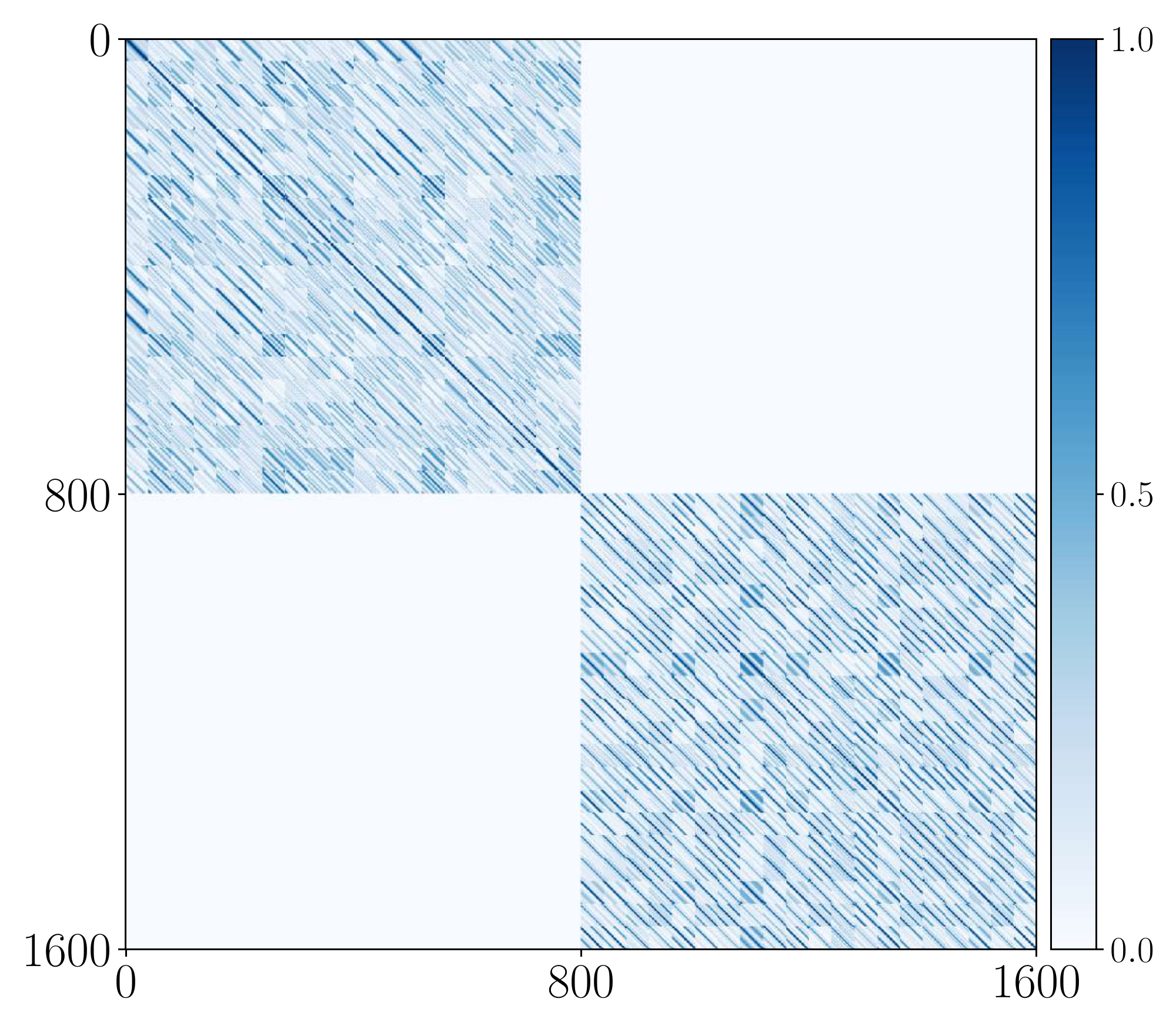}\vspace{-0.05in}
        \caption{\small $\bar{\Z}_{\text{shift}}$ (MNIST)}
    \label{fig:1d-invariance-plots-f}
    \end{subfigure}
    \begin{subfigure}[b]{0.25\textwidth}
        \centering
        \includegraphics[width=\textwidth]{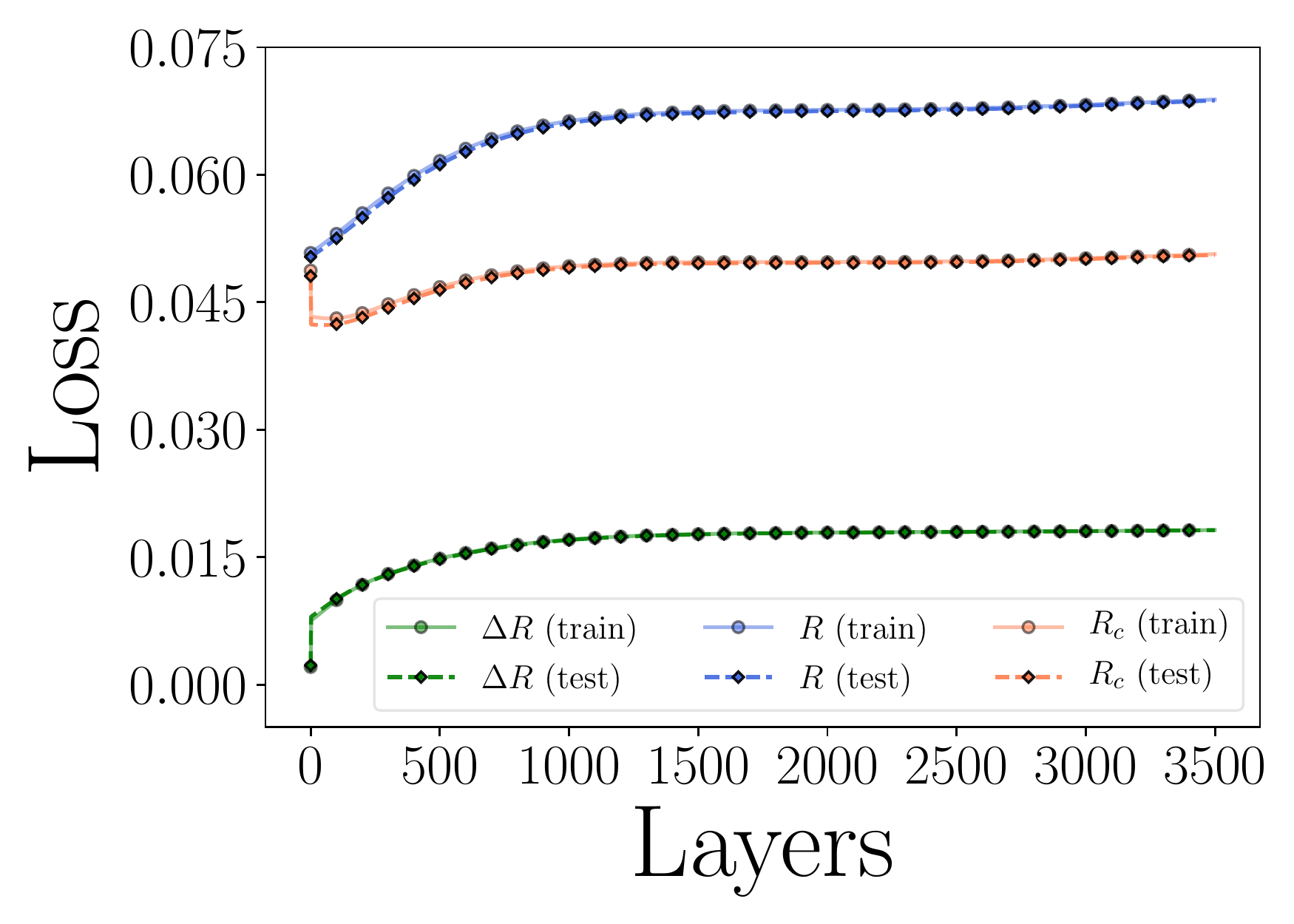}\vspace{-0.05in}
        \caption{\small Loss (MNIST)}
    \label{fig:1d-invariance-plots-g}
    \end{subfigure}
    \begin{subfigure}[b]{0.25\textwidth}
        \centering
        \includegraphics[width=\textwidth]{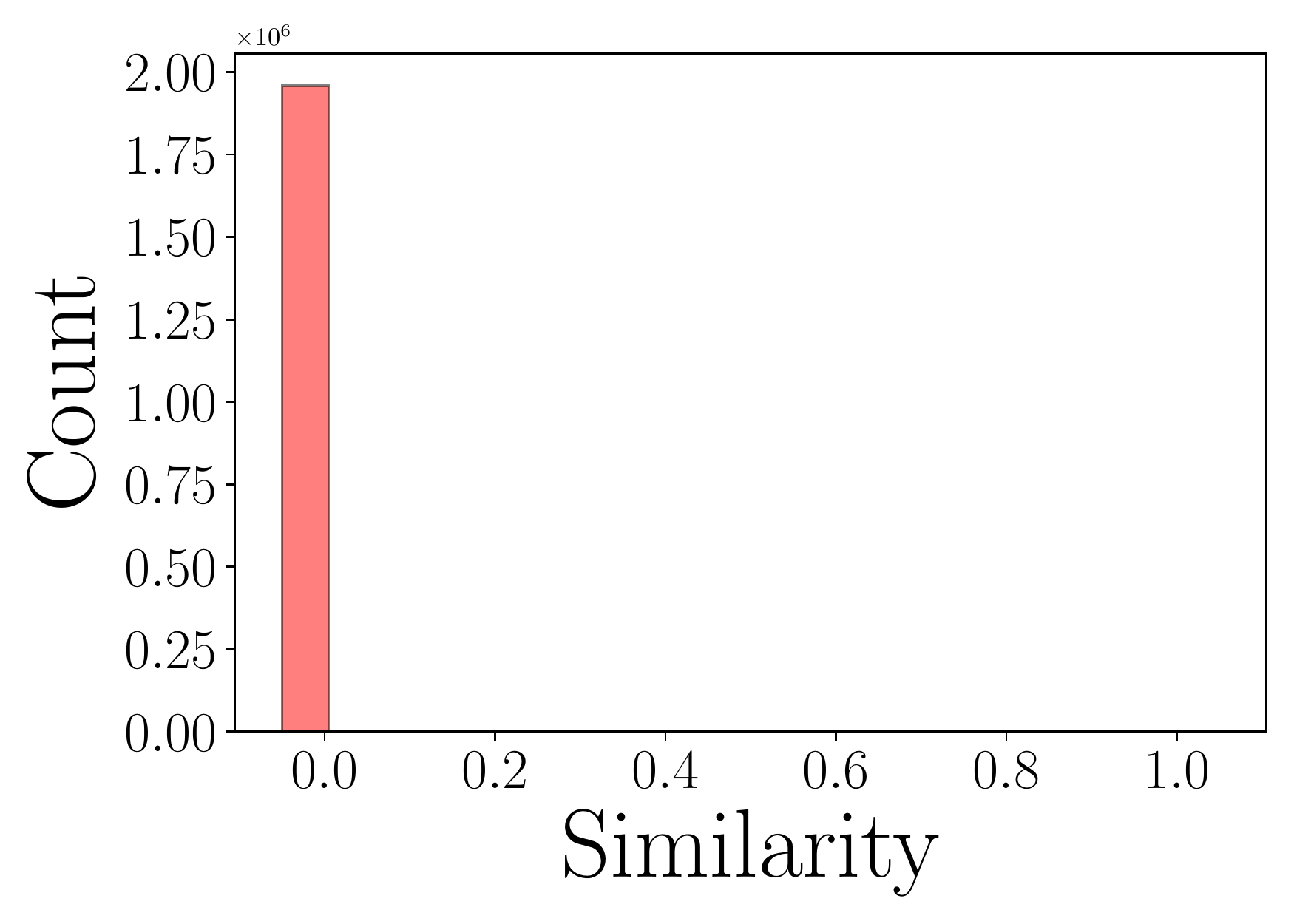}\vspace{-0.05in}
        \caption{\small Similarity (MNIST)}
    \label{fig:1d-invariance-plots-h}
    \end{subfigure}
    \caption{\small 
    Heatmaps of cosine similarity between data $\X_{\text{shift}}$/learned features $\bar{\Z}_{\text{shift}}$, MCR$^2$ loss, and distance between shift samples and subspaces. 
    For (a), (b), (e), (f), we pick one sample from each class and augment the sample with its every possible shifted ones, then calculate the cosine similarity between these augmented samples.
    For (d), (h), we first augment each samples in the dataset with its every possible shifted ones, then we evaluate the cosine similarity (in absolute value) between pairs across classes: for each pair, one sample is from training and one sample is from test which belong to different classes.
    }
    \label{fig:1d-invariance-plots}
    \vspace{-0.1in}
\end{figure}

\textbf{Learning Shift Invariant Features.}
As described in \textsection~\ref{sec:shift-invariant}, by maximizing the rate reduction via Eq.~\eqref{eqn:approximate-convolution}, we are able to explicitly construct operators that are invariant to (circular) shifts. To verify the effectiveness of our proposed network on shift invariance tasks, we apply our network to classify signals sampled from two different 1D functions. The underlying function of the first class is sinusoidal signal $h_1(t) = \textsf{sin}(t) + \epsilon$, and the second class is a composition of \textsf{sign} and \textsf{sin} function, $h_2(t) = \textsf{sign}(\textsf{sin}(t)) + \epsilon$, where $\epsilon\sim \mathcal{N}(0, 0.1)$. (See Figure~\ref{fig:appendix-1d-function-visualize} in Appendix~\ref{sec:appendix-exp}). Each sample is generated by first picking $t_0 \in [0, 10\pi]$, then obtaining $n$ equidistant point within the boundaries $[t_0, t_0 + 2\pi]$ with i.i.d Gaussian noise. 
Detailed implementations for sampling from $h_1$ and $h_2$ can be found in Appendix~\ref{code:sampling-sinusoid}. We generate a dataset which contains $m$ samples, with $m/2$ samples in each class, i.e., $\X=[\X_1, \X_2] \in \R^{n\times m}$. Then each sample is lifted to a $C$-channel feature as defined in \eqref{eqn:lift-1d}, i.e., $\bar{\X}\in \R^{(n\cdot C)\times m}$. 
For training data, We set the number of features $n=150$,  samples $m=400$, channels $C=7$, iterations/layers $L=2,000$, step size $\eta=0.1$, and precision $\epsilon=0.1$. 
We sample the same number of test data points followed by the same procedure.
As shown in Figure~\ref{fig:appendix-sinusoids-heatmaps}, we observe that the network can map the two classes of signals to orthogonal subspaces both on training and test datasets. 
To verify invariance property of the network, we first pick 5 signal samples from each class (from test dataset) and get their corresponding augmented samples by shifting. 
Then we have $m=1,500$ augmented samples for each original signal, $\X_{\text{shift}}\in\R^{150\times 1,500}$, and we visualize the pairwise inner product of $\X_{\text{shift}}$ and their representations, $\bar{\Z}_{\text{shift}}\in\R^{(150\cdot 7)\times 1,500}$ in Figure~\ref{fig:1d-invariance-plots-a}-\ref{fig:1d-invariance-plots-b}. Moreover, we augment every sample (from the test dataset) with its all possible shifted versions and calculate the cosine similarity between their representations and the representations of all the training samples from the other class (in Figure~\ref{fig:1d-invariance-plots-d}). We find that the proposed network can map different classes of signals (including all shifted augmentations) to orthogonal subspaces, to increase the MCR$^2$ loss (shown in Figure~\ref{fig:1d-invariance-plots-c}).

\begin{figure}
    \centering
    \begin{subfigure}[b]{0.28\textwidth}
        \includegraphics[width=\textwidth]{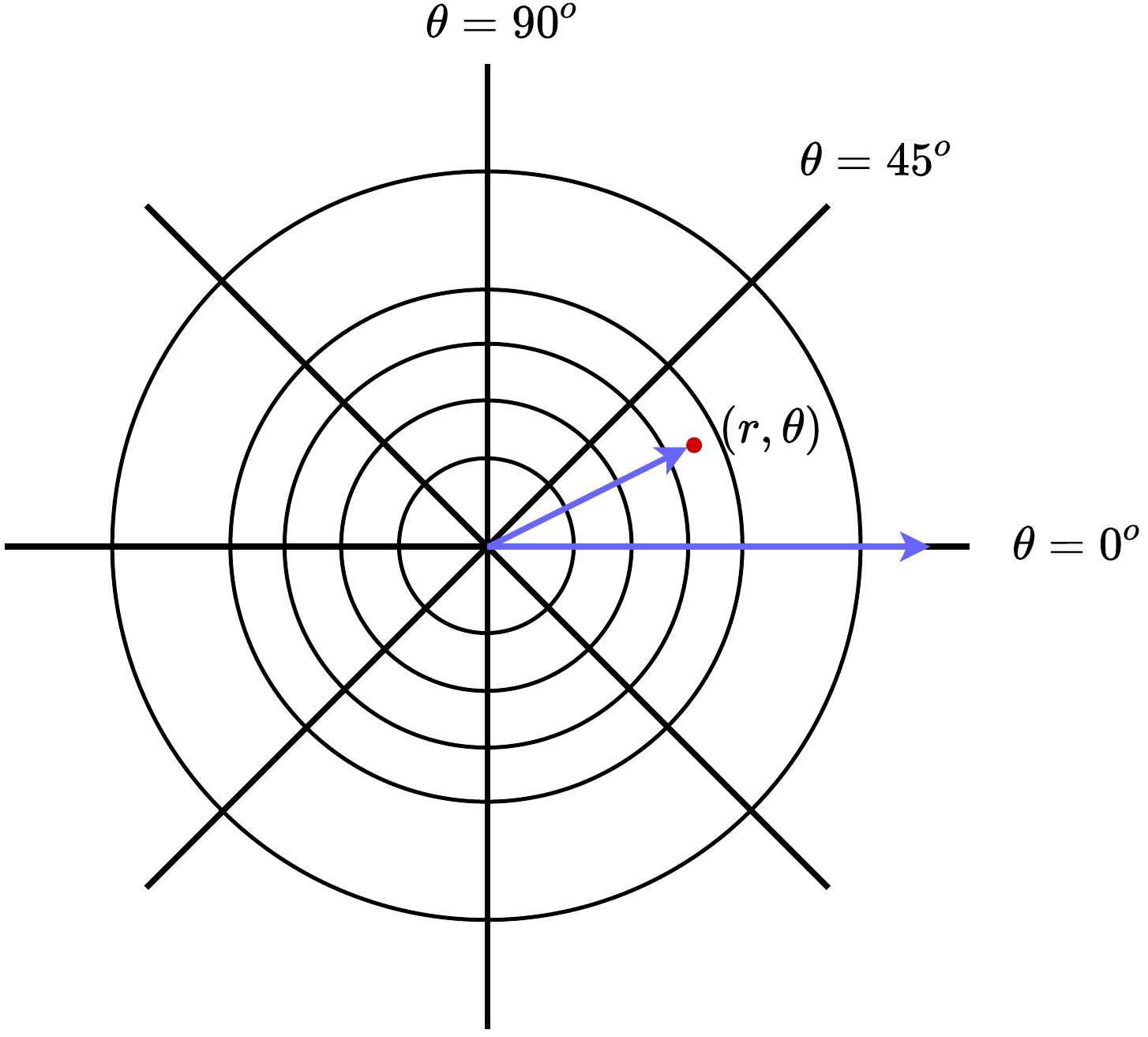}
    \end{subfigure}
    \hspace{5mm}
    \begin{subfigure}[b]{0.28\textwidth}
        \includegraphics[width=\textwidth]{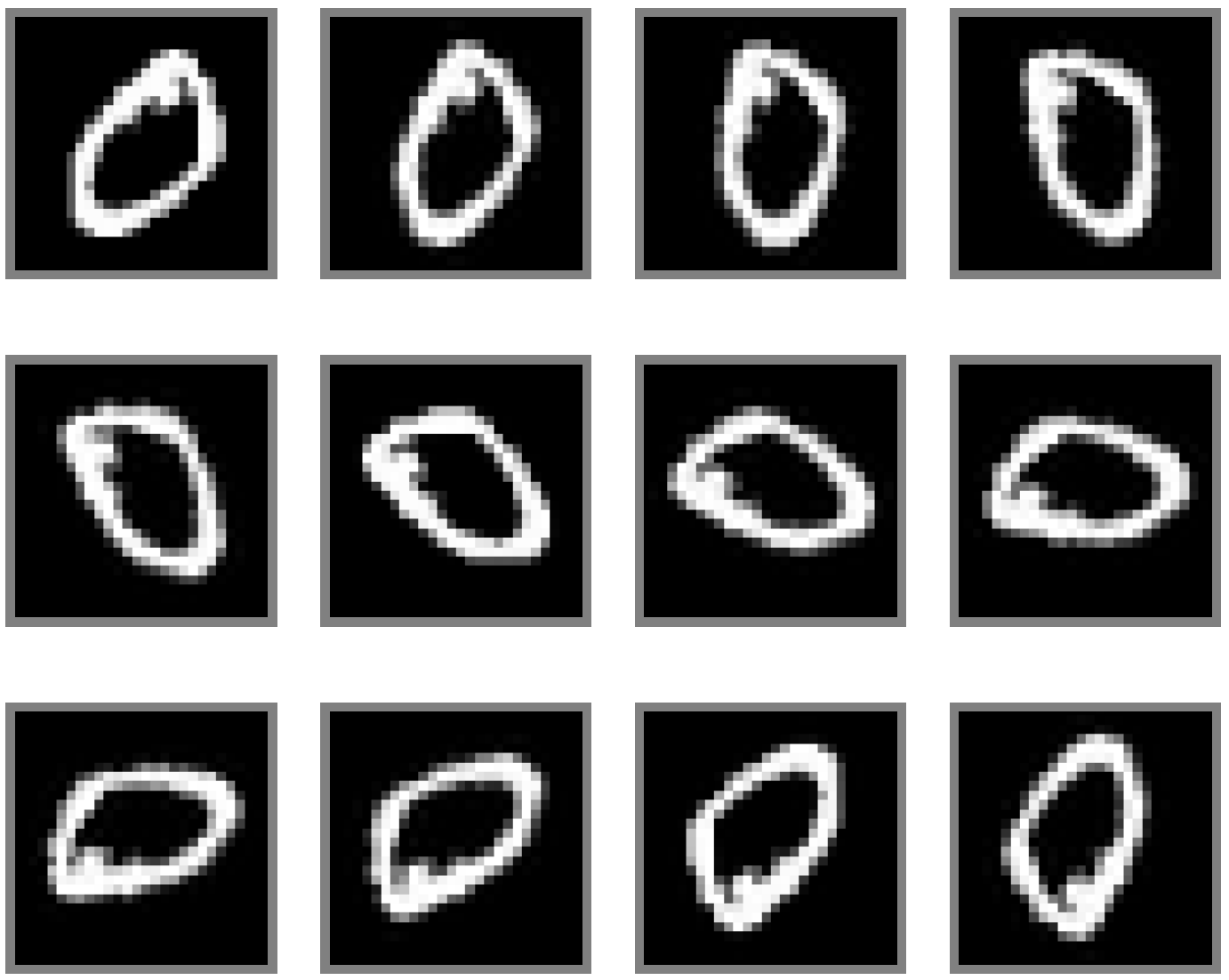}
    \end{subfigure}
    \begin{subfigure}[b]{0.28\textwidth}
        \includegraphics[width=\textwidth]{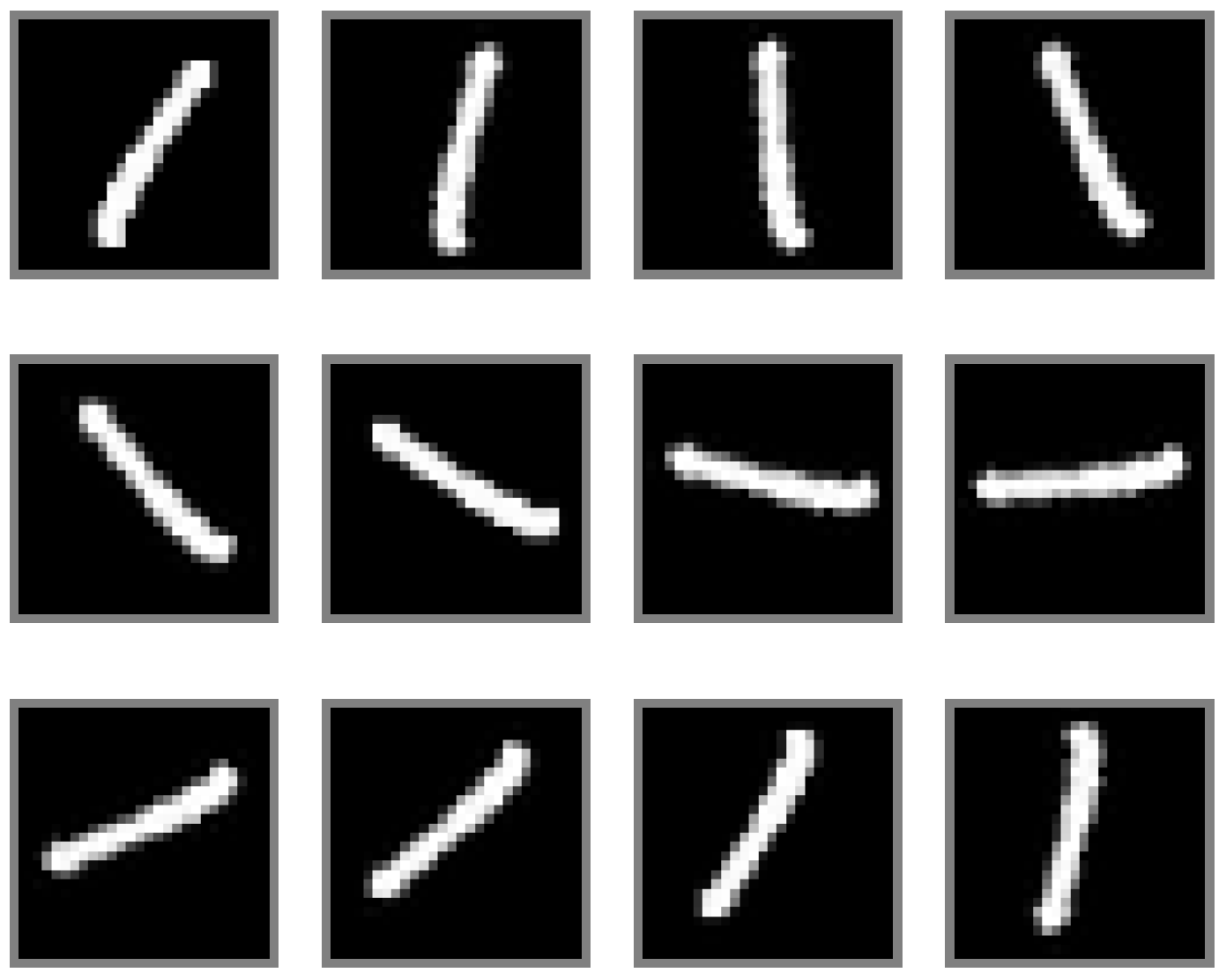}
    \end{subfigure}
    \caption{Examples of rotated images of MNIST digits for testing rotation invariance, each rotated by 18$^{\circ}$. (\textbf{Left}) Diagram for polar coordinate representation;  (\textbf{Right}) Rotated digit `0' and digit `1'.}
    \label{fig:mnist-rotation-visualize}
    \vspace{-0.1in}
\end{figure}

\begin{figure}
    \centering
    \begin{subfigure}[b]{0.25\textwidth}
        \includegraphics[width=\textwidth]{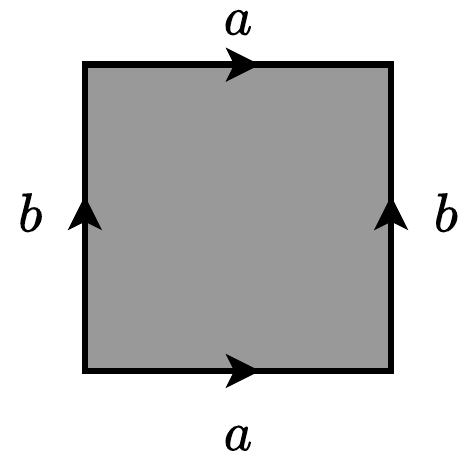}
    \end{subfigure}
    \hspace{8mm}
    \begin{subfigure}[b]{0.28\textwidth}
        \includegraphics[width=\textwidth]{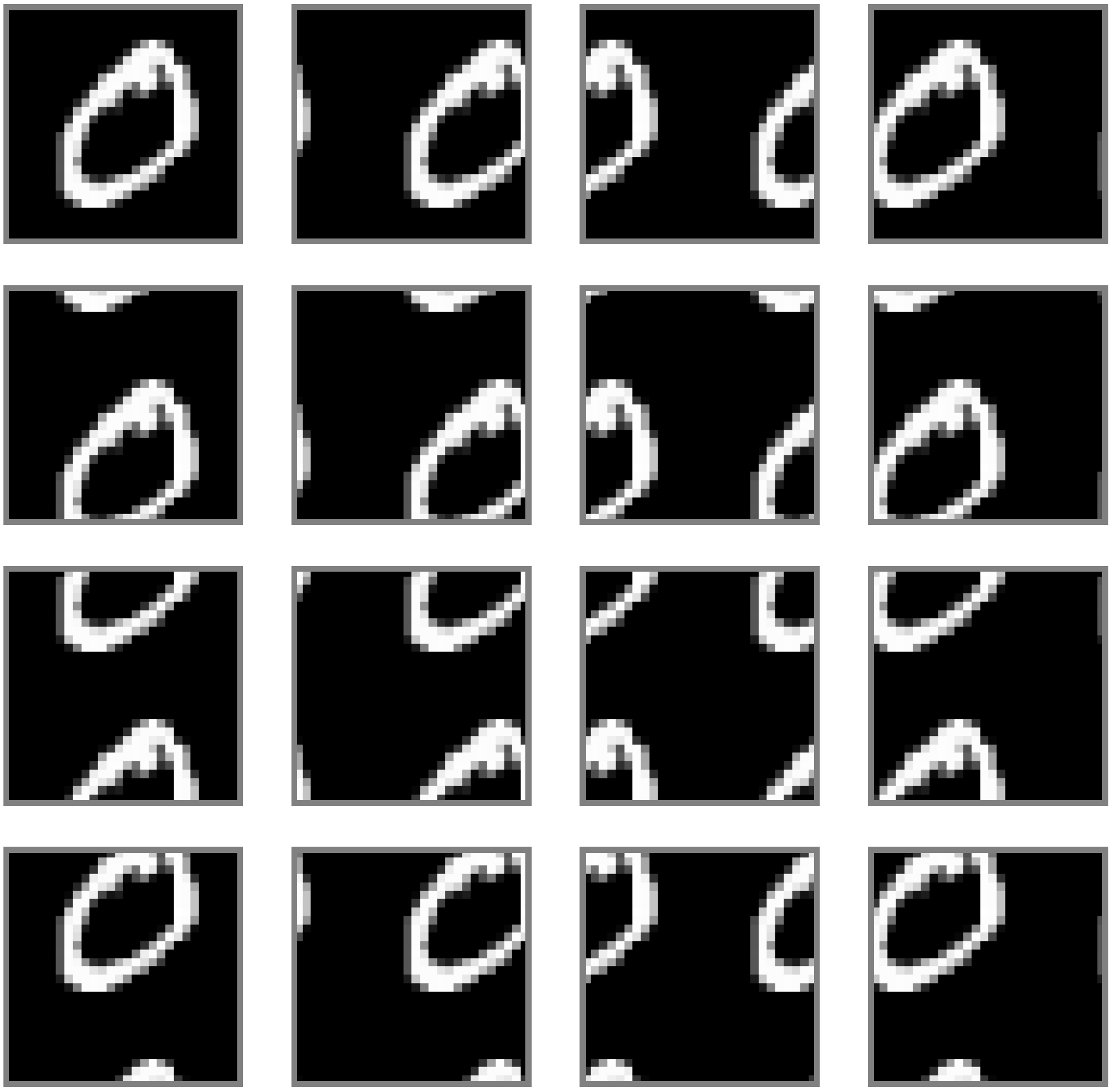}
    \end{subfigure}
    \begin{subfigure}[b]{0.28\textwidth}
        \includegraphics[width=\textwidth]{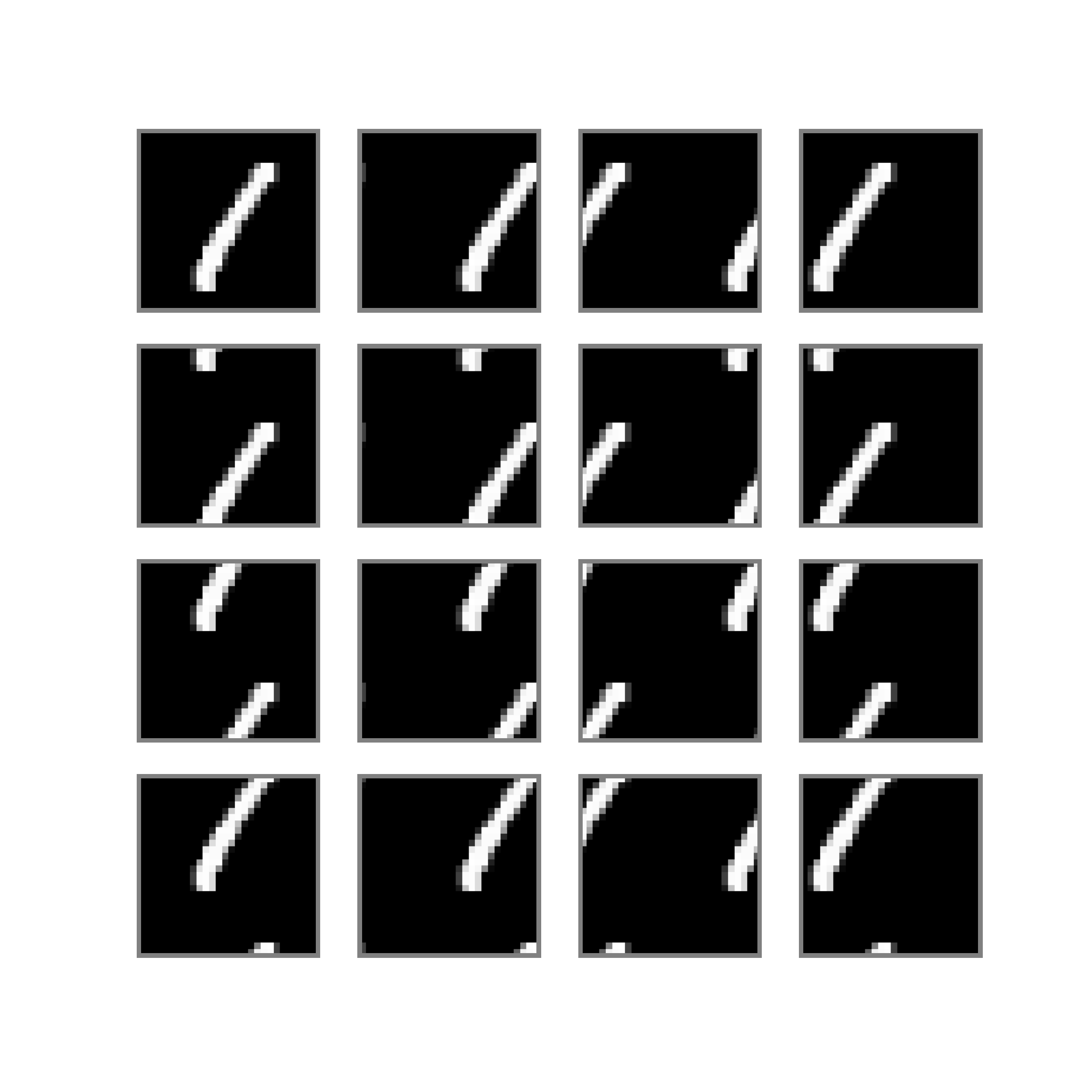}
    \end{subfigure}
    \caption{Examples of translated images of MNIST digits (with \textsf{stride=7}) for testing cyclic translation invariance of the ReduNet. (\textbf{Left}) Diagram for cyclic translation;  (\textbf{Right}) Translated digit `0' and digit `1'.}
    \label{fig:mnist-translation-visualize}
    \vspace{-0.2in}
\end{figure}

\textbf{Rotational Invariance on MNIST Digits.}
We study the ReduNet on learning \textit{rotation} invariant features on MNIST dataset~\citep{lecun1998mnist}. We impose a polar grid on the image $\x\in\R^{H\times W}$, with its geometric center being the center of the 2D polar grid. For each radius $r_i$, $i \in [C]$, we can sample $\Gamma$ pixels with respect to each angle $\gamma_l =l\cdot({2\pi}/\Gamma)$ with $l \in [\Gamma]$. 
Then given an image sample $\x$ from the dataset, we represent the image in a polar coordinate representation $\x(p) = (\gamma_{l,i}, r_{l,i})\in\R^{\Gamma\times C}$. 
Examples of rotated images are shown in Figure~\ref{fig:mnist-rotation-visualize}.
Our goal is to learn rotation invariant features, i.e., we expect to learn $f(\cdot, \bm{\theta})$ such that $\{f(\x(p) \circ \mathfrak{g}, \bm{\theta})\}_{\mathfrak{g} \in\mathbb{G}}$ lie in the same subspace, where $\mathfrak{g}$ is the shift transformation in polar angle.
By performing polar coordinate transformation for images from digit `0' and digit `1' in the training dataset, we can obtain the data matrix $\X(p) \in \mathbb{R}^{(\Gamma\cdot C) \times m}$. 
After performing polar coordinate transformation, the rotation operation in the original images is equivalent to the shift operation in the polar coordinate system. 
We use $m=2,000$ training samples, set $\Gamma=200$ and $C=5$ for polar transformation, and set iteration $L=3,500$, precision $\epsilon=0.1$, step-size $\eta=0.5$. We generate $1,000$ test samples followed by the same procedure. 
In Figure~\ref{fig:appendix-mnist1d-heatmaps}, we can see that our proposed ReduNet is able to map most samples from different classes to orthogonal subspaces (w.r.t. class) on test dataset. Meanwhile, in Figure~\ref{fig:1d-invariance-plots-e}, \ref{fig:1d-invariance-plots-f}, and \ref{fig:1d-invariance-plots-h}, we observe that the learnt features are invariant to shift transformation in polar angle (i.e., arbitrary rotation in $\x$).

We compare the accuracy (both on the original test data and the shifted test data) of the ReduNet (without considering invariance) and the shift invariant ReduNet. 
For ReduNet (without considering invariance), we use the same training dataset as the shift invariant ReduNet, we set iteration $L=3,500$, step size $\eta=0.5$, and precision $\epsilon=0.1$.
The results are summarized in Table~\ref{table:mcr-rotation-translation-mnist}. With the invariant design, we can see from Table~\ref{table:mcr-rotation-translation-mnist} that the shift invariant ReduNet achieves better performance in terms of invariance on the MNIST binary classification task.

\begin{table}[t]
\begin{center}
\caption{\small Comparing network performance on learning rotational-invariant representations and 2D translation-invariant representations on MNIST.}
\label{table:mcr-rotation-translation-mnist}
\begin{small}
\begin{sc}
\begin{tabular}{l | c c }
\toprule
\textbf{Shift Invariance}& ReduNet & ReduNet ({\scriptsize shift-invariant})  \\
\midrule
Acc ({\scriptsize Original Test Data}) & 0.983 & 0.996 \\
Acc ({\scriptsize Test Data with All Possible Shifts}) & 0.707 & 0.993 
\\
\toprule
\textbf{Translational Invariance}& ReduNet & ReduNet ({\scriptsize translation-invariant})  \\
\midrule
Acc ({\scriptsize Original Test Data}) & 0.980 & 0.975 \\
Acc ({\scriptsize Test Data with All Possible Shifts}) & 0.540 & 0.909 \\
\bottomrule
\end{tabular}
\end{sc}
\end{small}
\end{center}
\end{table}

\textbf{Translational Invariance on MNIST Digits.} We provide experimental results for verifying the invariance property of ReduNet under 2D translations.  We construct 1). ReduNet (without considering invariance) and 2). 2D translation-invariant ReduNet for classifying digit `0' and digit `1' on MNIST dataset. We use $m=1,000$ samples (500 samples from each class) for training the models, and use another $500$ samples (250 samples from each class) for evaluation. To evaluate the 2D translational invariance, for each test image $\x_{\text{test}}\in\R^{H\times W}$, we consider {\em all} translation augmentations of the test image with a \textsf{stride=7}. More specifically, for the MNIST dataset, we have $H=W=28$. So for each image, the total number of all cyclic translation augmentations (with \textsf{stride=7}) is $4\times 4=16$. Examples of translated images are shown in Figure~\ref{fig:mnist-translation-visualize}. Notice that such translations are considerably larger than normally considered in the literature since we consider invariance to the entire group of cyclic translations on the $H\times W$ grid as a torus. 

For ReduNet (without considering translation invariance), we set iteration $L=2,000$, step size $\eta=0.1$, and precision $\epsilon=0.1$. For translation-invariant ReduNet, we set $L=2,000$, step size $\eta=0.5$, precision $\epsilon=0.1$, number of channels $C=5$, and kernel size is set as $3\times 3$.
We summarize the results in Table~\ref{table:mcr-rotation-translation-mnist}. Similar to the 1D rotational results on the MNIST dataset, the translation-invariant ReduNet achieves better performance under translations compared with the RedeNet without considering invariance. The accuracy drop of the translation-invariant ReduNet is much less than the one of ReduNet without invariance design.

%% file: appendix.tex
\section{Additional Remarks and Extensions}
\label{app:remarks}

\begin{remark}[Approximate with a ReLU Network]\label{rem:ReLU}
In practice, there are many other simpler nonlinear activation functions that one can use to approximate the membership $\widehat{\bm \pi}(\cdot)$ and subsequently the nonlinear operation $\bm \sigma$ in \eqref{eqn:soft-residual}. Notice that the geometric meaning of $\bm \sigma$ in \eqref{eqn:soft-residual} is to compute the ``residual'' of each feature against the subspace to which it belongs. So when we restrict all our features to be in the first (positive) quadrant of the feature space,\footnote{Most current neural networks seem to adopt this regime.}
one may approximate this residual using the rectified linear units operation, ReLUs, on $\p_{j} = \bm C_{\ell}^j \z_{\ell}$ or its orthogonal complement:
\begin{equation}
\bm \sigma(\z_\ell) \; \propto \; \z_\ell - \sum_{j=1}^k   \mbox{ReLU}\big(\bm P_{\ell}^j \z_{\ell}\big), 
\label{eq:approx-relu}
\end{equation}
where $\bm P_{\ell}^j = (\bm C_{\ell}^j)^\perp$ is the projection onto the $j$-th class\footnote{$\bm P_{\ell}^j$ can be viewed as the orthogonal complement to $\bm C_{\ell}^j$.} and $\text{ReLU}(x) = \max(0, x)$. The above approximation is good under the more restrictive assumption that projection of $\z_{\ell}$ on the correct class via  $\bm P_{\ell}^j$ is mostly large and positive and yet small or negative for other classes. 

The resulting ReduNet will be a network primarily involving ReLU operations and feature normalization (onto $\mathbb S^{n-1})$ between each layer. Although in this work, we have argued that the forward-constructed ReduNet network already works to a large extent, in practice one certainly can conduct back-propagation to further fine tune the so-obtained network, say to correct some remaining errors in predicting labels of the training data. Empirically, people have found that deep networks with ReLU activations are easier to train via back propagation \citep{krizhevsky2012imagenet}. 
\end{remark}

\begin{remark}[Accelerated Optimization via Additional Skip Connections]\label{rem:acceleration}
Empirically, people have found that additional skip connections across multiple layers may improve the network performance, e.g. the DenseNet \citep{dense-net}. In our framework, the role of each layer is precisely interpreted as one iterative gradient ascent step for the objective function $\Delta R$. In our experiments (see Section \ref{sec:experiments}), we have observed that the basic gradient scheme sometimes converges slowly, resulting in deep networks with thousands of layers (iterations)!  To improve the efficiency of the basic ReduNet, one may consider in the future accelerated gradient methods such as the Nesterov acceleration \citep{nesterov1983method} or perturbed accelerated gradient descent \citep{Jin-2018}. Say to minimize or maximize a function $h(\z)$, such accelerated methods usually take the form:
\begin{equation}
\left\{
\begin{array}{ccl}
    \bm p_{\ell +1} &=& \z_\ell + \beta_\ell\cdot (\z_\ell - \z_{\ell-1}),\\
    \z_{\ell+1} &=& \bm p_{\ell+1} + \eta \cdot \nabla h(\bm p_{\ell+1}). 
\end{array}  
\right.
\label{eqn:acceleration}
\end{equation}
Hence they require introducing additional skip connections among three layers $\ell-1$, $\ell$ and $\ell+1$. For typical convex or nonconvex programs, the above accelerated schemes can often reduce the number of iterations by a magnitude. 
\end{remark}

\newpage
\section{1D Circular Shift Invariance}\label{app:1D}
It has been long known that to implement a convolutional neural network, one can achieve higher computational efficiency by implementing the network in the spectral domain via the fast Fourier transform \citep{mathieu2013fast,lavin2015fast,Vasilache2015FastCN}. However, our purpose here is different: We want to show that the linear operators $\bm E$ and $\bm C^j$ derived from the gradient flow of MCR$^2$ are naturally convolutions when we enforce shift-invariance rigorously. Their convolution structure is derived from the rate reduction objective, rather than heuristically imposed upon the network. Furthermore, the computation involved in constructing these linear operators has a naturally efficient implementation in the spectral domain via fast Fourier transform. Arguably this work is the first to show multi-channel convolutions, together with other convolution-preserving nonlinear operations in the ReduNet, are both necessary and sufficient to ensure shift invariance.

To be somewhat self-contained and self-consistent, in this section, we first introduce our notation and review some of the key properties of circulant matrices which will be used to characterize the properties of the linear operators $\bm E$ and $\bm C^j$ and to compute them efficiently. The reader may refer to \cite{Kra2012OnCM} for a more rigorous exposition on circulant matrices.

\subsection{Properties of Circulant Matrix and Circular Convolution}\label{ap:circulant}

Given a vector $\z = [z_0, z_1, \ldots, z_{n-1}]^* \in \Re^n$,\footnote{We use superscript $^*$ to indicate (conjugate) transpose of a vector or a matrix} we may arrange all its circular shifted versions in a circulant matrix form as
\begin{equation}\label{eq:def-circulant}
\circm(\z) \quad\doteq\quad \left[ \begin{array}{ccccc} z_0 & z_{n-1} & \dots & z_2 & z_1 \\ z_1 & z_0 & z_{n-1} & \cdots & z_2 \\ \vdots & z_1 & z_0 &\ddots & \vdots \\ z_{n-2} &  \vdots & \ddots & \ddots & z_{n-1} \\ z_{n-1} & z_{n-2} & \dots & z_1 & z_0   \end{array} \right] \quad \in \Re^{n \times n}.
\end{equation}
\begin{fact}[Convolution as matrix multiplication via circulant matrix] The multiplication of a circulant matrix $\circm(\z)$ with a vector $\x \in \Re^n$ gives a circular (or cyclic) convolution, i.e., 
    \begin{equation}
        \circm(\z) \cdot \x = \z \circledast \x,
    \end{equation} 
    where
    \begin{equation}\label{eq:def-convolution}
    (\bm z \circledast \bm x)_{i} = \sum_{j=0}^{n-1} x_j z_{i+ n-j \, \mathrm{mod} \,n}.
    \end{equation}
\end{fact}

\begin{fact}[Properties of circulant matrices]
\label{fact:circ-properties}
Circulant matrices have the following properties:
\begin{itemize}
    \item Transpose of a circulant matrix, say $\circm(\z)^*$, is circulant;
    \item Multiplication of two circulant matrices is circulant, for example $\circm(\z)\circm(z)^*$;
    \item For a non-singular circulant matrix,  its inverse is also circulant (hence representing a circular convolution). 
\end{itemize}
\end{fact}

These properties of circulant matrices are extensively used in this work as for characterizing the convolution structures of the operators $\E$ and $\C^j$.

Given a set of vectors $[\z^{1}, \dots, \z^{m}] \in \Re^{n\times m}$, let $\circm(\z^i) \in \Re^{n\times n}$ be the circulant matrix for $\z^i$. 
Then we have the following:
\begin{proposition}[Convolution structures of $\bm E$ and $\bm C^j$]
Given a set of vectors $\bm Z =[\z^1, \ldots, \z^m]$, the matrix: 
$$\E = \alpha\big(\bm I + \alpha\sum_{i=1}^m \circm(\z^i)\circm(\z^i)^*\big)^{-1}
$$ 
is a circulant matrix and represents a circular convolution: $$\E \z = \bm e \circledast \z,$$ where $\bm e$ is the first column vector of $\E$. Similarly, the matrices $\bm C^j$ associated with any subsets of $\bm Z$ are also circular convolutions. 
\label{prop:circular-conv}
\end{proposition}

\subsection{Circulant Matrix and Circulant Convolution for Multi-channel Signals}\label{ap:multichannel-circulant}
In the remainder of this section, we view $\z$ as a 1D signal such as an audio signal. Since we will deal with the more general case of multi-channel signals, we will use the traditional notation $T$ to denote the temporal length of the signal and $C$ for the number of channels. Conceptually, the ``dimension'' $n$ of such a multi-channel signal, if viewed as a vector, should be $n = CT$.\footnote{Notice that in the main paper, for simplicity, we have used $n$ to indicate both the 1D ``temporal'' or 2D ``spatial'' dimension of a signal, just to be consistent with the vector case, which corresponds to $T$ here. All notation should be clear within the context.} As we will also reveal additional interesting structures of the operators $\bm E$ and $\bm C^j$ in the spectral domain, we use $t$ as the index for time, $p$ for the index of frequency, and $c$ for the index of channel.

Given a multi-channel 1D signal  $\bar\z \in \Re^{C \times T}$, we denote
\begin{equation}\label{eq:index-multichannel}
    \bar\z = 
        \begin{bmatrix}
        \bar\z[1]^*\\
        \vdots\\
        \bar\z[C]^*\\
    \end{bmatrix}
    = [\bar\z(0), \bar\z(1), \ldots, \bar\z(T-1)] = \{\bar\z[c](t)\}_{c=1, t=0}^{c=C, t=T-1}.
\end{equation}
To compute the coding rate reduction for a collection of such multi-channel 1D signals, we may flatten the matrix representation into a vector representation by stacking the multiple channels of $\bar\z$ as a column vector.  
In particular, we let
\begin{equation}
    \vec(\bar\z) = \left[ \bar\z[1](0), \bar\z[1](1), \ldots, \bar\z[1](T-1), \bar\z[2](0), \ldots \right] \quad \in \Re^{(C\times T)}.
\end{equation}
Furthermore, to obtain shift invariance for the coding rate reduction, we may generate a collection of shifted copies of $\bar\z$ (along the temporal dimension). 
Stacking the vector representations for such shifted copies as column vectors, we obtain
\begin{equation}
    \circm(\bar\z) \doteq 
    \begin{bmatrix}
    \circm(\bar\z[1])\\
    \vdots\\
    \circm(\bar\z[C])
    \end{bmatrix} \quad
    \in \Re^{(C\times T) \times T}.
\end{equation}
In above, we overload the notation ``$\circm(\cdot)$'' defined in \eqref{eq:def-circulant}. 

We now consider a collection of $m$ multi-channel 1D signals  $\{\bar\z^i \in \Re^{C \times T}\}_{i=1}^m$.
Compactly representing the data by $\bar\Z \in \Re^{C \times T \times m}$ in which the $i$-th slice on the last dimension is $\bar\z^i$, we denote
\begin{equation}\label{eq:index-multichannel-collection}
    \bar\Z[c] = [\bar\z^1[c], \ldots, \bar\z^m[c]] \in \Re^{T \times m}, \qquad \bar\Z(t) = [\bar\z^1(t), \ldots, \bar\z^m(t)] \in \Re^{C \times m}.
\end{equation}
In addition, we denote
\begin{equation}
\begin{split}
    \vec(\bar\Z) &= [\vec(\bar\z^1), \ldots, \vec(\bar\z^m)] \in \Re^{(C\times T) \times m}, \\
    \circm(\bar\Z) &= [\circm(\bar\z^1), \ldots, \circm(\bar\z^m)] \in \Re^{(C\times T) \times (T \times m)}.
\end{split}
\end{equation}
Then, we define the \emph{shift invariant coding rate reduction} for $\bar\Z \in \Re^{C \times T \times m}$ as
\begin{multline}
\label{eq:1D-MCR2}
    \Delta R_\circm(\bar\Z, \bm{\Pi}) \doteq \frac{1}{T}\Delta R(\circm(\bar\Z), \bar{\bm{\Pi}}) \\= 
    \frac{1}{2T}\log\det \Bigg(\I + \alpha \cdot \circm(\bar\Z) \cdot \circm(\bar\Z)^{*} \Bigg) 
    - \sum_{j=1}^{k}\frac{\gamma_j}{2T}\log\det\Bigg(\I + \alpha_j \cdot \circm(\bar\Z) \cdot \bar{\bm{\Pi}}^{j} \cdot \circm(\bar\Z)^{*} \Bigg),
\end{multline}
where $\alpha = \frac{CT}{mT\epsilon^{2}} = \frac{C}{m\epsilon^{2}}$, $\alpha_j = \frac{CT}{\textsf{tr}\left(\bm{\Pi}^{j}\right)T\epsilon^{2}} = \frac{C}{\textsf{tr}\left(\bm{\Pi}^{j}\right)\epsilon^{2}}$, $\gamma_j = \frac{\textsf{tr}\left(\bm{\Pi}^{j}\right)}{m}$, and $\bar{\bm \Pi}^j$ is augmented membership matrix in an obvious way.
Note that we introduce the normalization factor $T$ in \eqref{eq:1D-MCR2} because the circulant matrix $\circm(\bar\Z)$ contains $T$ (shifted) copies of each signal.

By applying \eqref{eqn:expand-directions} and \eqref{eqn:compress-directions}, we obtain the derivative of $\Delta R_\circm(\bar\Z, \bm{\Pi})$ as
\begin{equation}\label{eqn:expand-directions-multichannel} 
\begin{split}
    \frac{1}{2T}\frac{\partial \log \det \Big(\I + \alpha \circm(\bar\Z) \circm(\bar\Z)^{*} \Big)}{\partial  \vec(\bar\Z)} 
    &= \frac{1}{2T}\frac{\partial \log \det \Big(\I + \alpha \circm(\bar\Z) \circm(\bar\Z)^{*} \Big)}{\partial \circm(\bar\Z)} \frac{\partial \circm(\bar\Z)}{\partial  \vec(\bar\Z)}\\
    &= \underbrace{\alpha\Big(\I + \alpha\circm(\bar\Z) \circm(\bar\Z)^{*}\Big)^{-1}}_{\bar\E{} \; \in \Re^{(C\times T)\times (C \times T)}}\vec(\bar\Z), 
\end{split}
\end{equation}    
\begin{equation}\label{eqn:compress-directions-multichannel}
    \frac{\gamma_j }{2T}\frac{\partial \log \det \Big(\I + \alpha_j \circm(\bar\Z) \bm \Pi^j \circm(\bar\Z)^{*} \Big)}{\partial  \vec(\bar\Z)} = \gamma_j  \underbrace{ \alpha_j  \Big(\I +  \alpha_j \circm(\bar\Z) \bm \Pi^j \circm(\bar\Z)^{*}\Big)^{-1}}_{\bar\C^j \; \in \Re^{(C\times T)\times (C \times T)}} \vec(\bar\Z) \bm \Pi^j.
\end{equation}

In the following, we show that $\bar\E \cdot \vec(\bar\z)$ represents a multi-channel circular convolution. 
Note that
\begin{equation}\label{eq:calc-E-bar}
    \bar{\bm E} =
    \alpha 
    \left[\begin{smallmatrix}
    \bm I + \alpha \sum_{i=1}^m\circm(\z^i[1])\circm(\z^i[1])^* & \cdots & \sum_{i=1}^m\circm(\z^i[1])\circm(\z^i[C])^* \\
    \vdots & \ddots & \vdots \\
    \sum_{i=1}^m\circm(\z^i[C])\circm(\z^i[1])^* & \cdots & \bm I + \sum_{i=1}^m\alpha \circm(\z^i[C])\circm(\z^i[C])^* \\
    \end{smallmatrix}\right]^{-1}.
\end{equation}
By using Fact~\ref{fact:circ-properties}, the matrix in the inverse above is a \emph{block circulant matrix}, i.e., a block matrix where each block is a circulant matrix. 
A useful fact about the inverse of such a matrix is the following. 
\begin{fact}[Inverse of block circulant matrices]
The inverse of a block circulant matrix is a block circulant matrix (with respect to the same block partition).
\end{fact}

The main result of this subsection is the following.
\begin{proposition}[Convolution structures of $\bar{\bm E}$ and $\bar{\bm C}^j$]
Given a collection of multi-channel 1D signals  $\{\bar\z^i \in \Re^{C \times T}\}_{i=1}^m$, the matrix $\bar\E$
is a block circulant matrix, i.e.,
\begin{equation}\label{eq:E-bar}
    \bar{\bm E} \doteq 
    \begin{bmatrix}
        \bar{\bm E}_{1, 1} & \cdots & \bar{\bm E}_{1, C}\\
        \vdots & \ddots & \vdots \\
        \bar{\bm E}_{C, 1} & \cdots & \bar{\bm E}_{C, C}\\
    \end{bmatrix},
\end{equation}
where each $\bar{\bm E}_{c, c'}\in \Re^{T \times T}$ is a circulant matrix. Moreover, $\bar{\bm E}$ represents a multi-channel circular convolution, i.e., for any multi-channel signal $\bar\z \in \Re^{C \times T}$ we have 
$$\bar\E \cdot \vec(\bar\z) = \vec( \bar{\bm e} \circledast \bar\z).$$ 
In above, $\bar{\bm e} \in \Re^{C \times C \times T}$ is a multi-channel convolutional kernel  with $\bar{\bm e}[c, c'] \in \Re^{T}$ being the first column vector of $\bar{\bm E}_{c, c'}$, and $\bar{\bm e} \circledast \bar\z \in \Re^{C \times T}$ is the multi-channel circular convolution (with ``$\circledast$'' overloading the notation from Eq.  \eqref{eq:def-convolution}) defined as
\begin{equation}
    (\bar{\bm e} \circledast \bar\z)[c] = \sum_{c'=1}^C \bar{\bm e}[c, c'] \circledast \bar{\z}[c'], \quad \forall c = 1, \ldots, C.
\end{equation}
Similarly, the matrices $\bar{\bm C}^j$ associated with any subsets of $\bar{\bm Z}$ are also multi-channel circular convolutions. 
\label{prop:multichannel-circular-conv}
\end{proposition}

Note that the calculation of $\bar\E$ in \eqref{eq:calc-E-bar} requires inverting a matrix of size $(C\times T) \times (C \times T)$. 
In the following, we show that this computation can be accelerated by working in the frequency domain.

\subsection{Fast Computation in Spectral Domain}
\label{ap:1D-shift}

\paragraph{Circulant matrix and Discrete Fourier Transform.}
A remarkable property of circulant matrices is that {\em they all share the same set of eigenvectors that form a unitary matrix}. 
We define the matrix:
\begin{equation}\label{eq:dft-matrix}
    \F_T \doteq \frac{1}{\sqrt{T}} 
    \begin{bmatrix}
        \omega_T^0 & \omega_T^0 & \cdots & \omega_T^0 &  \omega_T^0\\
        \omega_T^0 & \omega_T^1 & \cdots & \omega_T^{T-2} &  \omega_T^{T-1}\\
        \vdots     & \vdots     & \ddots     & \vdots & \vdots \\
        \omega_T^0 & \omega_T^{ T-2}& \cdots & \omega_T^{(T-2)^2} &  \omega_T^{(T-2)(T-1)} \\
        \omega_T^0 & \omega_T^{T-1}& \cdots & \omega_T^{(T-2)(T-1)} &  \omega_T^{(T-1)^2}
    \end{bmatrix} \quad \in \mathbb{C}^{T\times T},
\end{equation} 
where $\omega_T \doteq \exp(- \frac{2\pi\sqrt{-1}}{T})$ is the roots of unit (as $\omega_T^T = 1$). The matrix $\F_T$ is a unitary matrix: $\F_T \F_T^* = \bm I$ and is the well known {\em Vandermonde matrix}. Multiplying a vector with $\F_T$ is known as the {\em discrete Fourier transform} (DFT). Be aware that the conventional DFT matrix differs from our definition of $\bm F_T$ here by a scale: it does not have the $\frac{1}{\sqrt{T}}$ in front. Here for simplicity, we scale it so that $\bm F_T$ is a unitary matrix and its inverse is simply its conjugate transpose $\bm F_T^*$, columns of which represent the eigenvectors of a circulant matrix \citep{abidi2016optimization}. 

\begin{fact}[DFT as matrix-vector multiplication]\label{fact:dft}
The DFT of a vector $\z \in \Re^T$ can be computed as
        \begin{equation}
            \dft(\z) \doteq \F_T \cdot \z \quad \in \Co^T,
        \end{equation}
where
\begin{equation}
    \dft(\z)(p) = \frac{1}{\sqrt{T}} \sum_{t=0}^{T-1} z(t) \cdot \omega_T ^{p \cdot t}, \quad \forall p = 0, 1, \ldots, T-1.
\end{equation}
The Inverse Discrete Fourier Transform (IDFT) of a signal $\bv \in \Co^{T}$ can be computed as
        \begin{equation}
            \idft(\bv) \doteq \F_T^* \cdot \bv \quad \in \Co^T
        \end{equation}
        where
\begin{equation}
    \idft(\bv)(t) = \frac{1}{\sqrt{T}} \sum_{p=0}^{T-1} v(p) \cdot \omega_T ^{-p \cdot t}, \quad \forall t = 0, 1, \ldots, T-1.
\end{equation}
\end{fact}

Regarding the relationship between a circulant matrix (convolution) and discrete Fourier transform, we have:
\begin{fact}
An $n\times n$ matrix $\bm M \in \mathbb{C}^{n\times n}$ is a circulant matrix if and only if it is diagonalizable by the unitary matrix $\bm F_n$:
\begin{equation}
   \F_n \bm M \F_n^* = \D \quad \mbox{or} \quad \bm M = \F_n^* \D \F_n,
\end{equation}
where $\D$ is a diagonal matrix of eigenvalues. 
\label{fact:circulant}
\end{fact}

\begin{fact}[DFT are eigenvalues of the circulant matrix] Given a vector $\z \in \Co^T$, we have
    \begin{equation}\label{eq:dft-circulant}
        \F_T \cdot \circm(\z) \cdot \F_T^* = \diag(\dft(\z))  \quad \mbox{or} \quad \circm(\z) = \F_T^* \cdot \diag(\dft(\z)) \cdot \F_T.
    \end{equation}
That is, the eigenvalues of the circulant matrix associated with a vector are given by its DFT.
\label{fact:dft-circulant}
\end{fact}

\begin{fact}[Parseval's theorem]
Given any $\z \in \Co^T$, we have $\|\z\|_2 = \|\dft(\z)\|_2$. More precisely, 
\begin{equation}
    \sum_{t = 0}^{T-1} |\z[t]| ^2 =  \sum_{p = 0}^{T-1} |\dft(\z)[p]| ^2.
\end{equation}
\label{fact:parseval}
\end{fact}
This property allows us to easily ``normalize'' features after each layer onto the sphere $\mathbb S^{n-1}$ directly in the spectral domain (see Eq. \eqref{eqn:layer-approximate} and \eqref{eqn:layer-approximate-spectral}).

\paragraph{Circulant matrix and Discrete Fourier Transform for multi-channel signals.}

We now consider multi-channel 1D signals $\bar\z \in \Re^{C \times T}$. 
Let $\dft(\bar\z) \in \Co^{C \times T}$ be a matrix where the $c$-th row is the DFT of the corresponding signal $\z[c]$, i.e., 
\begin{equation}
    \dft(\bar\z) \doteq 
    \begin{bmatrix}
    \dft(\z[1]) ^*\\
    \vdots\\
    \dft(\z[C]) ^*
    \end{bmatrix} \quad
    \in \Co^{C \times T}.
\end{equation}
Similar to the notation in \eqref{eq:index-multichannel}, we denote
\begin{multline}
    \dft(\bar\z) = 
    \left[\begin{smallmatrix}
        \dft(\bar\z)[1]^*\\
        \vdots\\
        \dft(\bar\z)[C]^*\\
    \end{smallmatrix}\right]
    = [\dft(\bar\z)(0), \dft(\bar\z)(1), \ldots, \dft(\bar\z)(T-1)] \\
    = \{\dft(\bar\z)[c](t)\}_{c=1, t=0}^{c=C, t=T-1}.
\end{multline}
As such, we have $\dft(\z[c]) = \dft(\bar\z)[c]$. 

By using Fact~\ref{fact:dft-circulant}, $\circm(\bar\z)$ and $\dft(\bar\z)$ are related as follows:
\begin{equation}\label{eq:dft-circulant-multichannel}
    \circm(\bar\z) = 
    \begin{bmatrix}
    \F_T^*\cdot  \diag(\dft(\z[1])) \cdot \F_T\\
    \vdots\\
    \F_T^*\cdot  \diag(\dft(\z[C])) \cdot \F_T
    \end{bmatrix}
    =
    \begin{bmatrix}
        \F_T^* &  \cdots & \0 \\
        \vdots& \ddots & \vdots \\
        \0   &  \cdots & \F_T^* \\
    \end{bmatrix}
    \cdot
    \begin{bmatrix}
     \diag(\dft(\z[1]))\\
    \vdots\\
     \diag(\dft(\z[C]))
    \end{bmatrix}
     \cdot \F_T.
\end{equation}
We now explain how this relationship can be leveraged to produce a fast computation of $\bar\E$ defined in \eqref{eqn:expand-directions-multichannel}. 
First, there exists a permutation matrix $\P$ such that
\begin{equation}\label{eq:permute-dft}
    \begin{bmatrix}
        \diag(\dft(\z[1]))  \\
        \diag(\dft(\z[2]))  \\
        \vdots \\
        \diag(\dft(\z[C])) \\
    \end{bmatrix}
    = \P \cdot
    \begin{bmatrix}
        \dft(\bar\z)(0) & \0         & \cdots & 0\\
        \0         & \dft(\bar\z)(1) & \cdots & 0\\
        \vdots     & \vdots     & \ddots & \vdots \\
        \0         & \0         & \cdots & \dft(\bar\z)(T-1)
    \end{bmatrix}.
\end{equation}
Combining \eqref{eq:dft-circulant-multichannel} and \eqref{eq:permute-dft}, we have
\begin{equation}\label{eq:multi-channel-circ-covariance}
    \circm(\bar\z) \cdot \circm(\bar\z)^* = 
    \begin{bmatrix}
        \F_T^* &  \cdots & \0 \\
        \vdots& \ddots & \vdots \\
        \0   &  \cdots & \F_T^* \\
    \end{bmatrix}
    \cdot \P \cdot \D(\bar\z) \cdot \P^* \cdot
    \begin{bmatrix}
        \F_T &  \cdots & \0 \\
        \vdots& \ddots & \vdots \\
        \0   &  \cdots & \F_T \\
    \end{bmatrix},
\end{equation}
where
\begin{equation}
    \D(\bar\z) \doteq 
    \begin{bmatrix}
        \dft(\bar\z)(0) \cdot \dft(\bar\z)(0)^*  & \cdots & \0\\
         \vdots                              & \ddots & \vdots \\
        \0         & \cdots                 & \dft(\bar\z)(T-1) \cdot \dft(\bar\z)(T-1)^*
    \end{bmatrix}.
\end{equation}

Now, consider a collection of $m$ multi-channel 1D signals $\bar\Z \in \Re^{C \times T \times m}$. 
Similar to the notation in \eqref{eq:index-multichannel-collection}, we denote
\begin{equation}
\begin{split}
    \dft(\bar\Z)[c] &= [\dft(\bar\z^1)[c], \ldots, \dft(\bar\z^m)[c]] \in \Co^{T \times m}, \\
    \dft(\bar\Z)(p) &= [\dft(\bar\z^1)(p), \ldots, \dft(\bar\z^m)(p)] \in \Co^{C \times m}.
\end{split}
\end{equation}

By using \eqref{eq:multi-channel-circ-covariance}, we have
\begin{multline}\label{eq:bar-E-frequency}
    \bar\E =  
    \begin{bmatrix}
        \F_T^* &  \cdots & \0 \\
        \vdots& \ddots & \vdots \\
        \0   &  \cdots & \F_T^* \\
    \end{bmatrix}
    \cdot \P \cdot
    \alpha \cdot \left[\bm I + \alpha \cdot \sum_{i=1}^m \D(\bar\z^i)\right]^{-1} 
    \cdot \P^* \cdot
    \begin{bmatrix}
        \F_T &  \cdots & \0 \\
        \vdots& \ddots & \vdots \\
        \0   &  \cdots & \F_T \\
    \end{bmatrix}.
\end{multline}
Note that $\alpha \cdot \left[\bm I + \alpha \cdot \sum_{i=1}^m \D(\bar\z^i)\right]^{-1}$ is equal to
\begin{multline}\label{eq:E-bar-fast-inverse}
    \alpha
    \left[\begin{smallmatrix}
        \bm I + \alpha \dft(\bar\Z)(0) \cdot \dft(\bar\Z^i)(0)^*  & \cdots & \0\\
         \vdots                              & \ddots & \vdots \\
        \0         & \cdots                 & \bm I + \alpha \dft(\bar\Z)(T-1) \cdot \dft(\bar\Z)(T-1)^*
    \end{smallmatrix}\right] ^{-1} \\
    =
    \left[\begin{smallmatrix}
        \alpha \left(\bm I + \alpha \dft(\bar\Z)(0) \cdot \dft(\bar\Z)(0)^*\right)^{-1}  & \cdots & \0\\
         \vdots                              & \ddots & \vdots \\
        \0         & \cdots                 & \alpha \left(\bm I + \alpha \dft(\bar\Z)(T-1) \cdot \dft(\bar\Z)(T-1)^*\right)^{-1}
    \end{smallmatrix}\right].   
\end{multline}
Therefore, the calculation of $\bar\E$ only requires inverting $T$ matrices of size $C\times C$. 
This motivates us to construct the ReduNet in the spectral domain for the purpose of accelerating the computation, as we explain next.

\paragraph{Shift-invariant ReduNet in the Spectral  Domain.}

Motivated by the result in \eqref{eq:E-bar-fast-inverse}, we introduce the notations $\bar\cE \in \Re^{C \times C \times T}$ and $\bar\cC^j \in \Re^{C \times C \times T}$ given by
\begin{eqnarray}
    \bar\cE(p) &\doteq& \alpha \cdot \left[\I + \alpha \cdot \dft(\bar\Z)(p) \cdot \dft(\bar\Z)(p)^* \right]^{-1} \quad \in \Co^{C\times C}, \\
    \bar\cC^j(p) &\doteq& \alpha_j \cdot\left[\I + \alpha_j \cdot \dft(\bar\Z)(p) \cdot \bm{\Pi}_j \cdot \dft(\bar\Z)(p)^*\right]^{-1} \quad \in \Co^{C\times C}.
\end{eqnarray}
In above, $\bar\cE(p)$ (resp., $\bar\cC^j(p)$) is the $p$-th slice of $\bar\cE$ (resp., $\bar\cC^j$) on the last dimension. 
Then, the gradient of $\Delta R_\circm(\bar\Z, \bm{\Pi})$ with respect to $\bar\Z$ can be calculated by the following result.

\begin{theorem} [Computing multi-channel convolutions $\bar{\bm E}$ and $\bar{\bm C}^j$]
\label{thm:1D-convolution}
Let $\bar\U \in \Co^{C \times T \times m}$ and $\bar\W^{j} \in \Co^{C \times T \times m}, j=1,\ldots, k$ be given by 
\begin{eqnarray}
    \bar\U(p) &\doteq& \bar\cE(p) \cdot \dft(\bar\Z)(p), \\
    \bar\W^{j}(p) &\doteq& \bar\cC^j(p) \cdot \dft(\bar\Z)(p), \quad j=1,\ldots, k,
\end{eqnarray}
for each  $p \in \{0, \ldots, T-1\}$. Then, we have
\begin{eqnarray}
    \frac{1}{2T}\frac{\partial \log \det (\I + \alpha \cdot \circm(\bar\Z) \circm(\bar\Z)^{*} )}{\partial \bar\Z} &=& \idft(\bar\U), \\
    \frac{\gamma_j}{2T}\frac{\partial  \log\det (\I + \alpha_j \cdot \circm(\bar\Z) \bm \bar{\bm \Pi}^j \circm(\bar\Z)^{*})}{\partial \bar\Z}&=&
    \gamma_j \cdot \idft(\bar\W^{j} \bm \Pi^j).
\end{eqnarray}
\end{theorem}
By this result, the gradient ascent update in \eqref{eqn:gradient-descent} (when applied to $\Delta R_\circm(\bar\Z, \bm{\Pi})$) can be equivalently expressed as an update in frequency domain on $\bar\V_\ell \doteq \dft(\bar\Z_\ell)$ as
\begin{equation}
    \bar\V_{\ell+1}(p) \; \propto \; \bar\V_{\ell}(p) + \eta \; \Big(\bar\cE_\ell(p) \cdot \bar\V_\ell(p) - \sum_{j=1}^k \gamma_j \bar\cC_\ell^j(p) \cdot \bar\V_\ell(p)\Pi^j \Big), \quad p = 0, \ldots, T-1.
\end{equation}
Similarly, the gradient-guided feature map increment in \eqref{eqn:layer-approximate} can be equivalently expressed as an update in frequency domain on $\bar\bv_\ell \doteq \dft(\bar\z_\ell)$ as
\begin{equation}
\bar\bv_{\ell+1}(p) \propto \bar\bv_\ell(p) +  \eta \cdot  \bar\cE_{\ell}(p) \bar\bv_{\ell}(p) - \eta\cdot  \bm \sigma\Big([\bar\cC_{\ell}^{1}(p) \bar\bv_{\ell}(p), \dots, \bar\cC_{\ell}^{k}(p) \bar\bv_{\ell}(p)]\Big), \quad p = 0, \ldots, T-1,
\label{eqn:layer-approximate-spectral}
\end{equation}
subject to the constraint that $\|\bar\bv_{\ell+1}\|_F = \|\bar\z_{\ell+1}\|_F = 1$ (the first equality follows from Fact~\ref{fact:parseval}). 

We summarize the training, or construction to be more precise, of ReduNet in the spectral domain in Algorithm~\ref{alg:training-1D}.

\begin{algorithm}[t]
	\caption{\textbf{Training Algorithm} (1D Signal, Shift Invariance, Spectral Domain)}
	\label{alg:training-1D}
	\begin{algorithmic}[1]
		\REQUIRE $\bar\Z \in \Re^{C \times T \times m}$, $\bm{\Pi}$, $\epsilon > 0 $, $\lambda$, and a learning rate $\eta$.
		\STATE Set $\alpha = \frac{C}{m \epsilon^2}$, $\{\alpha_j = \frac{C}{\textsf{tr}\left(\bm{\Pi}^{j}\right)\epsilon^{2}}\}_{j=1}^k$, 
		$\{\gamma_j = \frac{\textsf{tr}\left(\bm{\Pi}^{j}\right)}{m}\}_{j=1}^k$.
		\STATE Set $\bar\V_0 = \{\bar\bv_0^{i}(p) \in \Co^C\}_{p=0,i=1}^{T-1, m}\doteq \dft(\bar\Z) \in \Co^{C \times T \times m}$.
		\FOR{$\ell = 1, 2, \dots, L$} 
		    \STATE \emph{\# Step 1: Compute $\cE$ and $\cC$.}
    		\FOR{$p = 0, 1, \dots, T-1$} 
        		\STATE Compute $\bar\cE_{\ell}(p)\in \Co^{C \times C}$ and $\{\bar\cC_{\ell}^j(p)\in \Co^{C \times C}\}_{j=1}^{k}$ as\\
        		$\bar\cE_\ell(p) \doteq \alpha \cdot \left[\I + \alpha \cdot \bar\V_{\ell-1}(p) \cdot \bar\V_{\ell-1}(p)^* \right]^{-1}$, \\
        		$\bar\cC_{\ell}^j(p) \doteq \alpha_j \cdot\left[\I + \alpha_j \cdot \bar\V_{\ell-1}(p) \cdot \bm{\Pi}^j \cdot \bar\V_{\ell-1}(p)^*\right]^{-1}$;
    		\ENDFOR
    		\STATE \emph{\# Step 2: Update $\bar\bv^i$ for each $i$.}
    		\FOR{$i=1, \ldots, m$}
    		    \STATE \emph{\# Compute projection at each frequency $p$.}
        		\FOR{$p = 0, 1, \dots, T-1$}     
    		        \STATE Compute $\{\bar\p_\ell^{ij} (p) \doteq \bar\cC^j_{\ell}(p) \cdot \bar\bv_\ell^{i}(p) \in \Co^{C \times 1}\}_{j=1}^k$;
        		\ENDFOR    	
    		    \STATE \emph{\# Compute overall projection by aggregating over frequency $p$.}
        		\STATE Let $\{\bar\P_\ell^{ij} = [\bar\p_\ell^{ij} (0), \ldots, \bar\p_\ell^{ij} (T-1)] \in \Co^{C \times T}\}_{j=1}^k$;
    		    \STATE \emph{\# Compute soft  assignment from projection.}
            	\STATE Compute $\Big\{\widehat{\bm \pi}_\ell^{ij} = \frac{\exp(-\lambda \|\bar\P_\ell^{ij}\|_F)}{\sum_{j=1}^k \exp(-\lambda \|\bar\P_\ell^{ij}\|_F)}\Big\}_{j=1}^k$;
    		    \STATE \emph{\# Compute update at each frequency $p$.}     	
        		\FOR{$p = 0, 1, \dots, T-1$}         		
            		\STATE $\bar\bv_\ell^{i}(p) = \bar\bv_{\ell-1}^{i}(p) + \eta \left(\bar\cE_\ell(p) \bar\bv_\ell^{i}(p) - \sum_{j=1}^k \gamma_j \cdot \widehat{\bm \pi}_\ell^{ij} \cdot \bar\cC_\ell^j(p) \cdot \bar\bv_\ell^{i}(p)\right)$;
        		\ENDFOR
        		\STATE Normalize $\bar\bv_\ell^{i} = \bar\bv_\ell^{i} \;/\; \|\bar\bv_\ell^{i}\|_F$;
        	\ENDFOR
    		\STATE Set $\bar\Z_\ell = \idft(\bar\V_\ell)$ as the feature at the $\ell$-th layer;	
    		\STATE \emph{\# Evaluate the objective value.}
    		\STATE $\frac{1}{2T}\sum_{p=0}^{T-1} \left( \log\det[\I + \alpha \bar\V_\ell(p) \cdot \bar\V_\ell(p)^* ] - \frac{\textsf{tr}\left(\bm{\Pi}^{j}\right)}{m}\log\det[\I + \alpha_j \bar\V_\ell(p) \cdot {\bm{\Pi}}^{j} \cdot \bar\V_\ell(p)^* ]\right)$;
		\ENDFOR
		\ENSURE features $\bar\Z_{L}$, the learned filters $\{\bar\cE_\ell(p)\}_{\ell, p}$ and  $\{\bar\cC^j_\ell(p)\}_{j, \ell, p}$.
	\end{algorithmic}
\end{algorithm}

\begin{proof}[Proof to Theorem \eqref{thm:1D-convolution}]

From \eqref{eqn:expand-directions}, \eqref{eq:bar-E-frequency} and \eqref{eq:dft-circulant-multichannel}, we have 
\begin{gather}
    \frac{1}{2}\frac{\partial \log\det \Big(\I + \alpha \circm(\bar\Z)  \circm(\bar\Z)^{*} \Big)}{\partial\circm(\bar\z^i)}
    = 
    \bar\E  \circm(\bar\z^i)
    =  
    \bar\E  
    \left[\begin{smallmatrix}
        \F_T^* &  \cdots & \0 \\
        \vdots& \ddots & \vdots \\
        \0   &  \cdots & \F_T^* \\
    \end{smallmatrix}\right]
    \left[\begin{smallmatrix}
     \diag(\dft(\z^i[1]))\\
    \vdots\\
     \diag(\dft(\z^i[C]))
    \end{smallmatrix}\right]
    \F_T\\
    =
    \left[\begin{smallmatrix}
        \F_T^* &  \cdots & \0 \\
        \vdots& \ddots & \vdots \\
        \0   &  \cdots & \F_T^* \\
    \end{smallmatrix}\right]
    \cdot \P \cdot
    \alpha \cdot \left[\bm I + \alpha \cdot \sum_{i} \D(\bar\z^i)\right]^{-1} \cdot 
    \left[\begin{smallmatrix}
        \dft(\bar\z^i)(0)   & \cdots & \0\\
        \vdots          & \ddots & \vdots \\
        \0              & \cdots & \dft(\bar\z^i)(T-1)
    \end{smallmatrix}\right]
    \cdot \F_T \\
    = 
    \left[\begin{smallmatrix}
        \F_T^* &  \cdots & \0 \\
        \vdots& \ddots & \vdots \\
        \0   &  \cdots & \F_T^* \\
    \end{smallmatrix}\right]
    \cdot \P  \cdot 
    \left[\begin{smallmatrix}
        \bar\cE(0) \cdot \dft(\bar\z^i)(0)   & \cdots & \0\\
        \vdots          & \ddots & \vdots \\
        \0              & \cdots & \bar\cE(T-1) \cdot \dft(\bar\z^i)(T-1)
    \end{smallmatrix}\right]
    \cdot \F_T\\
    =
    \left[\begin{smallmatrix}
        \F_T^* &  \cdots & \0 \\
        \vdots& \ddots & \vdots \\
        \0   &  \cdots & \F_T^* \\
    \end{smallmatrix}\right]
    \cdot \P  \cdot   
    \left[\begin{smallmatrix}
        \bar\u^i(0)   & \cdots & \0\\
        \vdots          & \ddots & \vdots \\
        \0              & \cdots & \bar\u^i(T-1)
    \end{smallmatrix}\right]
    \cdot \F_T
    =
    \left[\begin{smallmatrix}
        \F_T^* &  \cdots & \0 \\
        \vdots& \ddots & \vdots \\
        \0   &  \cdots & \F_T^* \\
    \end{smallmatrix}\right]
    \cdot     
    \left[\begin{smallmatrix}
        \diag(\bar\u^i[1])  \\
        \vdots \\
        \diag(\bar\u^i[C]) \\
    \end{smallmatrix}\right]
    \cdot \F_T\\
    = \circm(\idft(\bar\u^i)).
\end{gather}
Therefore, we have
\begin{equation}
\begin{split}
    \frac{1}{2}\frac{\partial \log\det \Big(\I + \alpha \cdot \circm(\bar\Z) \cdot \circm(\bar\Z)^{*} \Big)}{\partial \bar\z^i} 
    &= \frac{1}{2}\frac{\partial \log\det \Big(\I + \alpha \cdot \circm(\bar\Z) \cdot \circm(\bar\Z)^{*} \Big)}{\partial\circm(\bar\z^i)} \cdot \frac{\partial\circm(\bar\z^i)}{\partial \bar\z^i}\\
    &= T \cdot \idft(\bar\u^i).
\end{split}
\end{equation}
By collecting the results for all $i$, we have
\begin{gather}
    \frac{\partial \frac{1}{2T}\log\det \Bigg(\I + \alpha \cdot \circm(\bar\Z) \cdot \circm(\bar\Z)^{*} \Bigg)}{\partial \bar\Z}
    = \idft(\bar\U).
\end{gather}
In a similar fashion, we get
\begin{equation}
    \frac{\partial \frac{\gamma_j}{2T}\log\det\Bigg(\I + \alpha_j \cdot \circm(\bar\Z) \cdot \bar{\bm{\Pi}}^{j} \cdot \circm(\bar\Z)^{*} \Bigg)}{\partial \bar\Z} 
    = \gamma_j \cdot \idft(\bar\W^{j} \cdot \bm{\Pi}^j).
\end{equation}

\end{proof}

\clearpage

\newpage
\section{2D Circular Translation Invariance}\label{ap:2D-translation}
To a large degree, both conceptually and technically, the 2D case is very similar to the 1D case that we have studied carefully in the previous Appendix \ref{app:1D}. For the sake of consistency and completeness, we here gives a brief account.

\subsection{Doubly Block Circulant Matrix}

In this section, we consider $\z$ as a 2D signal such as an image, and use $H$ and $W$ to denote its ``height'' and ``width'', respectively. 
It will be convenient to work with both a matrix representation
\begin{equation}
    \z = 
    \begin{bmatrix}
    z(0, 0) & z(0, 1) & \cdots & z(0, W-1)\\
    z(1, 0) & z(1, 1) & \cdots & z(1, W-1)\\
    \vdots       & \vdots       & \ddots & \vdots        \\
    z(H-1, 0) & z(H-1, 1) & \cdots & z(H-1, W-1)\\
    \end{bmatrix} \quad \in \Re^{H \times W},
\end{equation}
as well as a vector representation
\begin{multline}
    \vec(\z)\doteq
    \Big[z(0, 0), \ldots, z(0, W-1), z(1, 0), \ldots, z(1, W-1), \ldots \\
\ldots, z(H-1, 0), \ldots, z(H-1, W-1)\Big]^* \in \Re^{(H \times W)}.
\end{multline}
We represent the circular translated version of $\z$ as $\trans_{p, q}(\z) \in \Re^{H \times W}$ by an amount of $p$ and $q$ on the vertical and horizontal directions, respectively. That is, we let
\begin{multline}
    \trans_{p, q}(\z) (h, w) \doteq \z(h - p \mod H, w - q \mod W), \\\forall (h, w) \in \{0, \ldots, H-1\} \times \{0, \ldots, W-1\}.
\end{multline}
It is obvious that $\trans_{0, 0}(\z) = \z$. 
Moreover, there is a total number of $H \times W$ distinct translations given by $\{\trans_{p, q}(\z), (p, q) \in \{0, \ldots, H-1\} \times \{0, \ldots, W-1\}\}$.  We may arrange the vector representations of them into a matrix and obtain 
\begin{multline}
    \circm(\z) \doteq \Big[\vec(\trans_{0, 0}(\z)), \ldots, \vec(\trans_{0, W-1}(\z)), \\ \vec(\trans_{1, 0}(\z)), \ldots, \vec(\trans_{1, W-1}(\z)),\\
    \ldots, \\
    \vec(\trans_{H-1, 0}(\z)), \ldots, \vec(\trans_{H-1, W-1}(\z))\Big] \in \Re^{(H\times W)\times (H\times W)}.
\end{multline}
The matrix $\circm(\z)$ is known as the \emph{doubly block circulant matrix} associated with $\z$ (see, e.g., \cite{abidi2016optimization,sedghi2018singular}). 

We now consider a multi-channel 2D signal represented as a tensor $\bar\z \in \Re^{C \times H \times W}$, where $C$ is the number of channels. 
The $c$-th channel of $\bar\z$ is represented as $\bar\z[c] \in \Re^{H \times W}$, and the $(h, w)$-th pixel is represented as $\bar\z(h, w) \in \Re^C$. 
To compute the coding rate reduction for a collection of such multi-channel 2D signals, we may flatten the tenor representation into a vector representation by concatenating the vector representation of each channel, i.e., we let
\begin{equation}
    \vec(\bar\z) = [\vec(\bar\z[1]) ^*, \ldots, \vec(\bar\z[C])^*]^* \quad \in \Re^{(C \times H \times W)}
\end{equation}
Furthermore, to obtain shift invariance for coding rate reduction, we may generate a collection of translated versions of $\bar\z$ (along two spatial dimensions). Stacking the vector representation for such translated copies as column vectors, we obtain
\begin{equation}
    \circm(\bar\z) \doteq 
    \begin{bmatrix}
    \circm(\bar\z[1])\\
    \vdots\\
    \circm(\bar\z[C])
    \end{bmatrix} \quad 
    \in \Re^{(C \times H \times W) \times (H \times W)}.
\end{equation}

We can now define a \emph{translation invariant coding rate reduction} for multi-channel 2D signals. 
Consider a collection of $m$ multi-channel 2D signals $\{\bar\z^i \in \Re^{C \times H \times W}\}_{i=1}^m$. 
Compactly representing the data by $\bar\Z \in \Re^{C \times H \times W \times m}$ where the $i$-th slice on the last dimension is $\bar\z^i$, we denote
\begin{equation}
    \circm(\bar\Z) = [\circm(\bar\z^1), \ldots, \circm(\bar\z^m)] \quad \in \Re^{(C\times H \times W) \times (H \times W \times m)}.
\end{equation}
Then, we define
\begin{multline}
\label{eq:2D-MCR2}
    \Delta R_\circm(\bar\Z, \bm{\Pi}) \doteq \frac{1}{HW}\Delta R(\circm(\bar\Z), \bar{\bm{\Pi}}) = 
    \frac{1}{2HW}\log\det \Bigg(\I + \alpha \cdot \circm(\bar\Z) \cdot \circm(\bar\Z)^{*} \Bigg) \\
    - \sum_{j=1}^{k}\frac{\gamma_j}{2HW}\log\det\Bigg(\I + \alpha_j \cdot \circm(\bar\Z) \cdot \bar{\bm{\Pi}}^{j} \cdot \circm(\bar\Z)^{*} \Bigg),
\end{multline}
where $\alpha = \frac{CHW}{mHW\epsilon^{2}} = \frac{C}{m\epsilon^{2}}$, $\alpha_j = \frac{CHW}{\textsf{tr}\left(\bm{\Pi}^{j}\right)HW\epsilon^{2}} = \frac{C}{\textsf{tr}\left(\bm{\Pi}^{j}\right)\epsilon^{2}}$, $\gamma_j = \frac{\textsf{tr}\left(\bm{\Pi}^{j}\right)}{m}$, and $\bar{\bm \Pi}^j$ is augmented membership matrix in an obvious way.

By following an analogous argument as in the 1D case, one can show that ReduNet for multi-channel 2D signals naturally gives rise to the multi-channel 2D circulant convolution operations. 
We omit the details, and focus on the construction of ReduNet in the frequency domain.

\subsection{Fast Computation in Spectral Domain}

\paragraph{Doubly block circulant matrix and 2D-DFT. }

Similar to the case of circulant matrices for 1D signals, all doubly block circulant matrices share the same set of eigenvectors, and these eigenvectors form a unitary matrix given by 
\begin{equation}
    \F \doteq \F_H \otimes \F_W  \quad \in \Co^{(H\times W) \times (H \times W)},
\end{equation}
where $\otimes$ denotes the Kronecker product and $\F_H, \F_W$ are defined as in \eqref{eq:dft-matrix}. 

Analogous to Fact \ref{fact:dft}, $\F$ defines 2D-DFT as follows.

\begin{fact}[2D-DFT as matrix-vector multiplication]\label{fact:2D-dft}
The 2D-DFT of a signal $\z \in \Re^{H \times W}$ can be computed as
    \begin{equation}
        \vec(\dft(\z)) \doteq \F \cdot \vec(\z) \quad \in \Co^{(H \times W)},
    \end{equation}
where
\begin{multline}\label{eq:2d-dft}
    \dft(\z)(p, q) = \frac{1}{\sqrt{H \cdot W}} \sum_{h=0}^{H-1} \sum_{w=0}^{W-1} \z(h, w) \cdot \omega_H ^{p \cdot h} \omega_W ^{q\cdot w},  \\~~\forall (p, q) \in \{0, \ldots, H-1\} \times \{0, \ldots, W-1\}.
\end{multline}
The 2D-IDFT of a signal $\bv \in \Co^{H \times W}$ can be computed as
    \begin{equation} 
        \vec(\idft(\bv)) \doteq \F_T^* \cdot \vec(\bv) \quad \in \Co^{(H \times W)},
    \end{equation}
    where
    \begin{multline}
        \idft(\bv)(h, w) = \frac{1}{\sqrt{H \cdot W}} \sum_{p=0}^{H-1}\sum_{q=0}^{W-1} v(p, q) \cdot \omega_H ^{-p \cdot h}\omega_W ^{-q \cdot w},
        \\~~\forall (h, w) \in \{0, \ldots, H-1\} \times \{0, \ldots, W-1\}.
    \end{multline}
\end{fact}

Analogous to Fact \ref{fact:2d-dft-circulant}, $\F$ relates $\dft(\z)$ and $\circm(\z)$ as follows.
\begin{fact}[2D-DFT are eigenvalues of the doubly block circulant matrix] Given a signal $\z \in \Co^{H \times W}$, we have
    \begin{equation}\label{eq:dft-circulant}
        \F \cdot \circm(\z) \cdot \F^* = \diag(\vec(\dft(\z)))  \quad \mbox{or} \quad \circm(\z) = \F^* \cdot \diag(\vec(\dft(\z))) \cdot \F.
    \end{equation}
\label{fact:2d-dft-circulant}
\end{fact}

\paragraph{Doubly block circulant matrix and 2D-DFT for multi-channel signals.}

We now consider multi-channel 2D signals $\bar\z \in \Re^{C \times H \times W}$. 
Let $\dft(\bar\z) \in \Co^{C \times H \times W}$ be a matrix where the $c$-th slice on the first dimension is the DFT of the corresponding signal $\z[c]$.
That is, $\dft(\bar\z)[c] = \dft(\z[c]) \in \Co^{H \times W}$. 
We use $\dft(\bar\z)(p, q) \in \Co^{C}$ to denote slicing of $\bar\z$ on the frequency dimensions. 

By using Fact~\ref{fact:2d-dft-circulant}, $\circm(\bar\z)$ and $\dft(\bar\z)$ are related as follows:
\begin{multline}
    \circm(\bar\z) 
    =
    \begin{bmatrix}
        \F^* \cdot \diag(\vec(\dft(\z[1])))  \cdot \F\\
        \vdots \\
        \F^* \cdot \diag(\vec(\dft(\z[C]))) \cdot \F\\
    \end{bmatrix}
    \\
    =
    \begin{bmatrix}
        \F^*  & \cdots & \0 \\
        \0    & \cdots & \0 \\
        \vdots  & \ddots & \vdots \\
        \0   & \cdots & \F^* \\
    \end{bmatrix}
    \cdot
    \begin{bmatrix}
        \diag(\vec(\dft(\z[1])))  \\
        \diag(\vec(\dft(\z[2])))  \\
        \vdots \\
        \diag(\vec(\dft(\z[C])))  \\
    \end{bmatrix}
    \cdot \F.
\end{multline}
Similar to the 1D case, this relation can be leveraged to produce a fast implementation of ReduNet in the spectral domain. 

\paragraph{Translation-invariant ReduNet in the Spectral Domain. }

Given a collection of multi-channel 2D signals $\bar\Z \in \Re^{C \times H \times W \times m}$, we denote
\begin{equation}
    \dft(\bar\Z)(p, q) \doteq [\dft(\bar\z^1)(p, q), \ldots, \dft(\bar\z^m)(p, q)] \quad \in \Re^{C \times m}.
\end{equation}
We introduce the notations $\bar\cE(p, q) \in \Re^{C \times C \times H \times W}$ and $\bar\cC^j(p, q) \in \Re^{C \times C \times H \times W}$ given by
\begin{eqnarray}
    \bar\cE(p, q) &\doteq& \alpha \cdot \left[\I + \alpha \cdot \dft(\bar\Z)(p, q) \cdot \dft(\bar\Z)(p, q)^* \right]^{-1} \quad \in \Co^{C\times C}, \\
    \bar\cC^j(p, q) &\doteq& \alpha_j \cdot\left[\I + \alpha_j \cdot \dft(\bar\Z)(p, q) \cdot \bm{\Pi}_j \cdot \dft(\bar\Z)(p, q)^*\right]^{-1} \quad \in \Co^{C\times C}.
\end{eqnarray}
In above, $\bar\cE(p, q)$ (resp., $\bar\cC^j(p, q)$) is the $(p, q)$-th slice of $\bar\cE$ (resp., $\bar\cC^j$) on the last two dimensions. 
Then, the gradient of $\Delta R_\circm(\bar\Z, \bm{\Pi})$ with respect to $\bar\Z$ can be calculated by the following result.

\begin{theorem}[Computing multi-channel 2D convolutions $\bar{\bm E}$ and $\bar{\bm C}^j$]\label{thm:2D-convolution}
Let $\bar\U \in \Co^{C \times H \times W \times m}$ and $\bar\W^{j} \in \Co^{C \times H \times W \times m}, j=1,\ldots, k$ be given by 
\begin{eqnarray}
    \bar\U(p, q) &\doteq& \bar\cE(p, q) \cdot \dft(\bar\Z)(p, q), \\
    \bar\W^{j}(p, q) &\doteq& \bar\cC^j(p, q) \cdot \dft(\bar\Z)(p, q), \quad j=1,\ldots, k,
\end{eqnarray}
for each  $(p, q) \in \{0, \ldots, H-1\}\times \{0, \ldots, W-1\}$. 
Then, we have
\begin{eqnarray}
    \frac{1}{2HW}\frac{\partial \log \det (\I + \alpha \cdot \circm(\bar\Z) \circm(\bar\Z)^{*} )}{\partial \bar\Z} &=& \idft(\bar\U), \\
    \frac{1}{2HW}\frac{\partial \left( \gamma_j  \log \det (\I + \alpha_j \cdot \circm(\bar\Z) \bm \bar{\bm \Pi}^j \circm(\bar\Z)^{*} )  \right)}{\partial \bar\Z}&=&
    \gamma_j \cdot \idft(\bar\W^{j} \bm \Pi^j).
\end{eqnarray}
\end{theorem}
This result shows that the calculation of the derivatives for the 2D case is analogous to that of the 1D case. 
Therefore, the construction of the ReduNet for 2D translation invariance can be performed using Algorithm~\ref{alg:training-1D} with straightforward extensions. 

%% file: appendix-experiment.tex
\clearpage
\section{Implementation Details and Additional Experiments}\label{sec:appendix-exp}
Disclaimer: in this work we do not particularly optimize any of the hyper parameters, such as the number of initial channels, kernel sizes, and learning rate etc., for the best performance. The choices are mostly for convenience and just minimally adequate to verify the concept, due to limited computational resource.

\subsection{Additional Experiments on learning mixture of Gaussians in $\mathbb{S}^1$ and $\mathbb{S}^2$}
\begin{figure}[ht]
    \centering
    \begin{subfigure}[b]{0.24\textwidth}
        \centering
        \includegraphics[width=\textwidth]{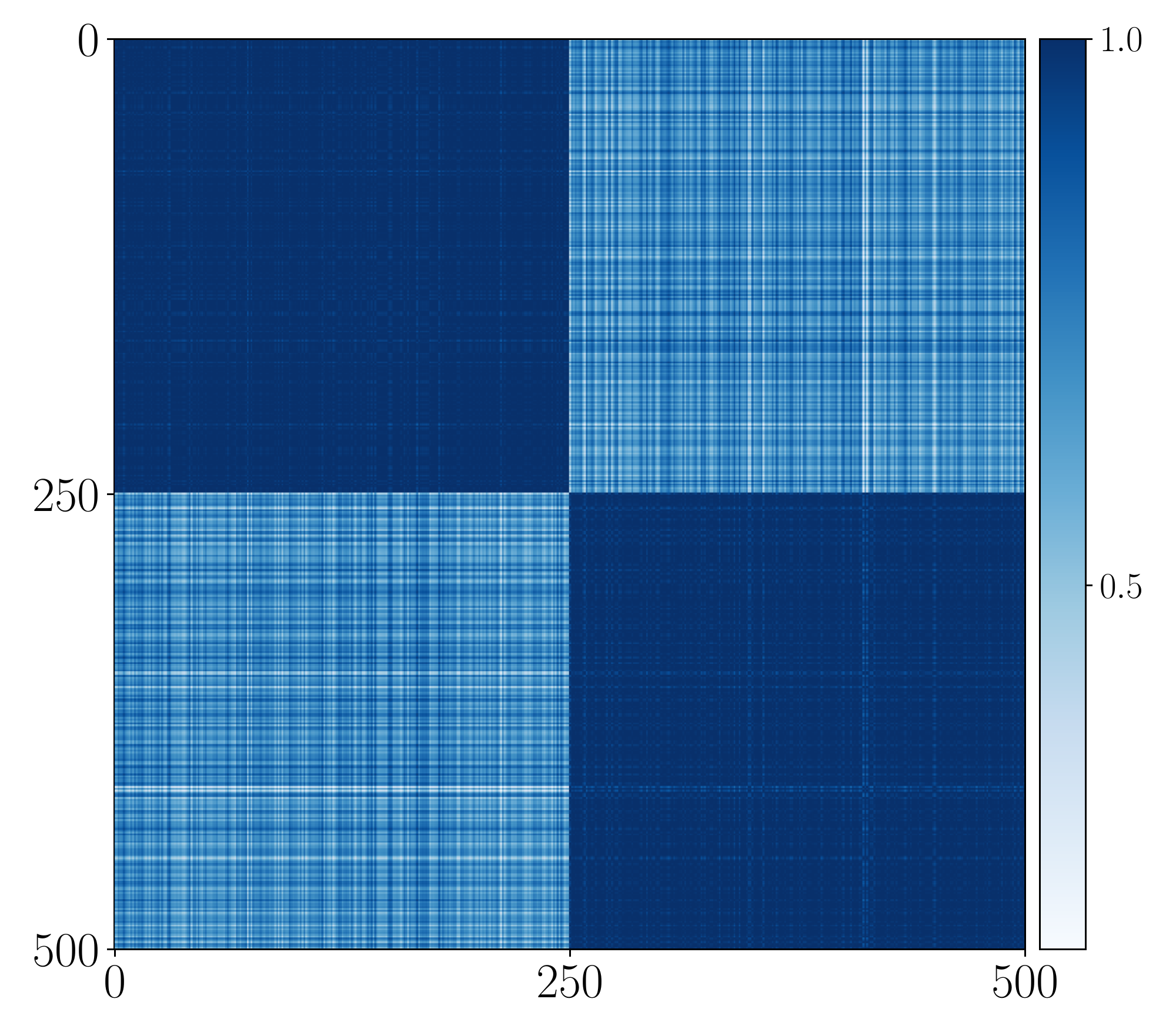}
        \caption{$\X_{\text{train}}$}
        \label{fig:gaussian-heatmaps-a}
    \end{subfigure}
    \begin{subfigure}[b]{0.24\textwidth}
        \centering
        \includegraphics[width=\textwidth]{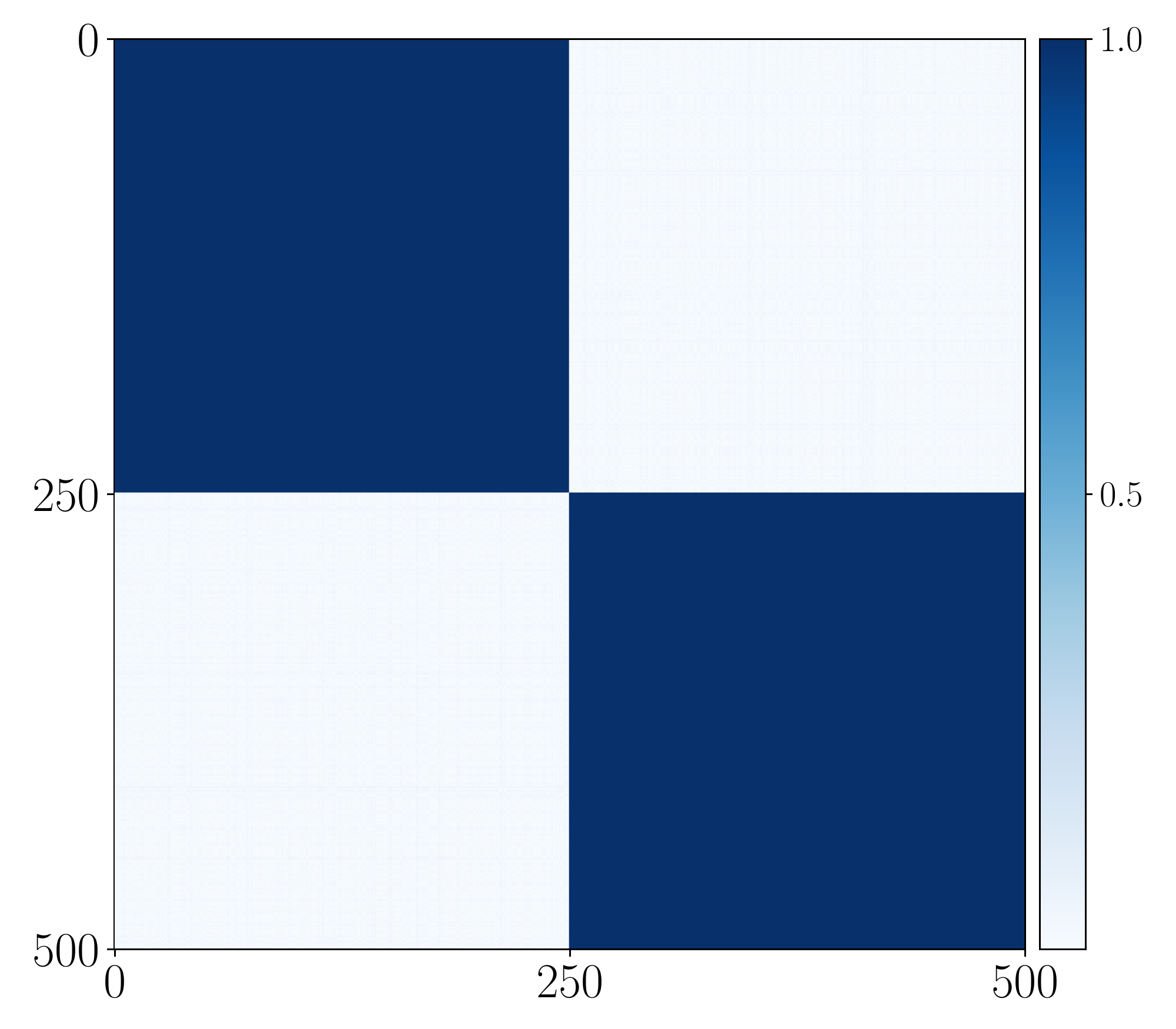}
        \caption{$\Z_{\text{train}}$}
        \label{fig:gaussian-heatmaps-b}
    \end{subfigure}
    \begin{subfigure}[b]{0.24\textwidth}
        \centering
        \includegraphics[width=\textwidth]{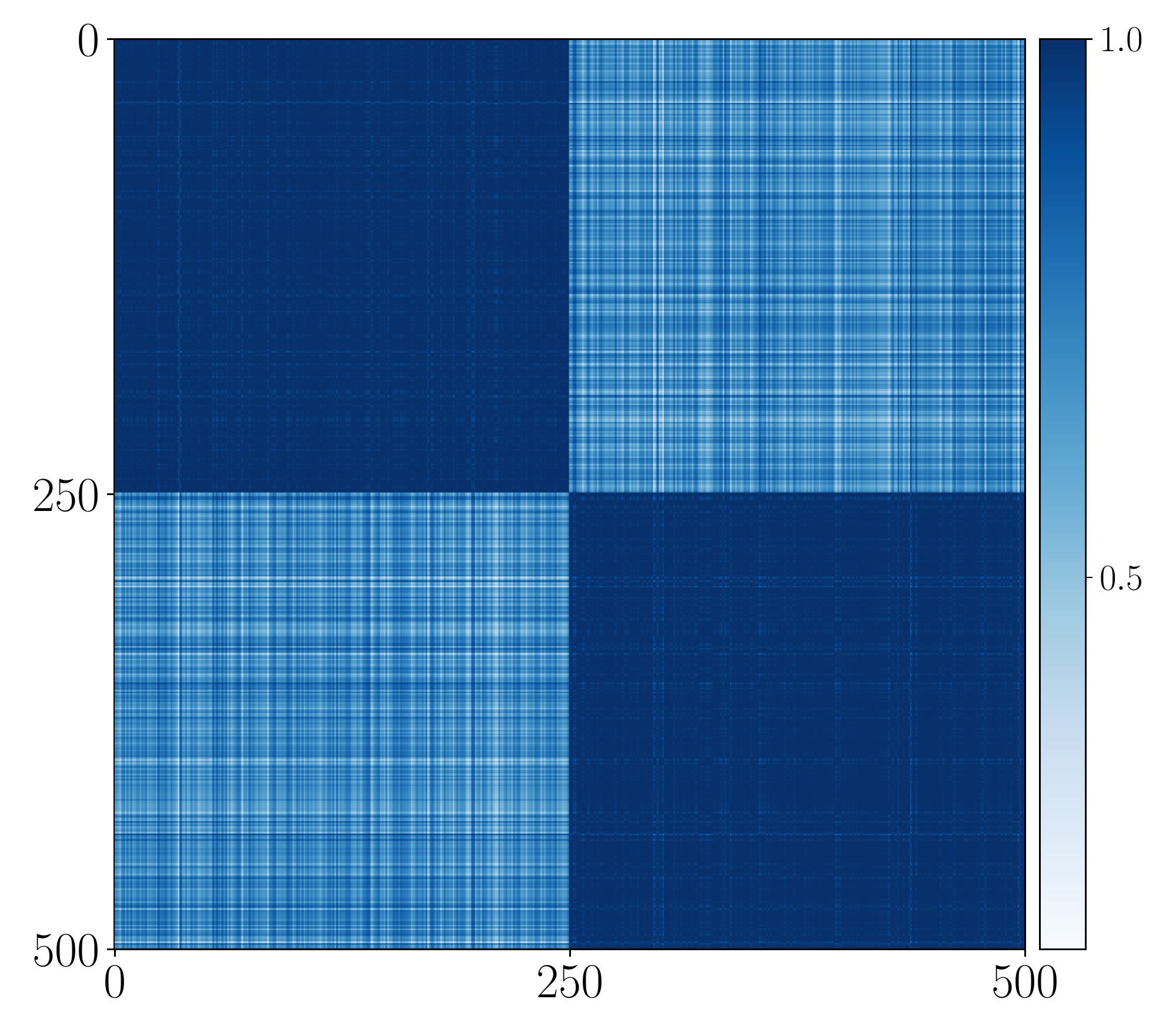}
        \caption{$\X_{\text{test}}$}
        \label{fig:gaussian-heatmaps-c}
    \end{subfigure}
    \begin{subfigure}[b]{0.24\textwidth}
        \centering
        \includegraphics[width=\textwidth]{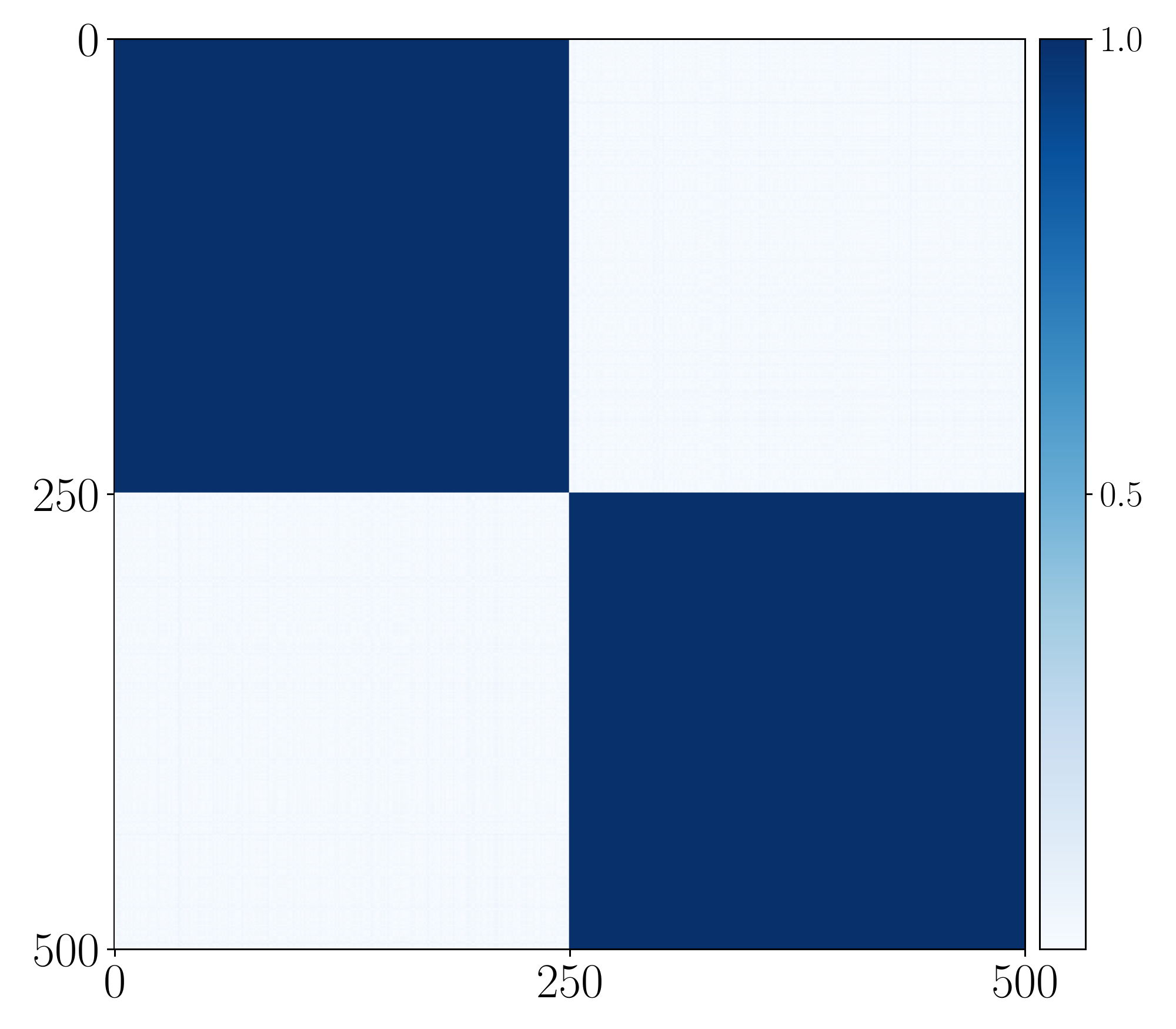}
        \caption{$\Z_{\text{test}}$}
        \label{fig:gaussian-heatmaps-d}
    \end{subfigure}
    \begin{subfigure}[b]{0.24\textwidth}
        \centering
        \includegraphics[width=\textwidth]{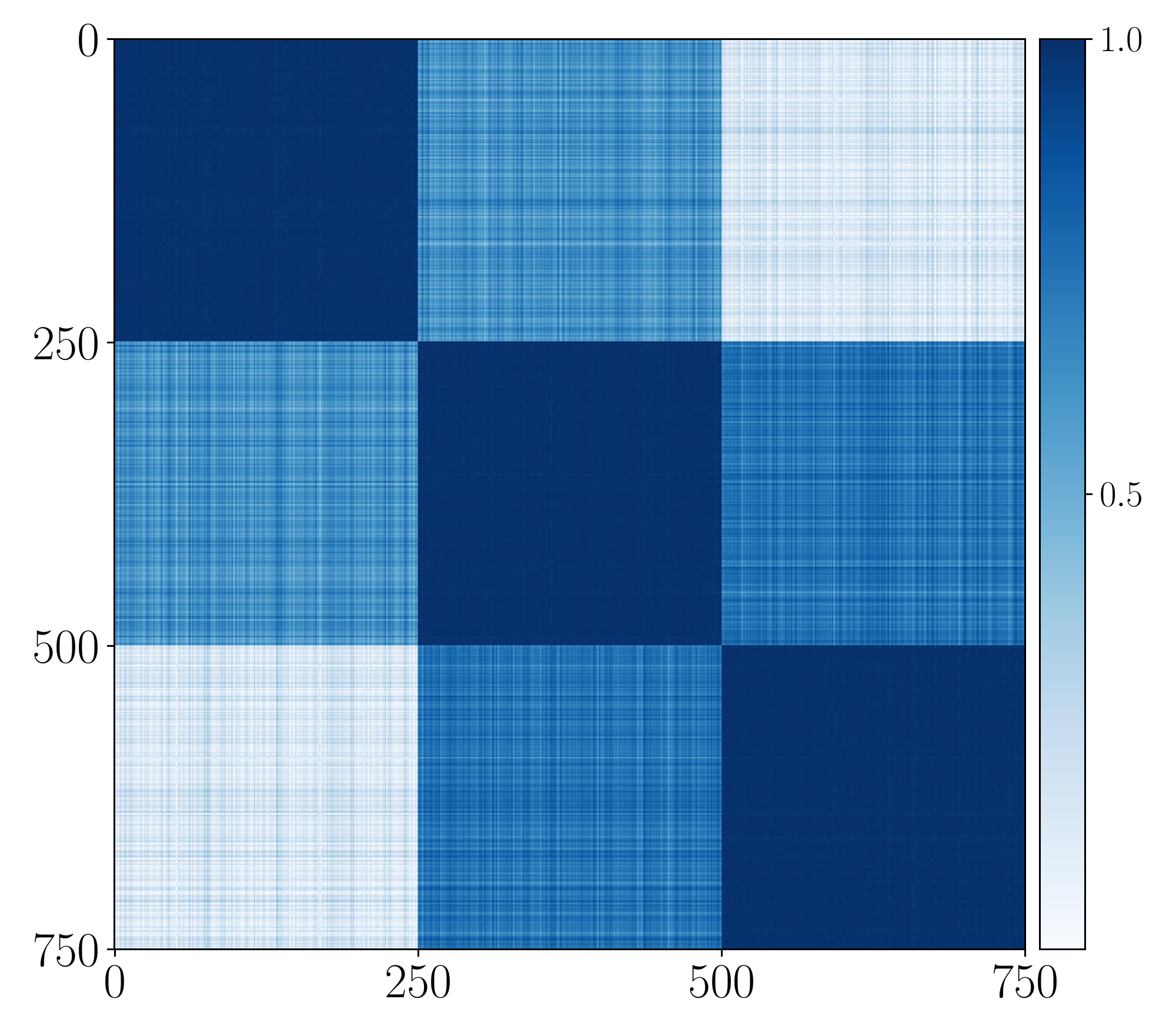}
        \caption{$\X_{\text{train}}$}
        \label{fig:gaussian-heatmaps-e}
    \end{subfigure}
    \begin{subfigure}[b]{0.24\textwidth}
        \centering
        \includegraphics[width=\textwidth]{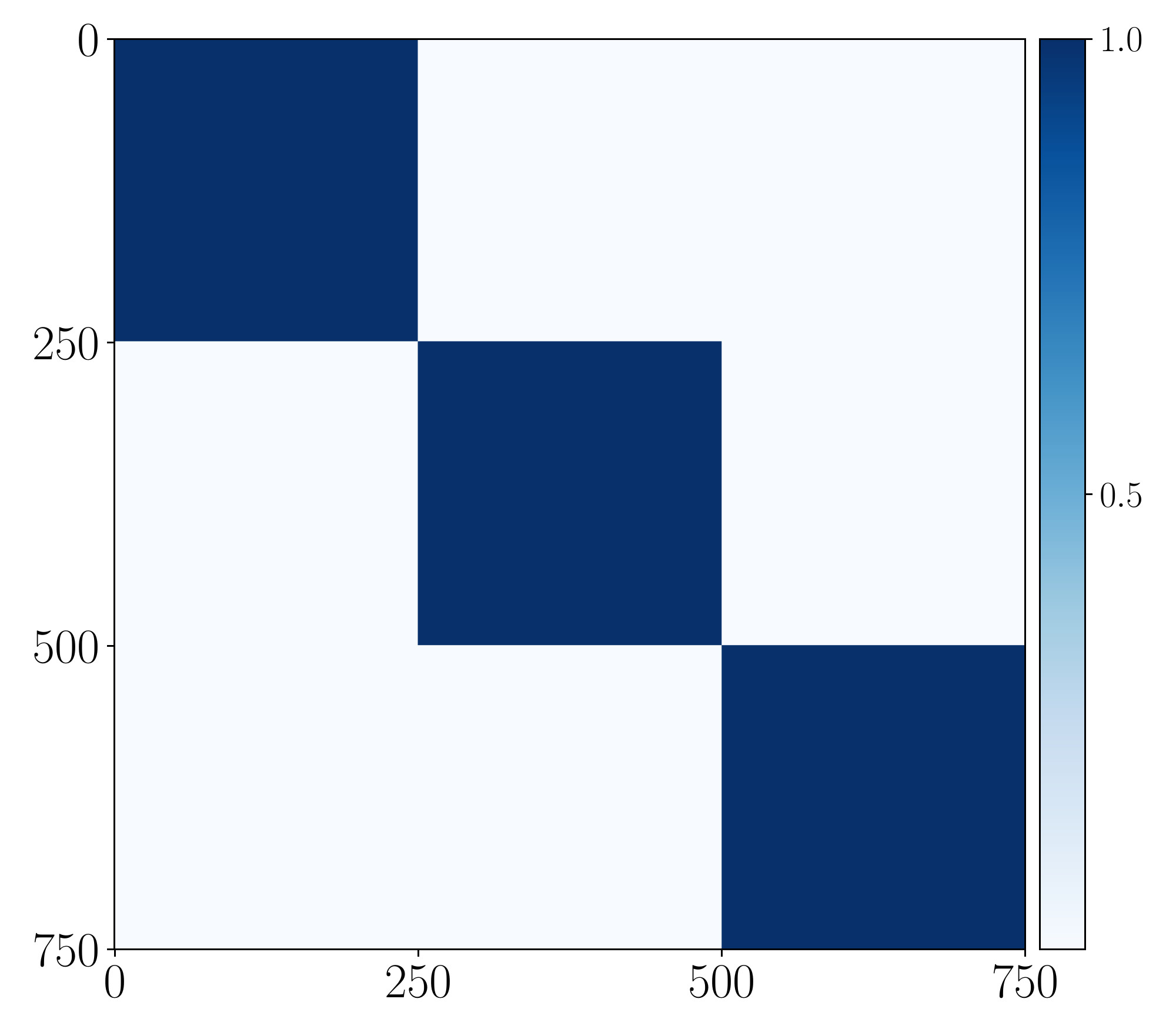}
        \caption{$\Z_{\text{train}}$}
        \label{fig:gaussian-heatmaps-f}
    \end{subfigure}
    \begin{subfigure}[b]{0.24\textwidth}
        \centering
        \includegraphics[width=\textwidth]{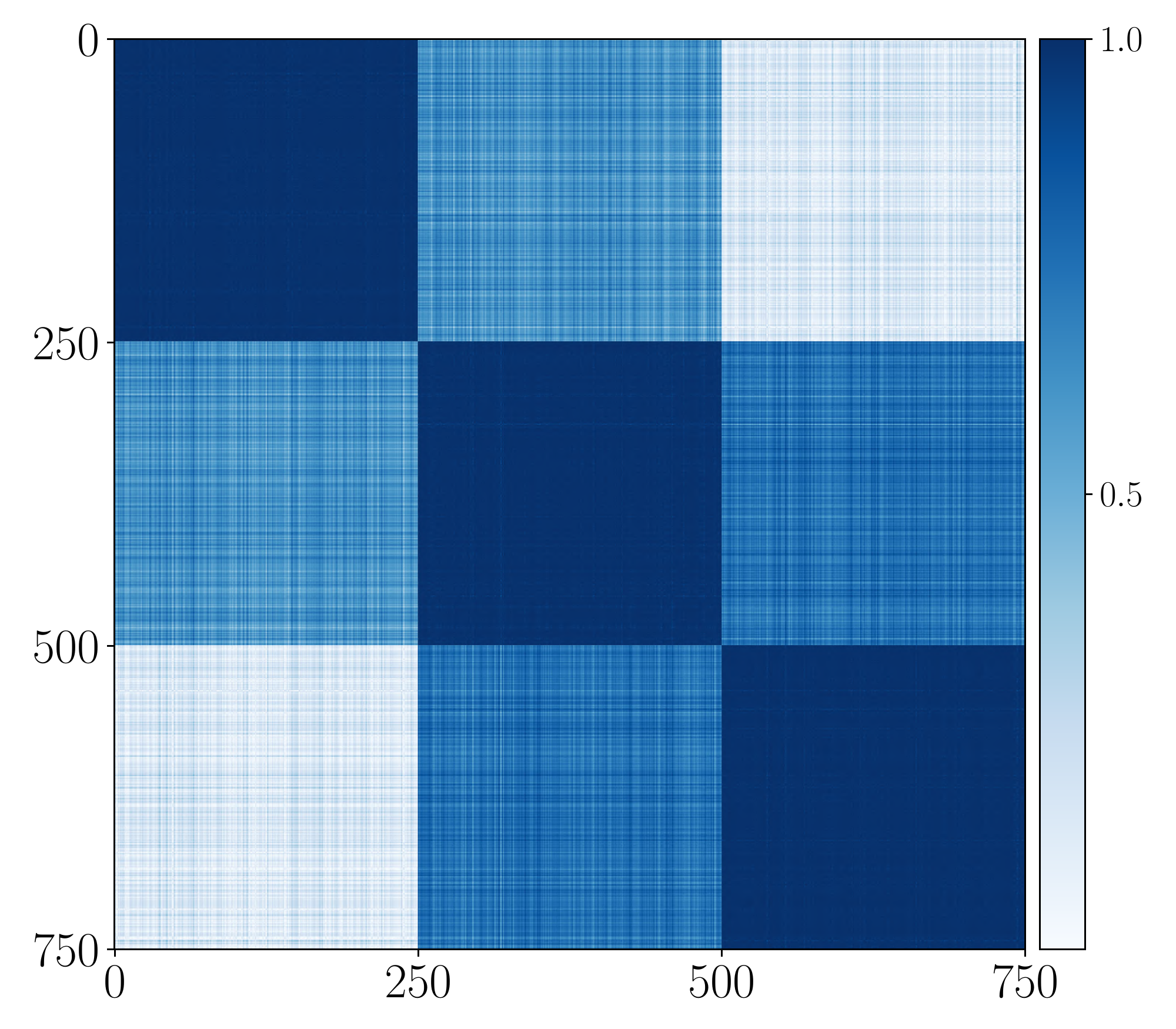}
        \caption{$\X_{\text{test}}$}
        \label{fig:gaussian-heatmaps-g}
    \end{subfigure}
    \begin{subfigure}[b]{0.24\textwidth}
        \centering
        \includegraphics[width=\textwidth]{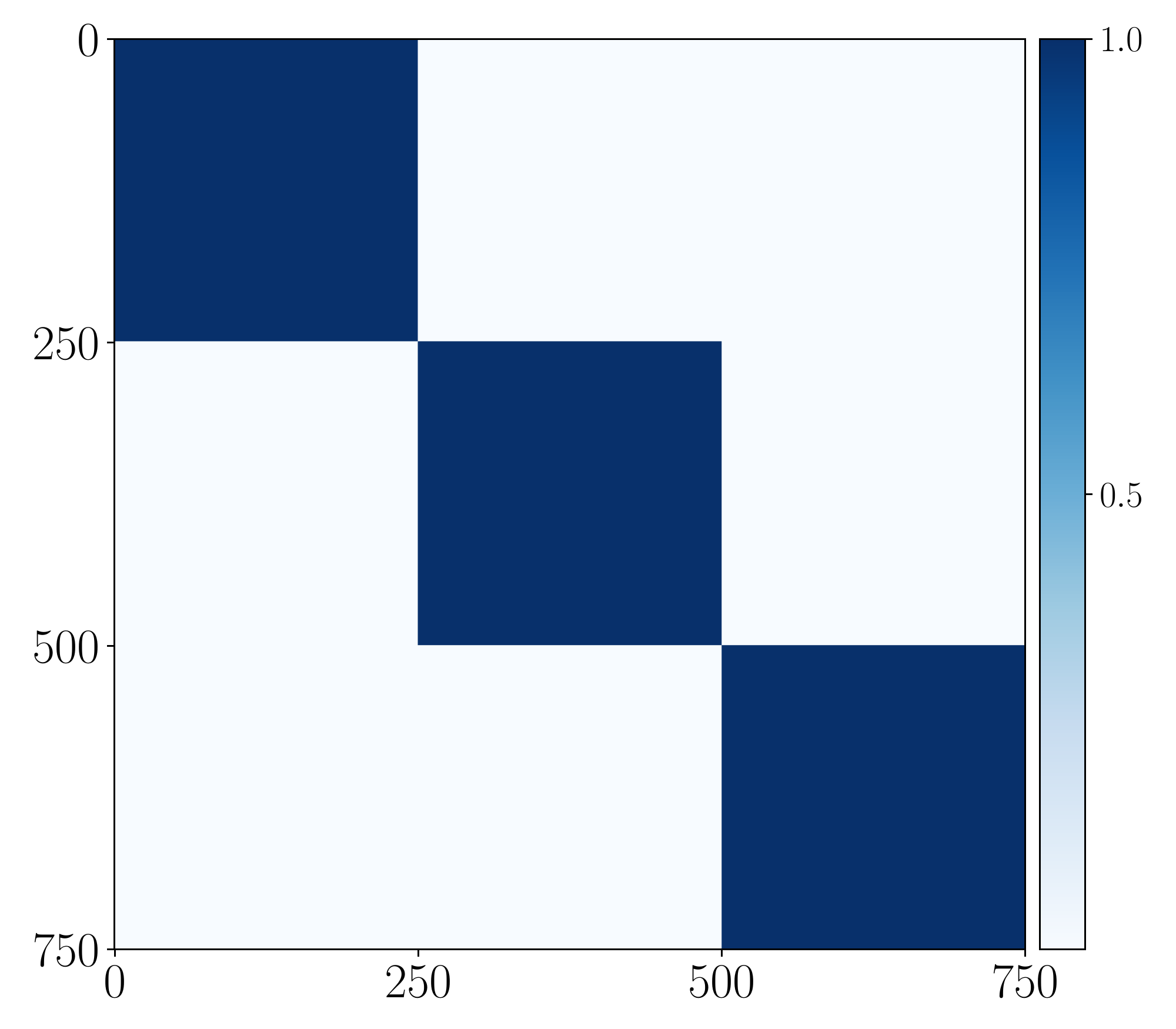}
        \caption{$\Z_{\text{test}}$}
        \label{fig:gaussian-heatmaps-h}
    \end{subfigure}
    \caption{Cosine similarity (absolute value) for $2D$ and $3D$ Mixture of Gaussians. Lighter color implies samples are more orthogonal. }
    \label{fig:appendix-gaussian-heatmaps}
\end{figure}

We provide the cosine similarity results for the experiments described in Figure~\ref{fig:gaussian2d3d-scatter-heatmap}. The results are shown in Figure~\ref{fig:appendix-gaussian-heatmaps}. We can observe that the network can map the data points to orthogonal subspaces.

\begin{figure}[ht]
    \centering
    \begin{subfigure}[b]{0.37\textwidth}
        \centering
        \begin{subfigure}[b]{0.49\textwidth}
            \centering
            \includegraphics[width=\textwidth]{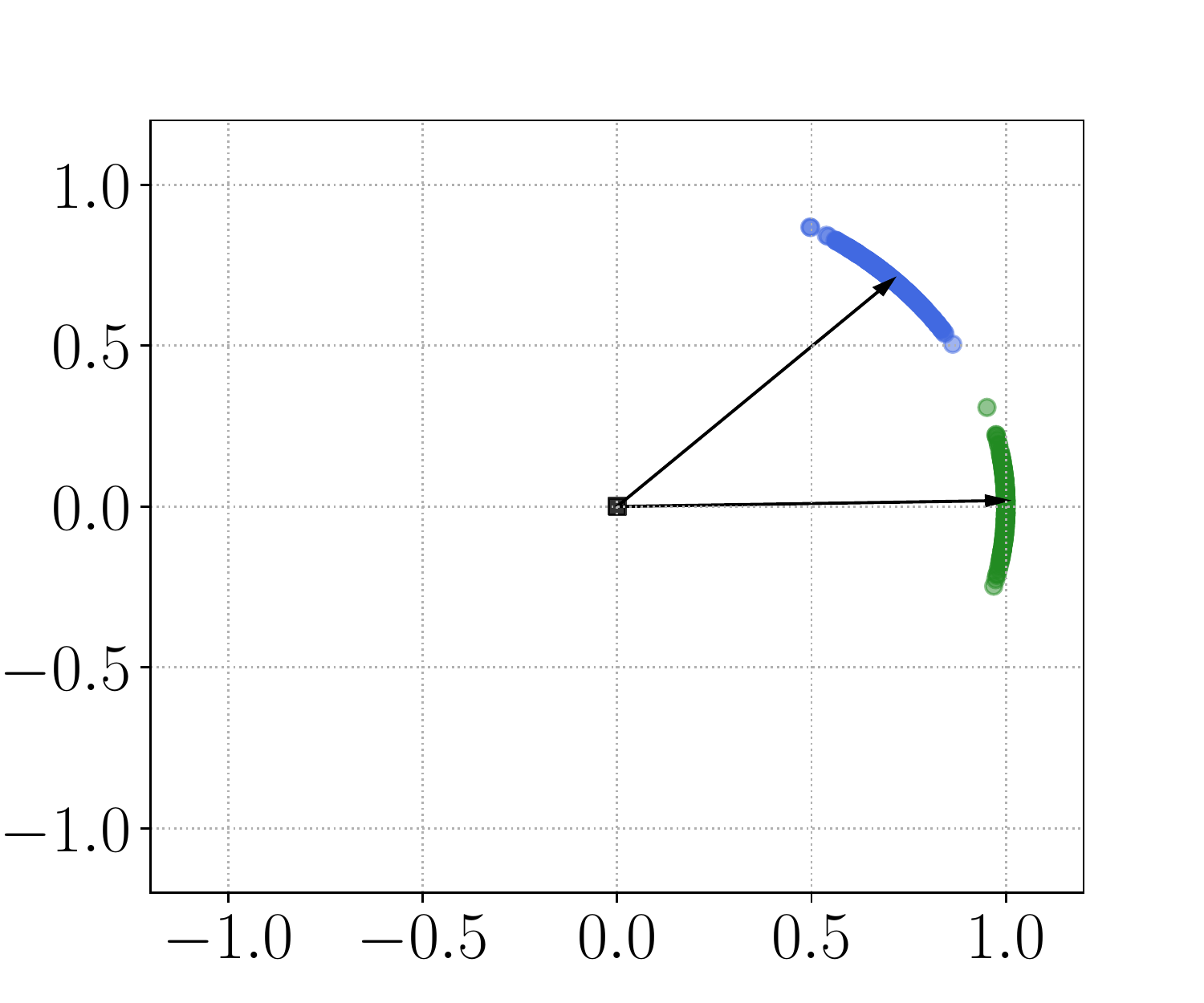}
        \end{subfigure}
        \begin{subfigure}[b]{0.49\textwidth}
            \centering
            \includegraphics[width=\textwidth]{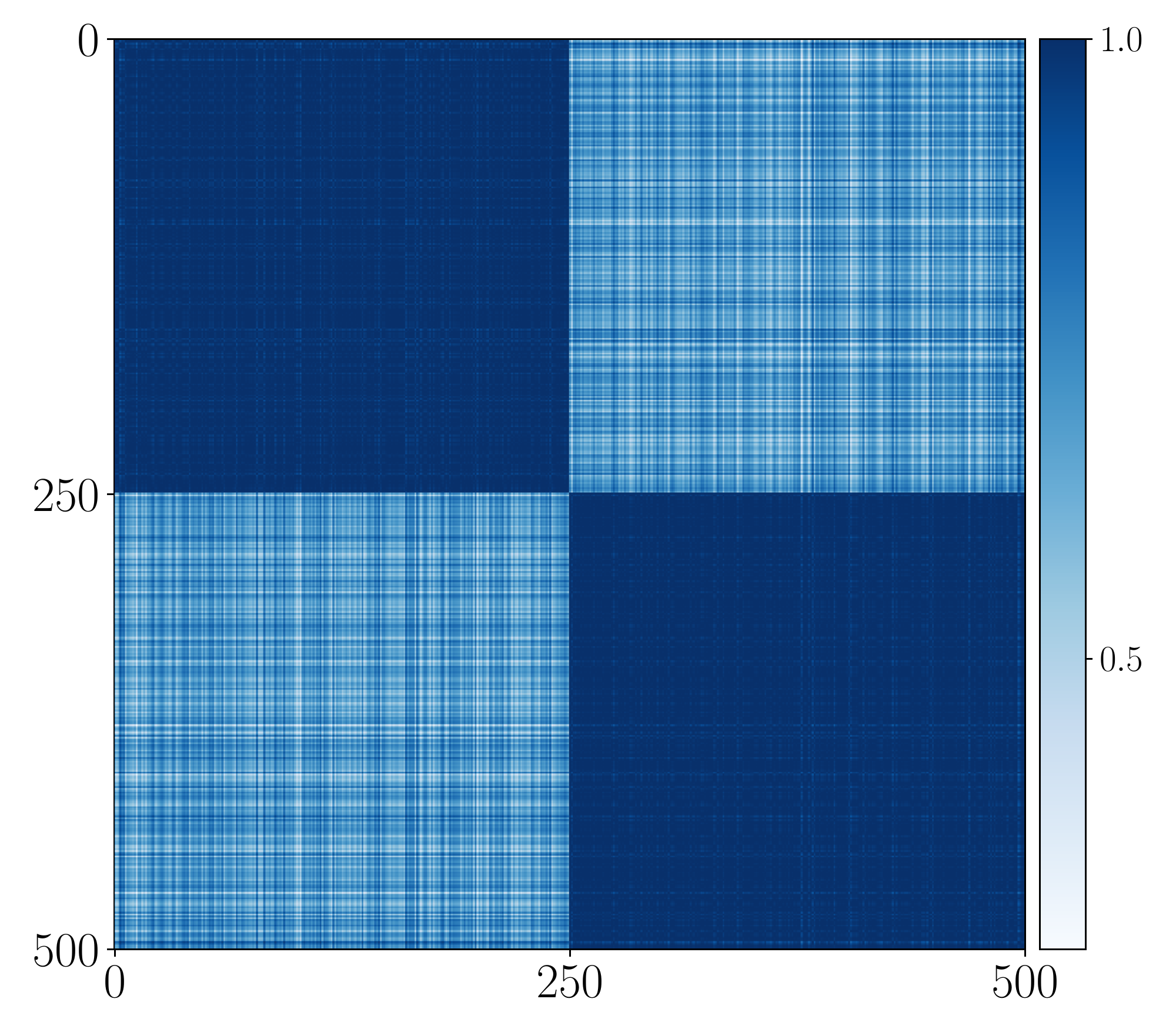}
        \end{subfigure}
        \caption{$\X (2D)$ (\textbf{left: }scatter plot; \textbf{right: }cosine similarity visualization)}
    \end{subfigure}
    \begin{subfigure}[b]{0.37\textwidth}
        \centering
        \begin{subfigure}[b]{0.49\textwidth}
            \centering
            \includegraphics[width=\textwidth]{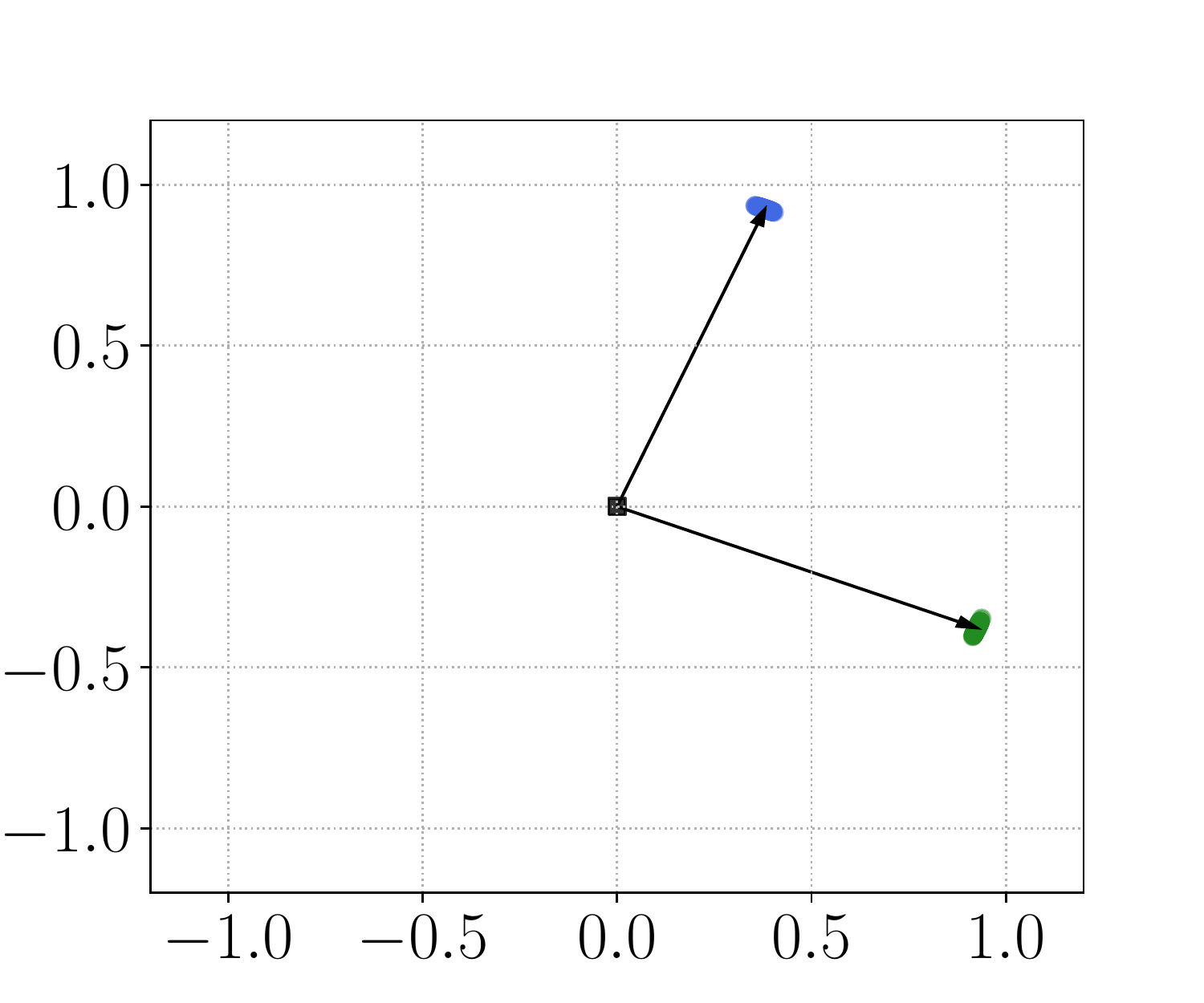}
        \end{subfigure}
        \begin{subfigure}[b]{0.49\textwidth}
            \centering
            \includegraphics[width=\textwidth]{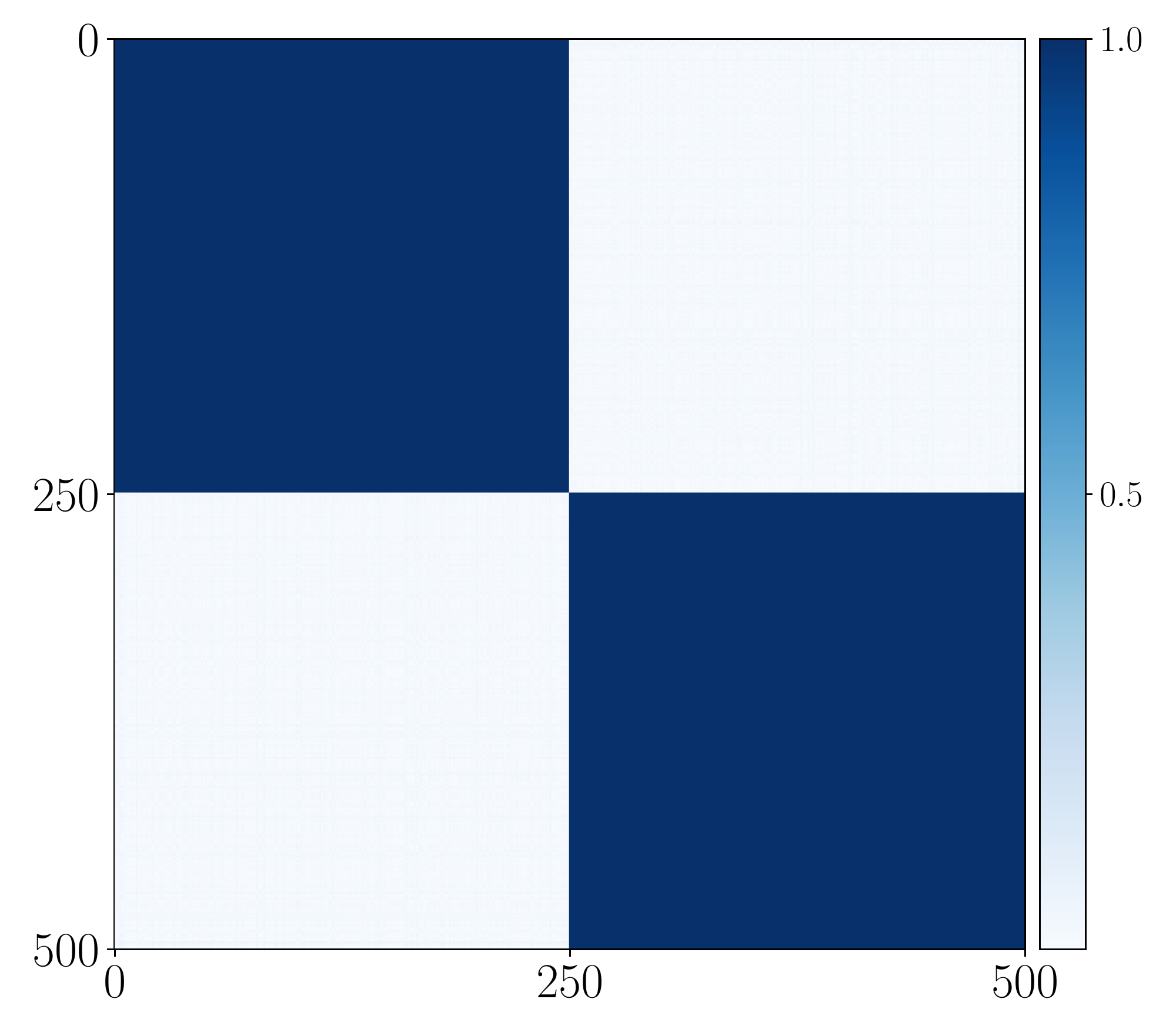}
        \end{subfigure}
        \caption{$\Z (2D)$ (\textbf{left: }scatter plot; \textbf{right: }cosine similarity visualization)}
    \end{subfigure}
    \begin{subfigure}[b]{0.24\textwidth}
        \centering
        \includegraphics[width=\textwidth]{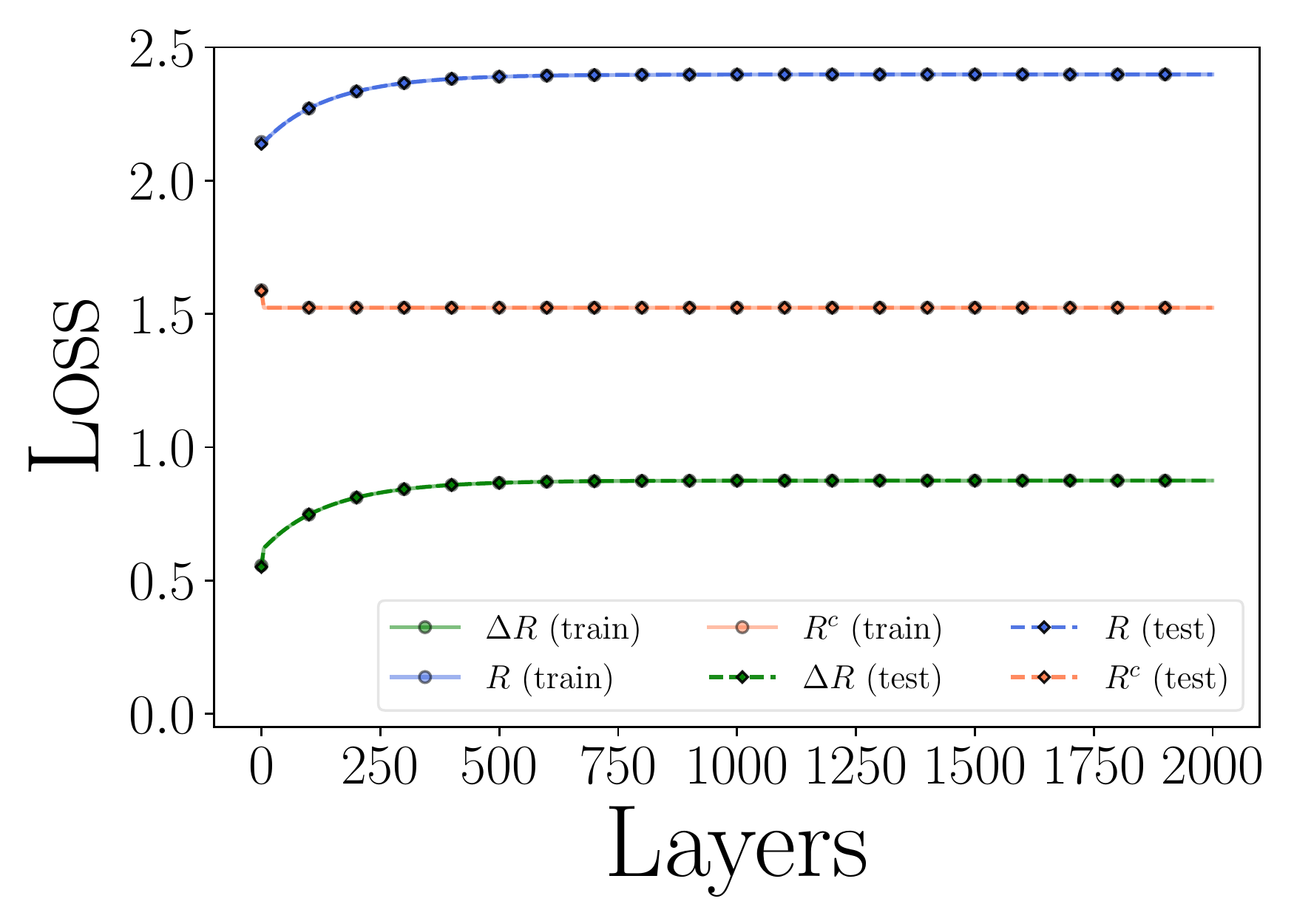}
        \caption{Loss}
    \end{subfigure}
    \caption{Learning mixture of Gaussians in $\mathbb{S}^1$ and $\mathbb{S}^2$. For $\mathbb{S}^1$, we set $\sigma_1 = \sigma_2 = 0.1$. }
    \label{fig:appendix-guassian-exp1}
\end{figure}

\textbf{Additional experiments on  $\mathbb{S}^1$ and $\mathbb{S}^2$.} We also provide additional experiments on learning mixture of Gaussians in $\mathbb{S}^1$ and $\mathbb{S}^2$ in Figure~\ref{fig:appendix-guassian-exp1}. We can observe similar behavior of the proposed ReduNet: the network can map data points from different classes to orthogonal subspaces.

\begin{figure}[ht]
    \centering
    \begin{subfigure}[b]{0.37\textwidth}
        \centering
        \begin{subfigure}[b]{0.49\textwidth}
            \centering
            \includegraphics[width=\textwidth]{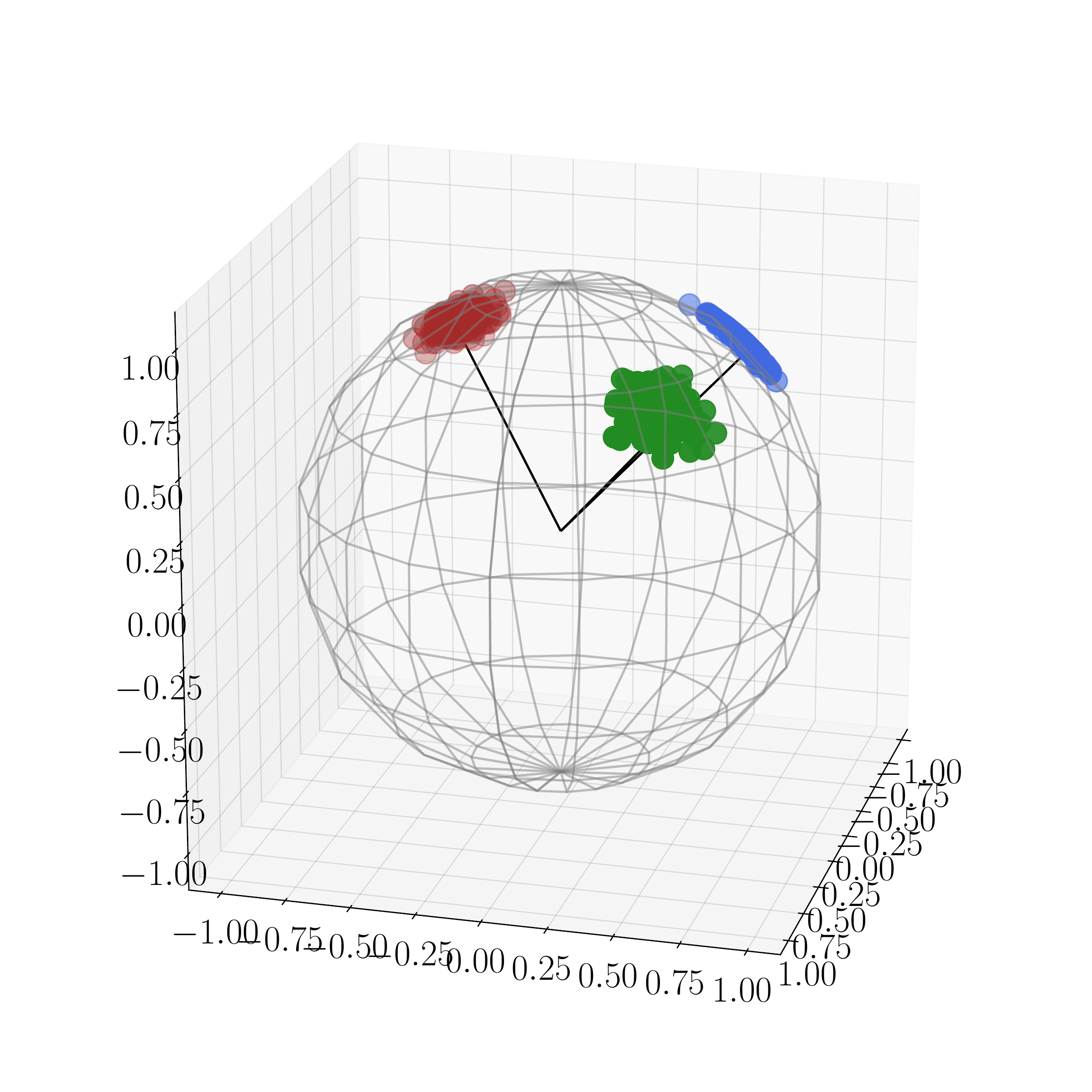}
        \end{subfigure}
        \begin{subfigure}[b]{0.49\textwidth}
            \centering
            \includegraphics[width=\textwidth]{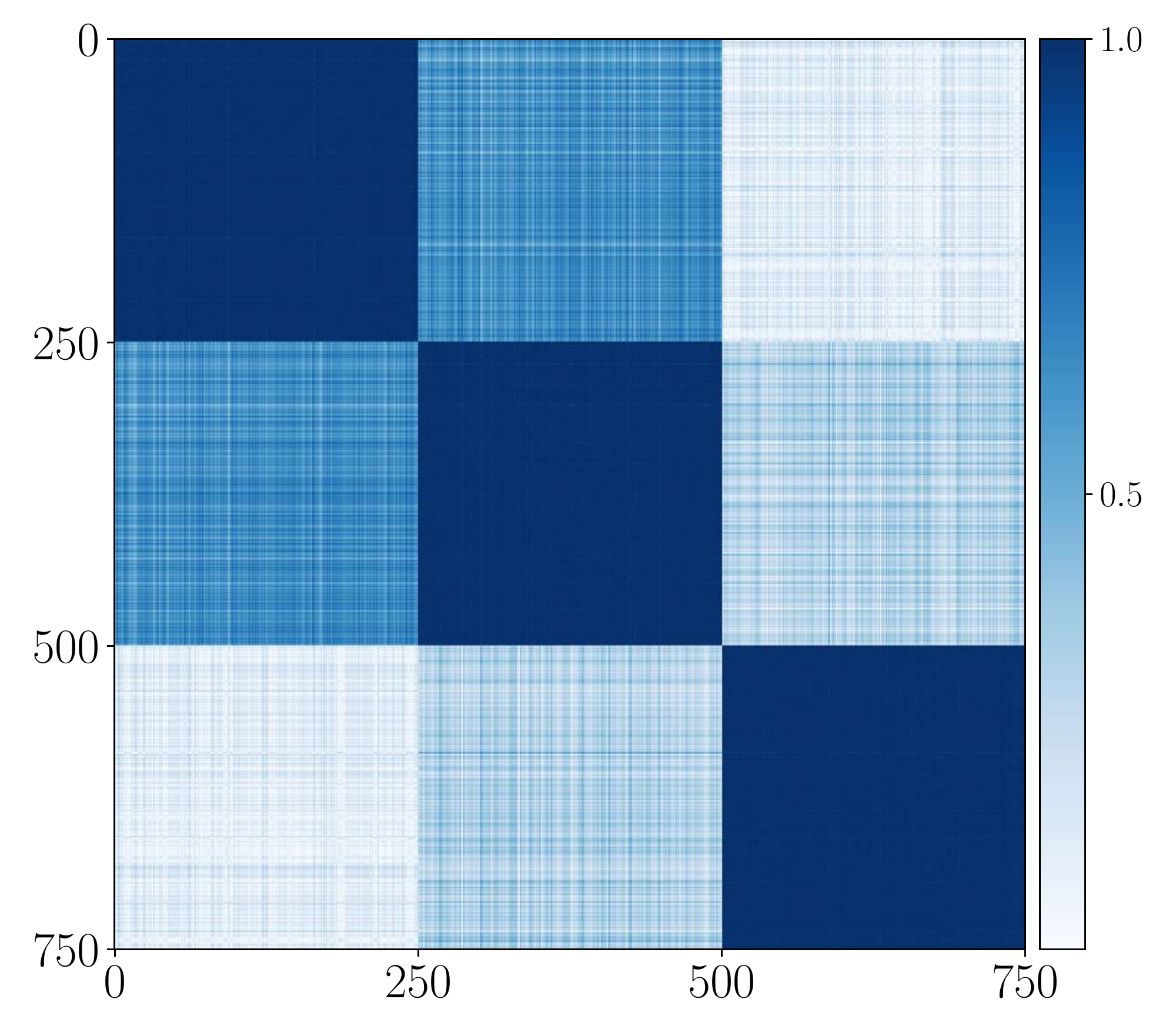}
        \end{subfigure}
        \caption{$\X (3D)$ (\textbf{left: }scatter plot; \textbf{right: }cosine similarity visualization)}
    \end{subfigure}
    \begin{subfigure}[b]{0.37\textwidth}
        \centering
        \begin{subfigure}[b]{0.49\textwidth}
            \centering
            \includegraphics[width=\textwidth]{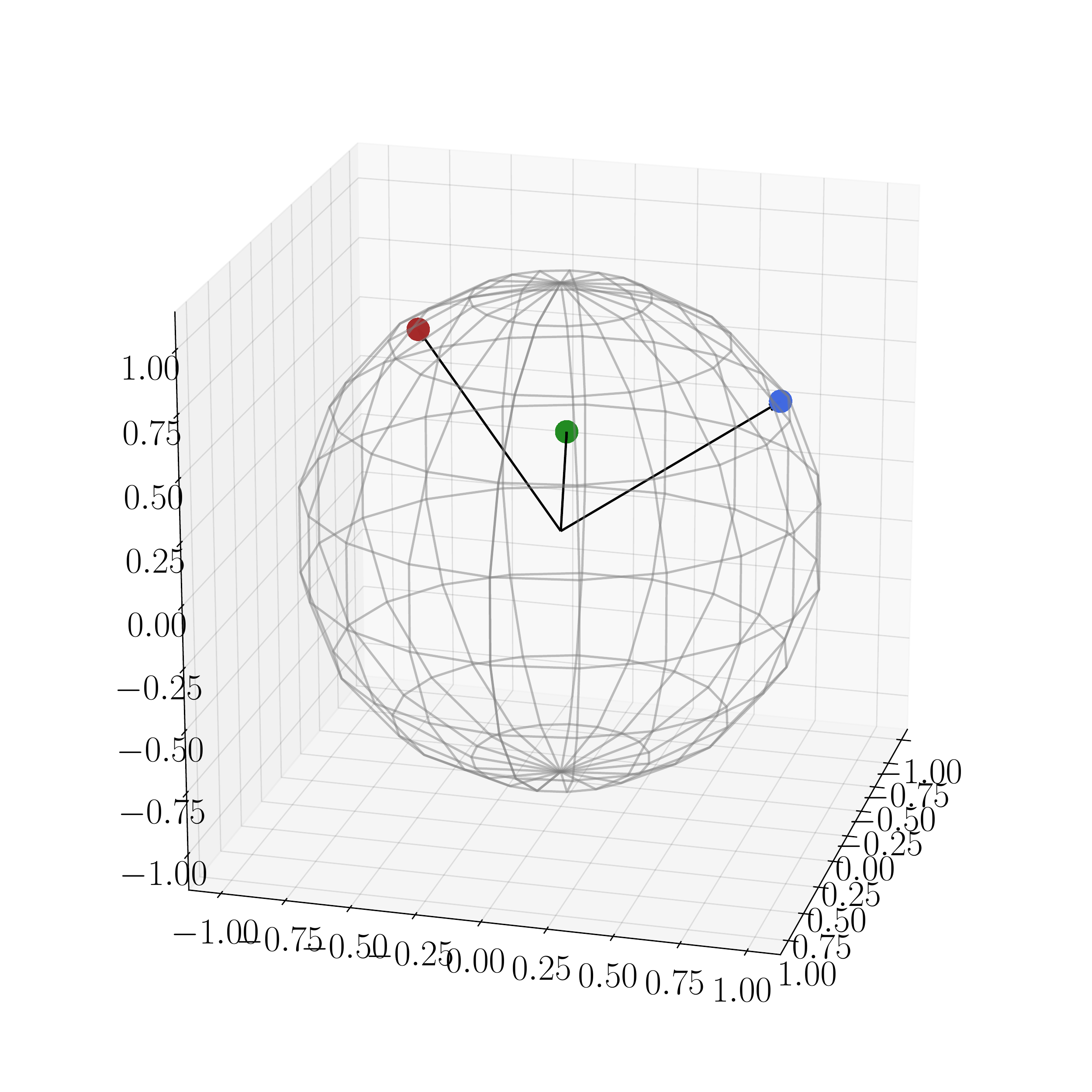}
        \end{subfigure}
        \begin{subfigure}[b]{0.49\textwidth}
            \centering
            \includegraphics[width=\textwidth]{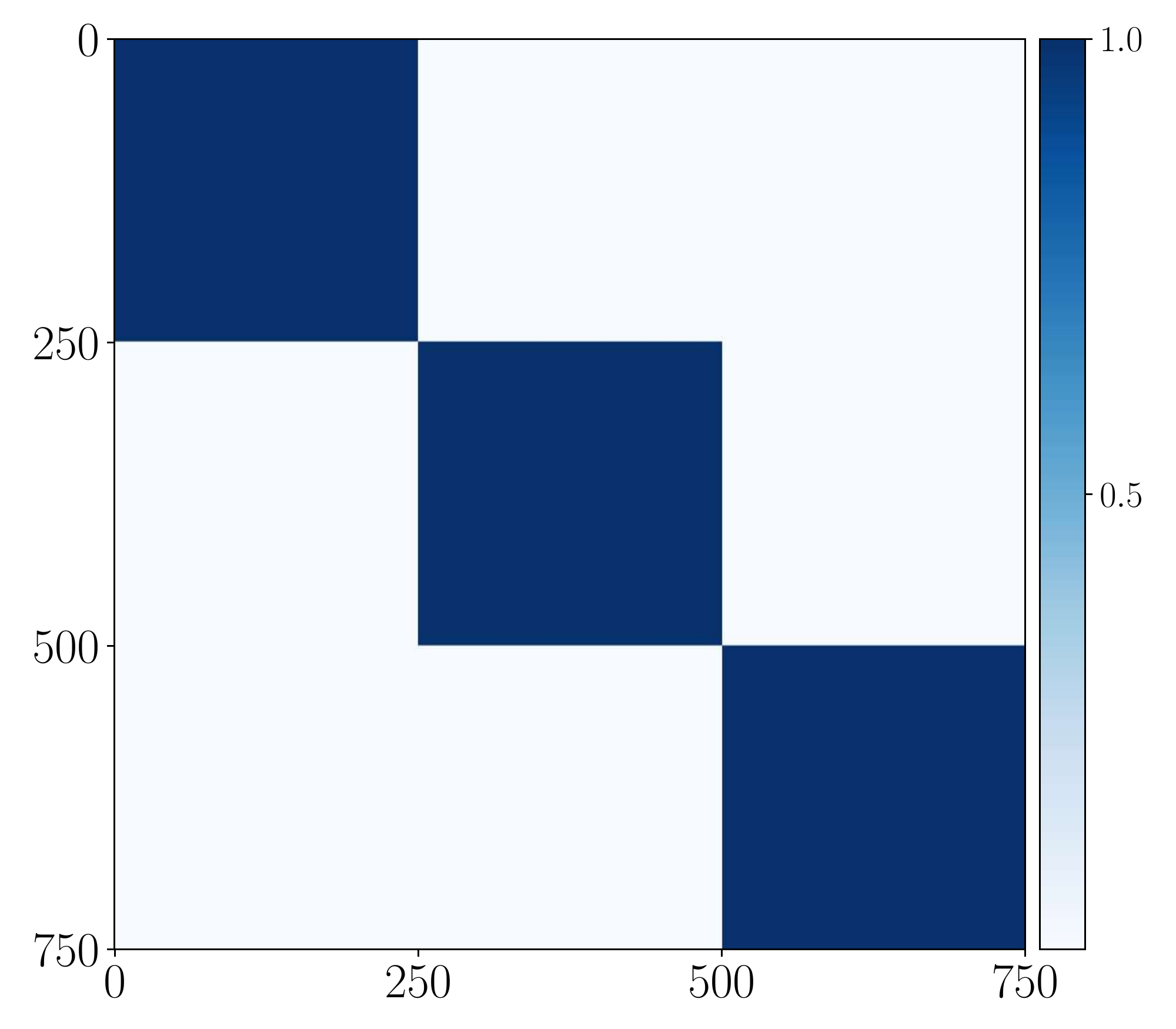}
        \end{subfigure}
        \caption{$\Z (3D)$ (\textbf{left: }scatter plot; \textbf{right: }cosine similarity visualization)}
    \end{subfigure}
    \begin{subfigure}[b]{0.24\textwidth}
        \centering
        \includegraphics[width=\textwidth]{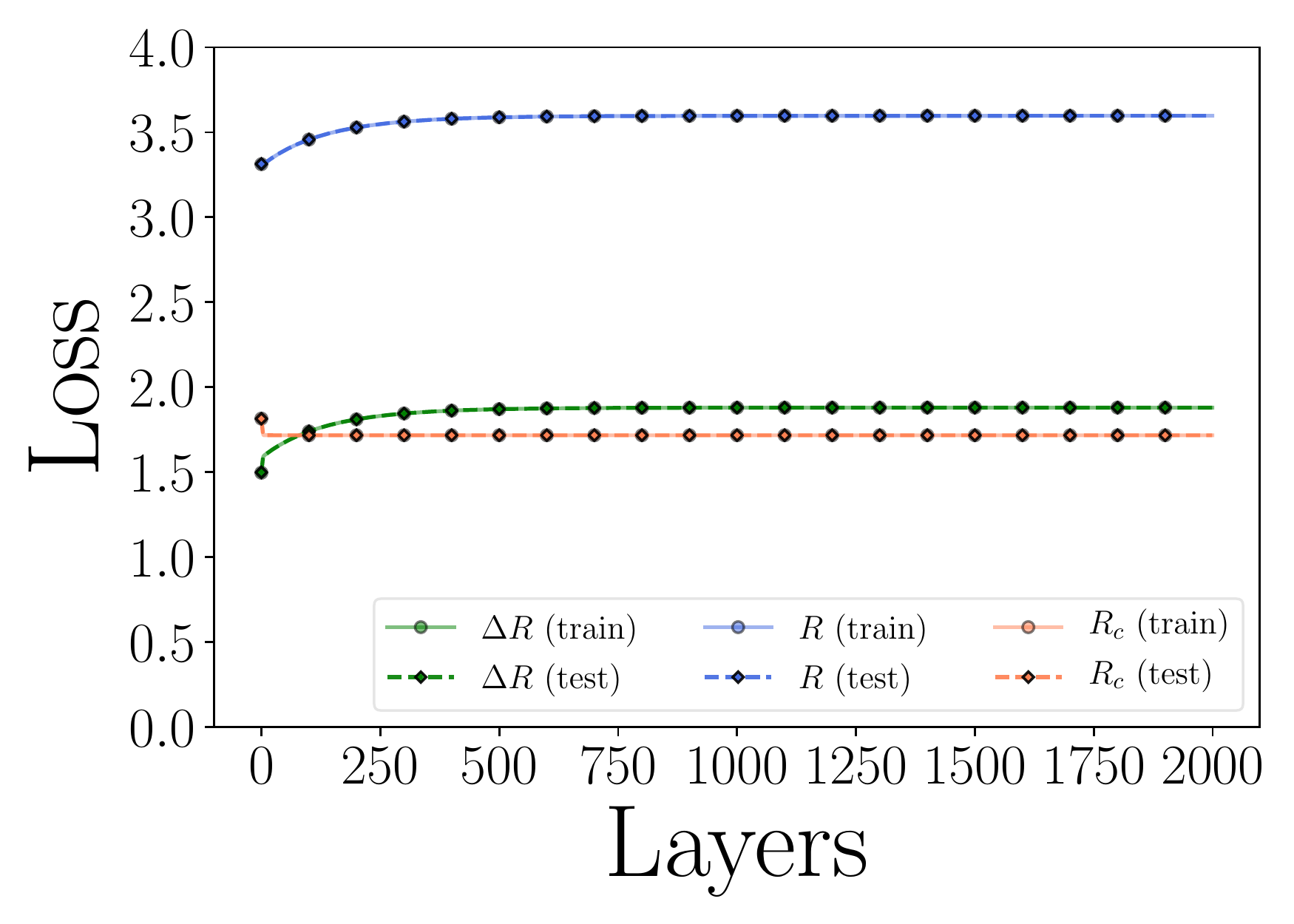}
        \caption{Loss}
    \end{subfigure}
    \caption{Learning mixture of Gaussians in $\mathbb{S}^1$ and $\mathbb{S}^2$. For $\mathbb{S}^2$, we set $\sigma_1 = \sigma_2 = \sigma_3 = 0.1$. }
    \label{fig:appendix-guassian-exp2}
\end{figure}

\begin{figure}[ht]
    \centering
    \begin{subfigure}[b]{0.49\textwidth}
        \begin{subfigure}[b]{0.49\textwidth}
            \centering
            \includegraphics[width=\textwidth]{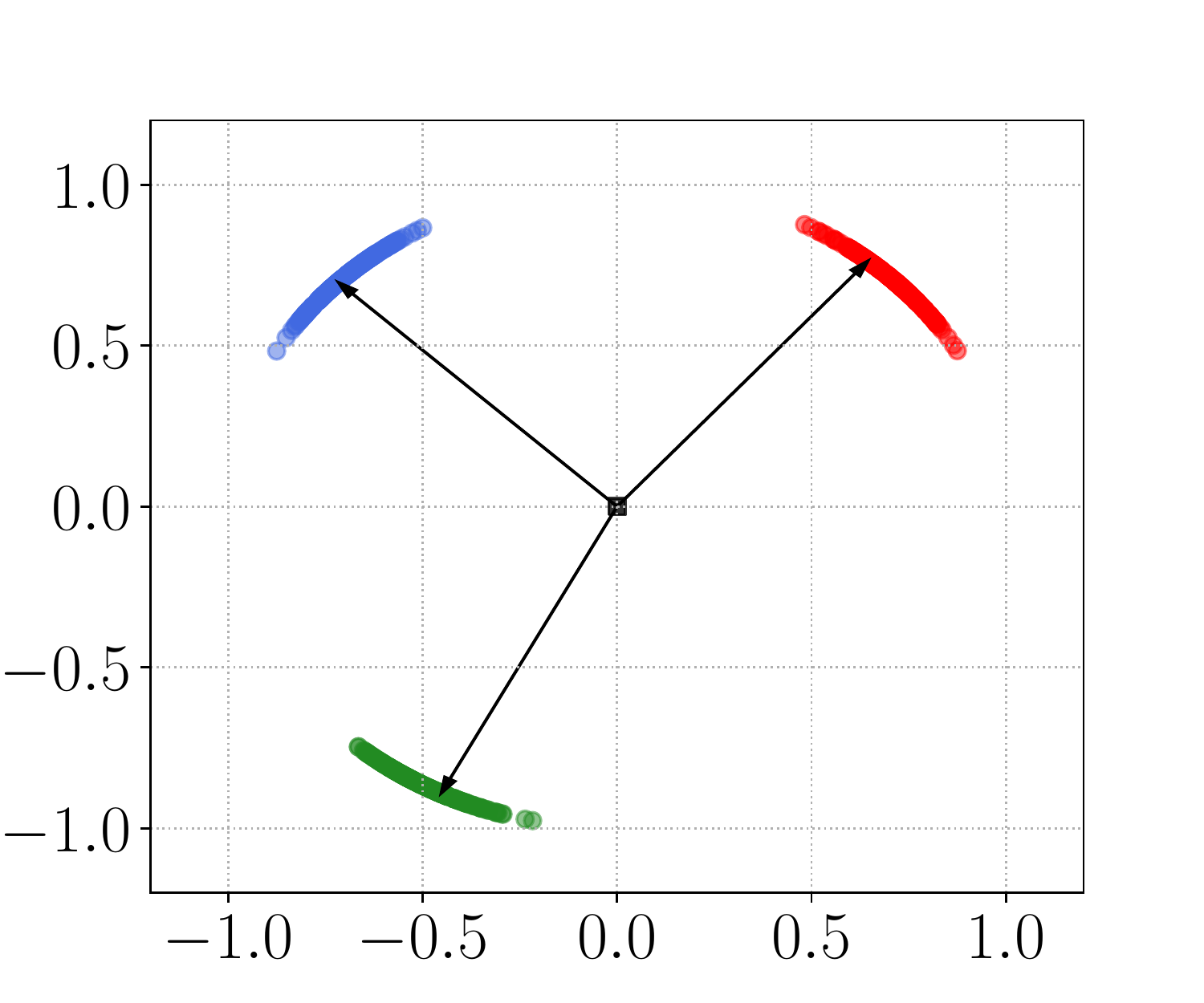}
        \end{subfigure}
    \begin{subfigure}[b]{0.49\textwidth}
            \centering
            \includegraphics[width=\textwidth]{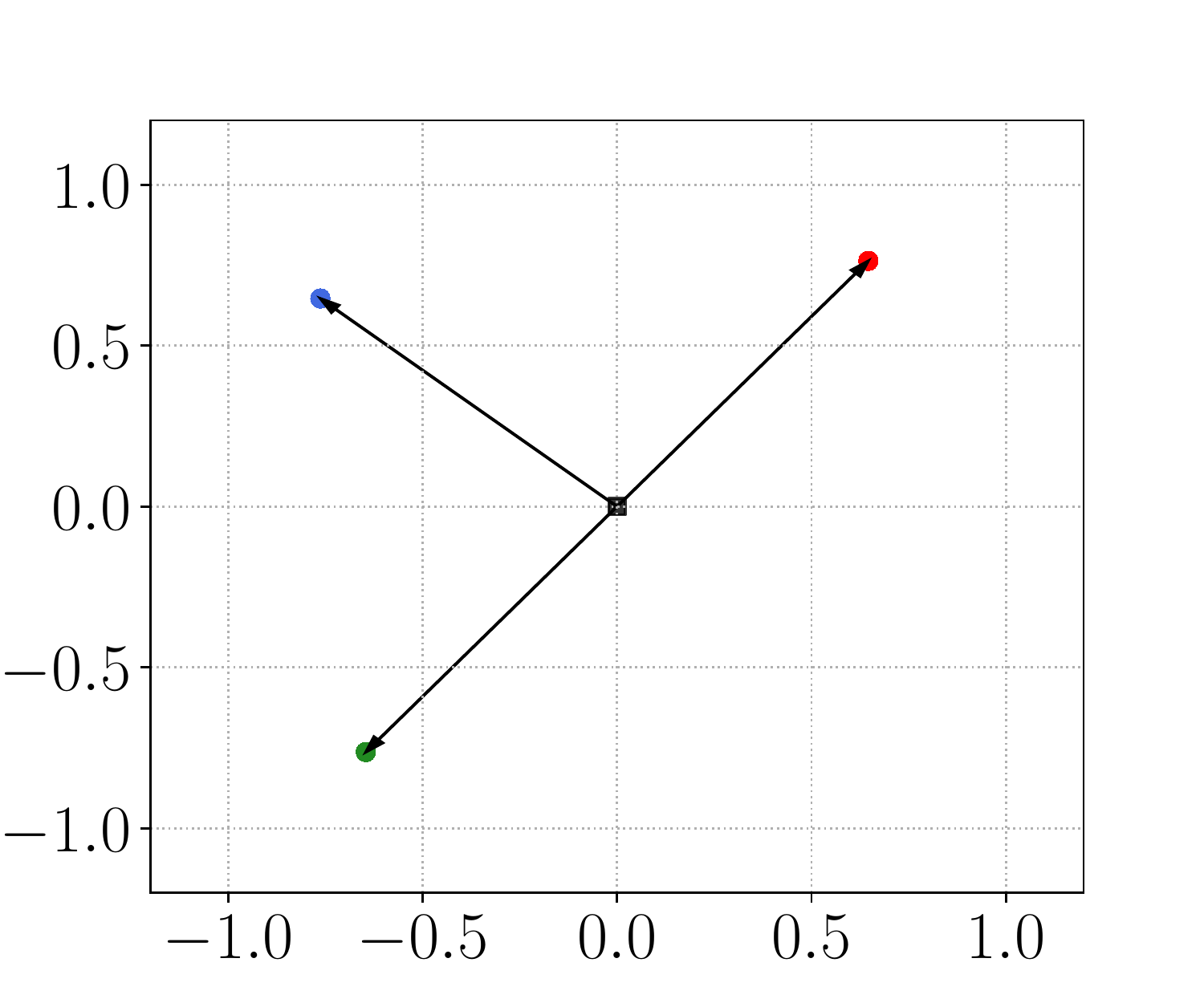}
        \end{subfigure}
    \caption{3 classes. (\textbf{Left}) $\X$; (\textbf{Right}) $\Z$}.
    \end{subfigure}
    \begin{subfigure}[b]{0.49\textwidth}
        \begin{subfigure}[b]{0.49\textwidth}
            \centering
            \includegraphics[width=\textwidth]{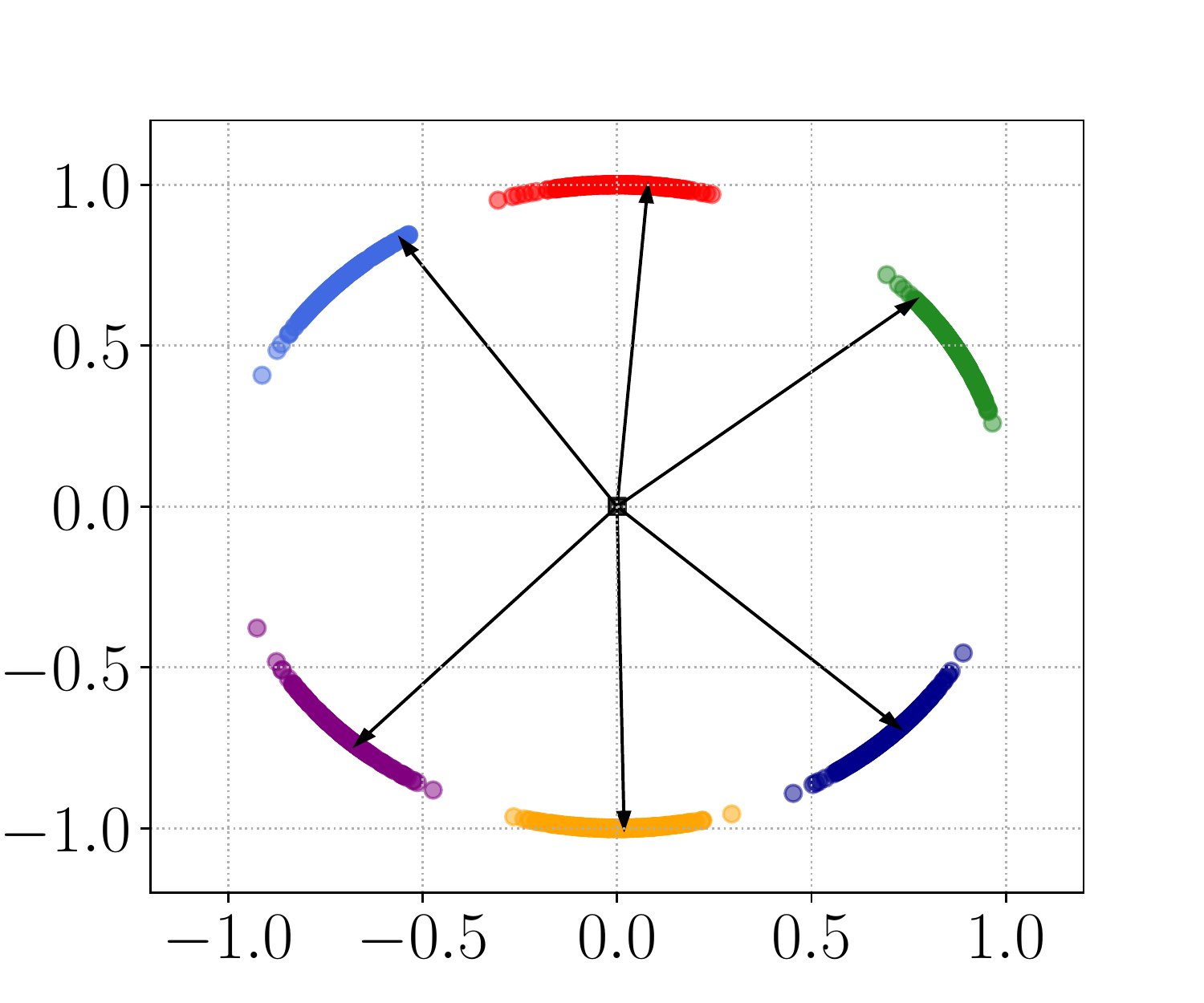}
        \end{subfigure}
    \begin{subfigure}[b]{0.49\textwidth}
            \centering
            \includegraphics[width=\textwidth]{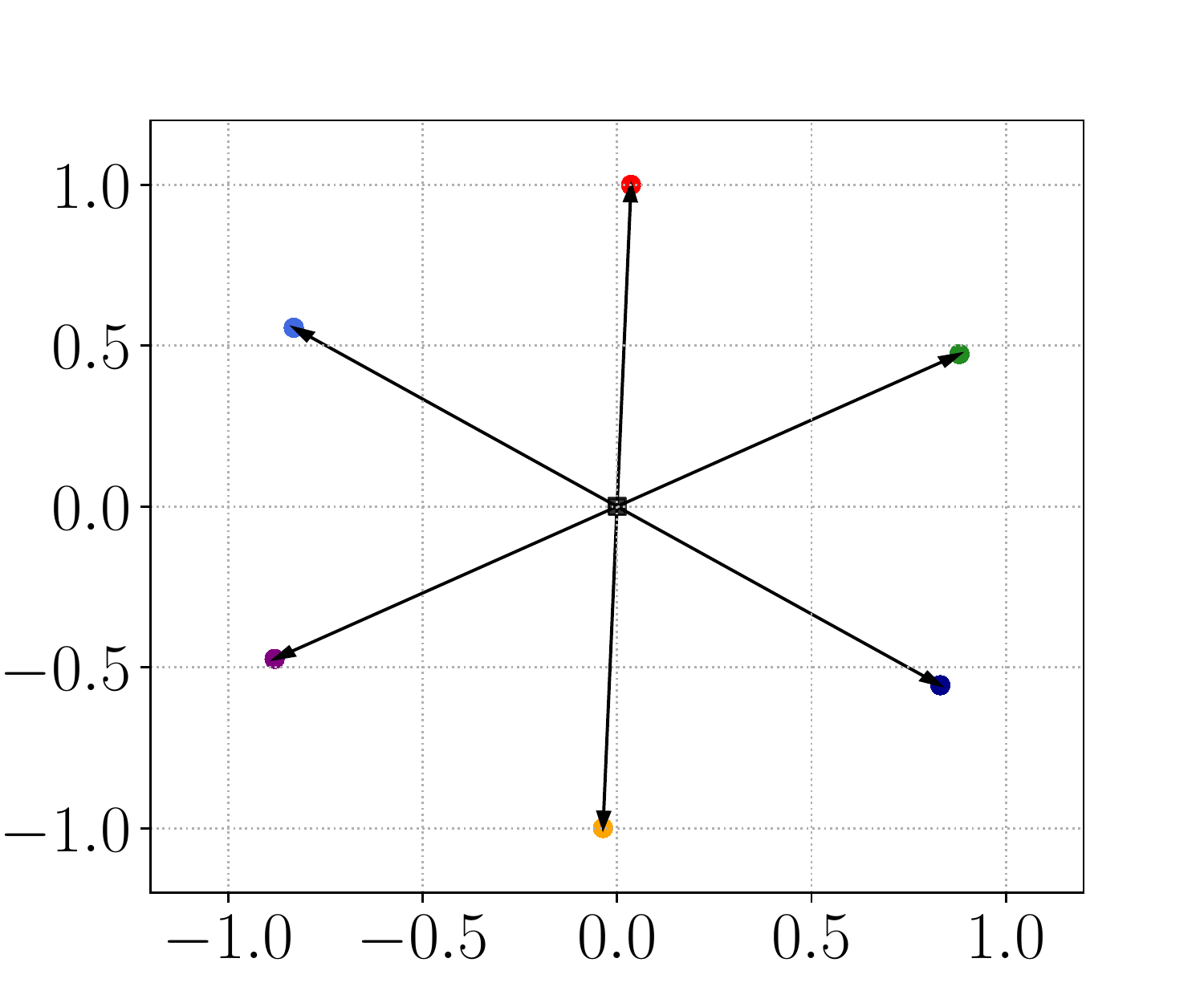}
        \end{subfigure}
        \caption{6 classes. (\textbf{Left}) $\X$; (\textbf{Right}) $\Z$}.
    \end{subfigure}
    \caption{Learning mixture of Gaussian distributions with more than 2 classes. For both cases, we use step size $\eta=0.5$ and precision $\epsilon=0.1$. For (a), we set iteration $L=2,500$; for (b), we set iteration $L=4,000$.}
    \label{fig:appendix-scatter-extra}
\end{figure}

\textbf{Additional experiments on  $\mathbb{S}^1$ with more than 2 classes.} We try to apply ReduNet to learn mixture of Gaussian distributions on $\mathbb{S}^1$ with the number of class is larger than 2. Notice that these are the cases to which the existing theory about MCR$^2$~\citep{yu2020learning} no longer applies. These experiments suggest that the MCR$^2$ still promotes between-class discriminativeness with so constructed ReduNet. In particular, the case on the left of Figure \ref{fig:appendix-scatter-extra}  indicates that the ReduNet has ``merged'' two linearly correlated clusters into one on the same line. This is consistent with the objective of rate reduction to group data as linear subspaces. 

\newpage
\subsection{Experiments on UCI datasets}
We evaluate the proposed ReduNet on some real datasets, namely the two UCI tasks~\citep{Dua2019}: \textsf{iris} and \textsf{mice}. There are 3 classes in \textsf{iris} dataset and the number of features is 4. For \textsf{mice} dataset, there are 8 classes and the number of features is 82. We randomly select 70\% data as the training data, and use the rest for evaluation.  The results are summarized in Table~\ref{table:appendix-mcr-uci}. We compare our method with logistic regression, SVM, and random forest, and we use the implementations by \textbf{\texttt{sklearn}}~\citep{pedregosa2011scikit}. From Table~\ref{table:appendix-mcr-uci}, we find that the forward-constructed ReduNet is able to achieve comparable performance with classic methods such as logistic regression, SVM, and random forest.

\begin{table}[ht]
\begin{center}
\caption{\small Performance (Accuracy) on \textsf{iris} and \textsf{mice} of the UCI datasets.}
\label{table:appendix-mcr-uci}
\begin{small}
\begin{sc}
\begin{tabular}{l | c c c c }
\toprule
& ReduNet & Logistic Regression & SVM & Random Forest  \\
\midrule
\textsf{iris} & 0.978 & 0.933 & 0.933 & 0.978 \\
\textsf{mice} & 0.972 & 0.855 & 0.975 & 0.985\\
\bottomrule
\end{tabular}
\end{sc}
\end{small}
\end{center}
\end{table}

\newpage
\subsection{Additional Experiments on Learning Shift Invariant Features}

\begin{figure}[ht]
    \centering
    \begin{subfigure}[b]{0.24\textwidth}
        \centering
        \includegraphics[width=\textwidth]{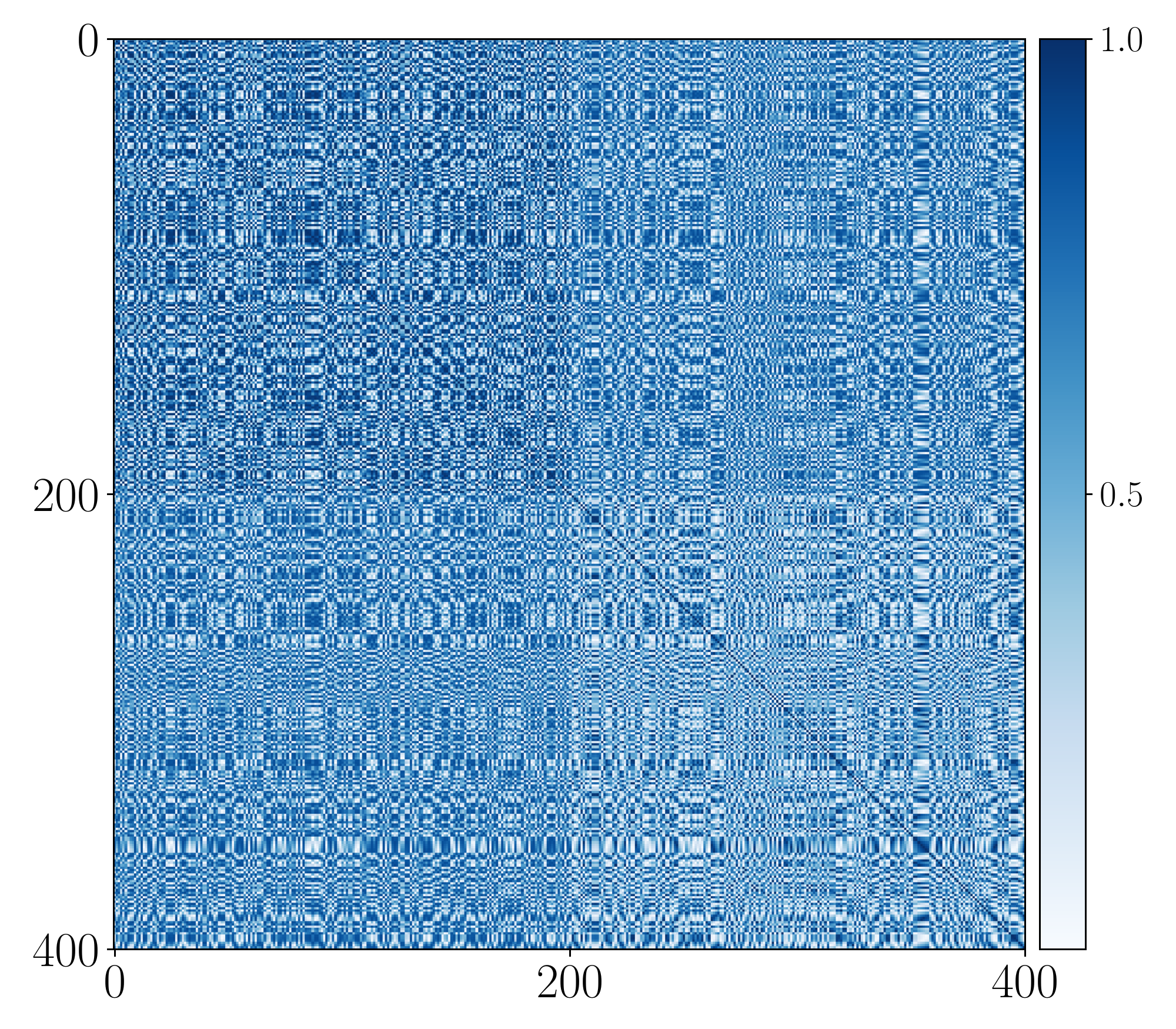}
        \caption{$\X_{\text{train}}$}
        \label{fig:sinusoids-heatmaps-a}
    \end{subfigure}
    \begin{subfigure}[b]{0.24\textwidth}
        \centering
        \includegraphics[width=\textwidth]{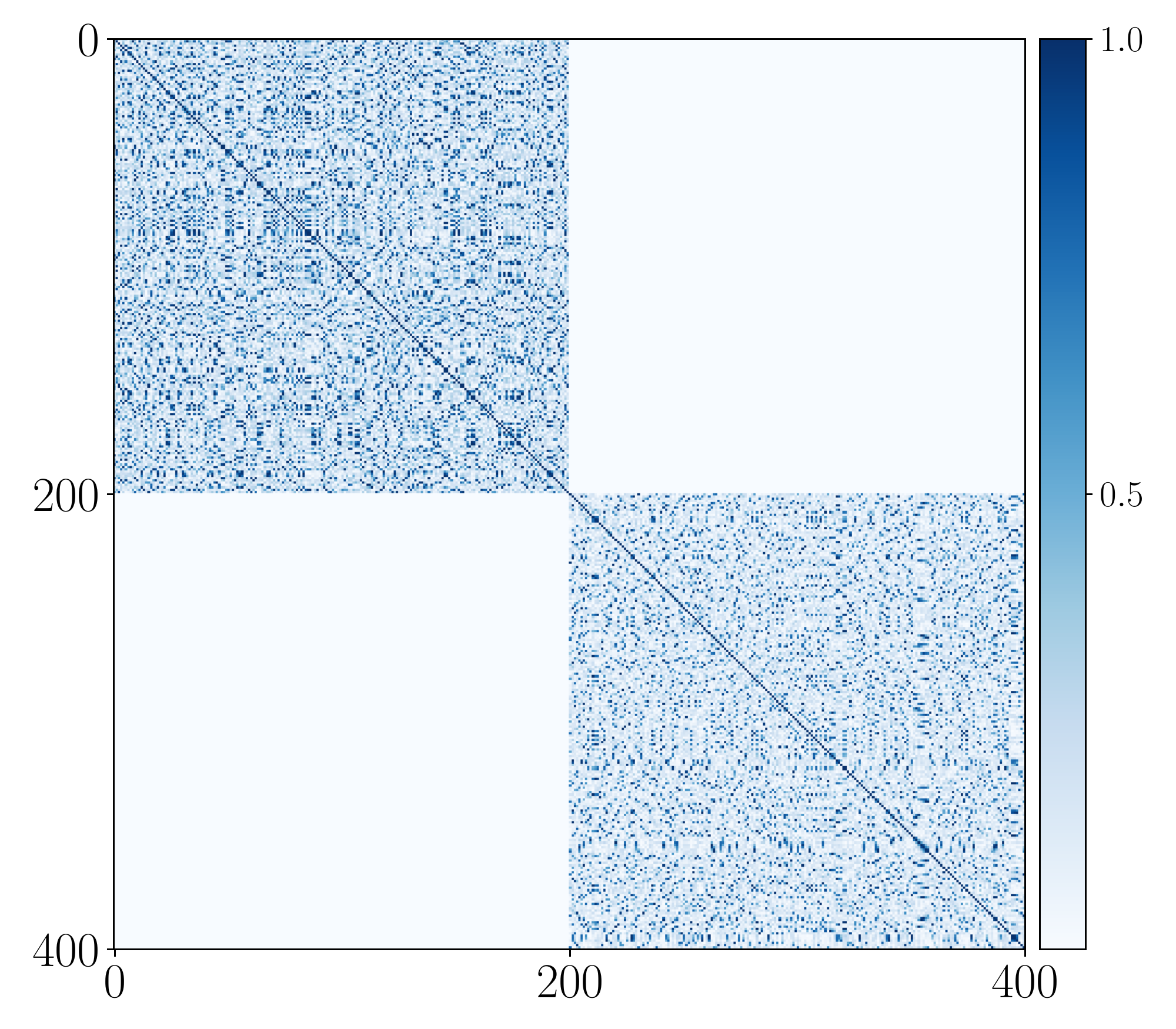}
        \caption{$\Z_{\text{train}}$}
        \label{fig:sinusoids-heatmaps-b}
    \end{subfigure}
    \begin{subfigure}[b]{0.24\textwidth}
        \centering        \includegraphics[width=\textwidth]{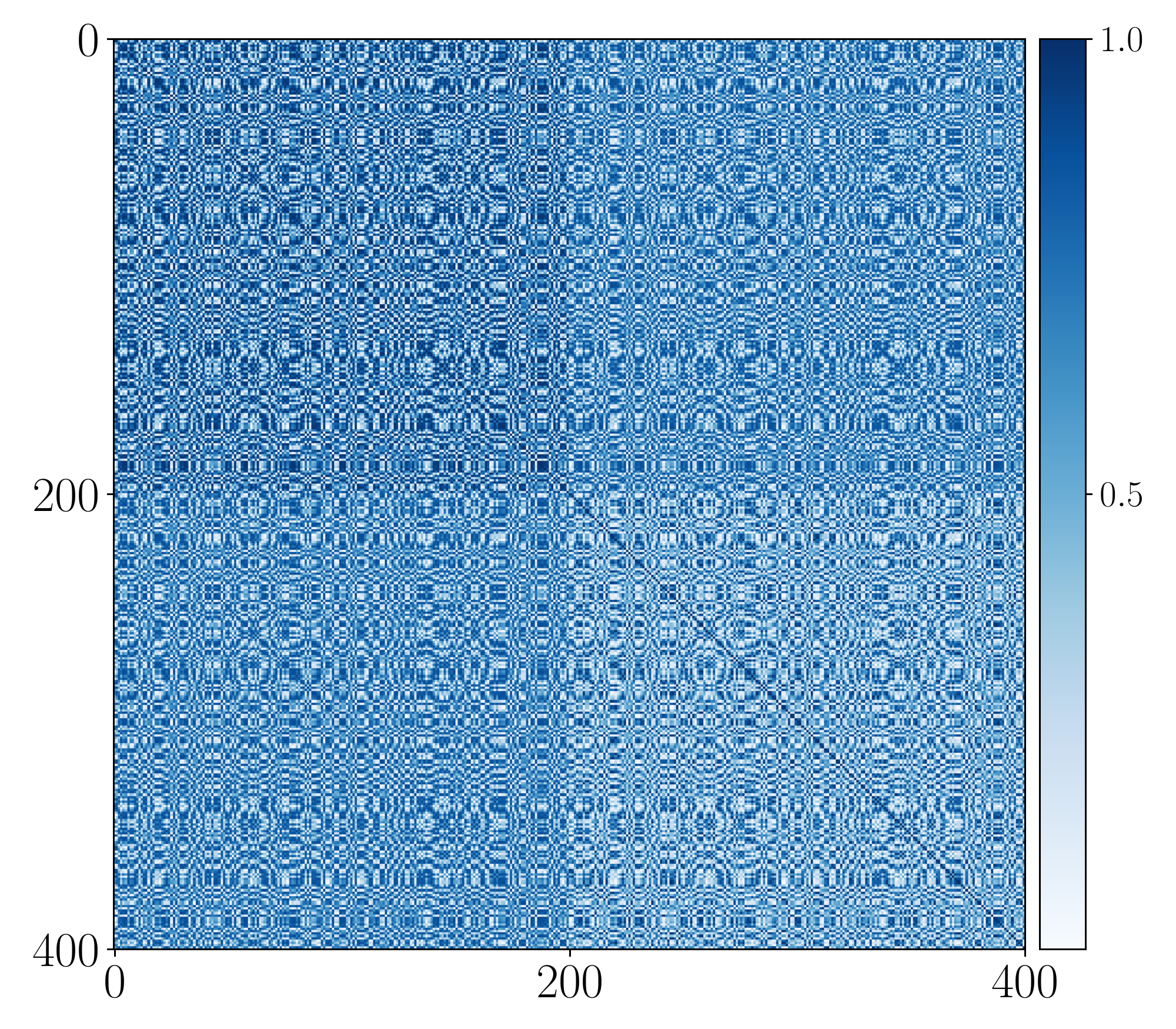}
        \caption{$\X_{\text{test}}$}
        \label{fig:sinusoids-heatmaps-c}
    \end{subfigure}
    \begin{subfigure}[b]{0.24\textwidth}
        \centering
        \includegraphics[width=\textwidth]{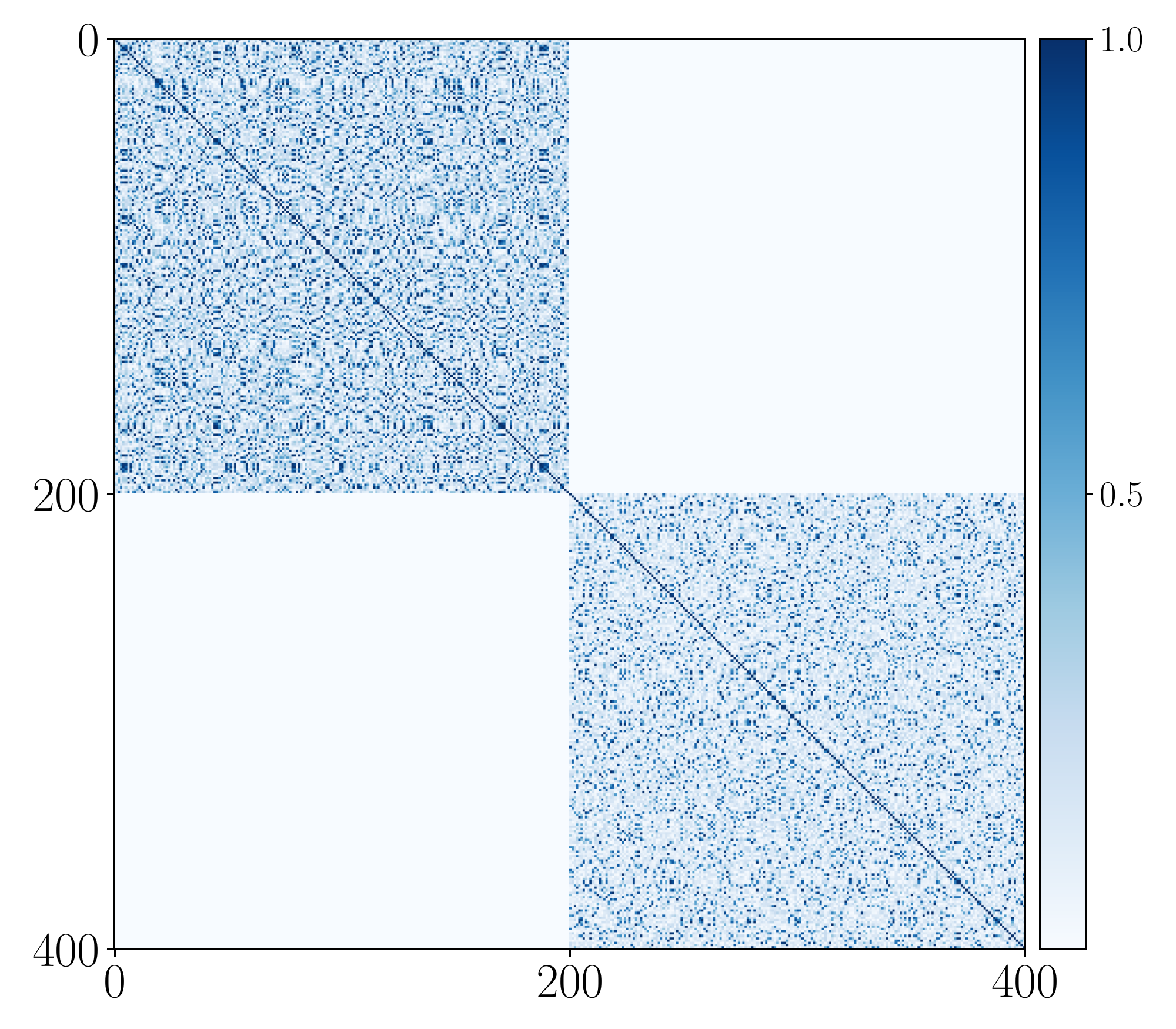}
        \caption{$\Z_{\text{test}}$}
        \label{fig:sinusoids-heatmaps-d}
    \end{subfigure}
     \caption{Cosine similarity (absolute value) of training/test data as well as training/test representations for learning 1D functions.}
    \label{fig:appendix-sinusoids-heatmaps}
\end{figure}

\begin{figure}[ht]
    \centering
    \begin{subfigure}[b]{0.24\textwidth}
        \centering
        \includegraphics[width=\textwidth]{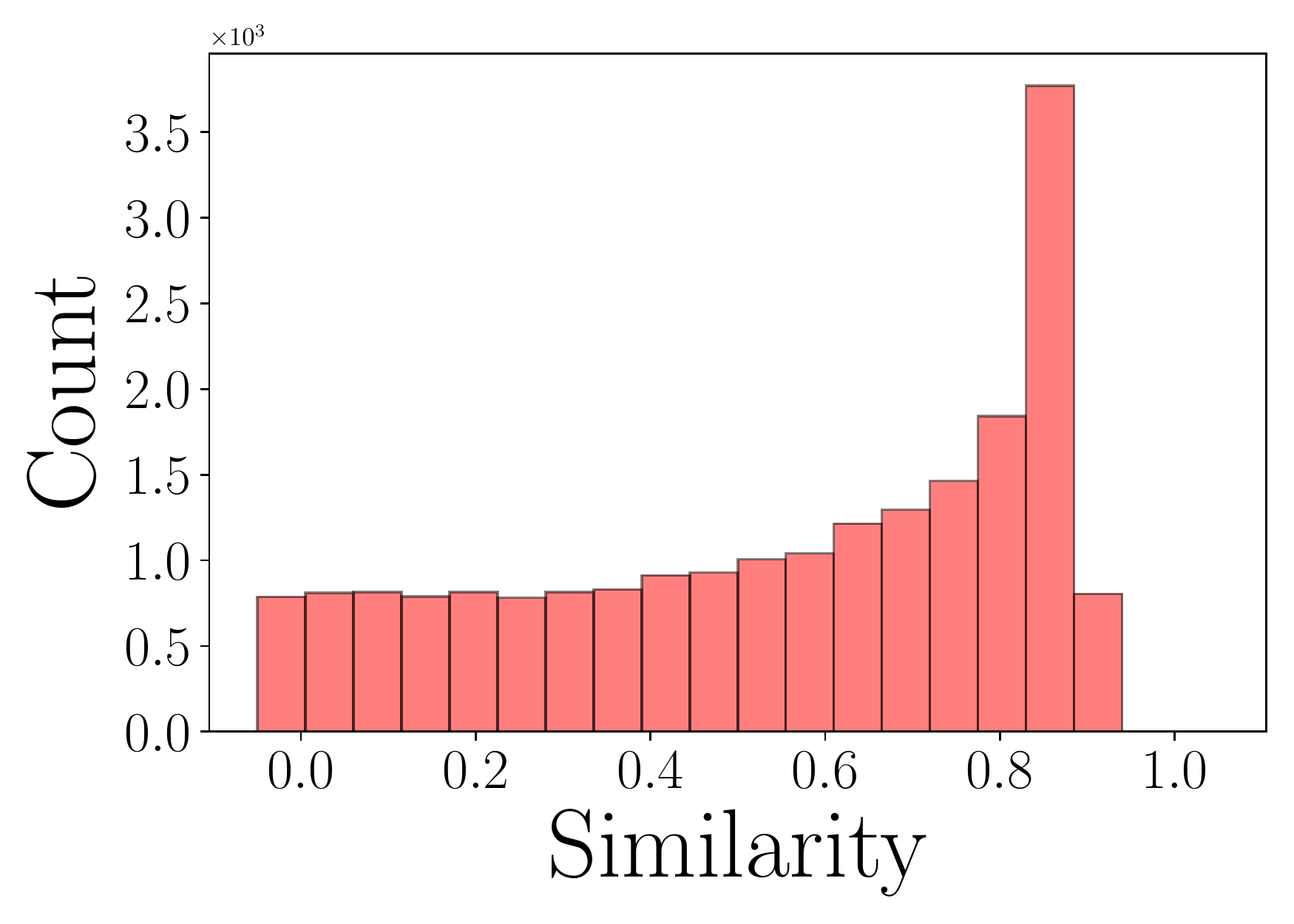}
        \caption{$\X_{1}${\scriptsize{(train)}} vs. $\X_{2}${\scriptsize{(test)}}}
    \end{subfigure}
    \begin{subfigure}[b]{0.24\textwidth}
        \centering
        \includegraphics[width=\textwidth]{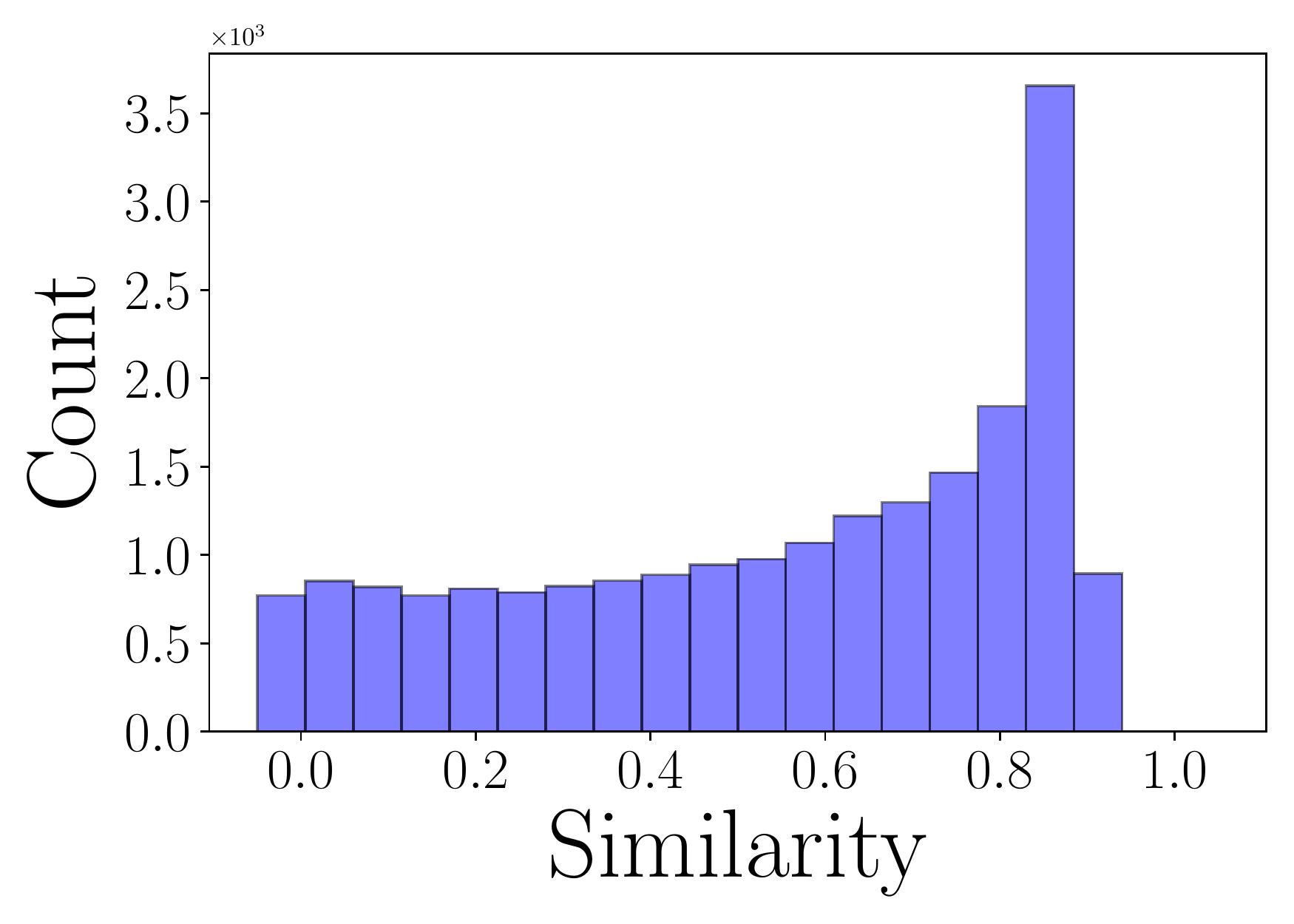}
        \caption{$\X_{2}${\scriptsize{(train)}} vs $\X_{1}${\scriptsize{(test)}}}
    \end{subfigure}
    \begin{subfigure}[b]{0.24\textwidth}
        \centering
        \includegraphics[width=\textwidth]{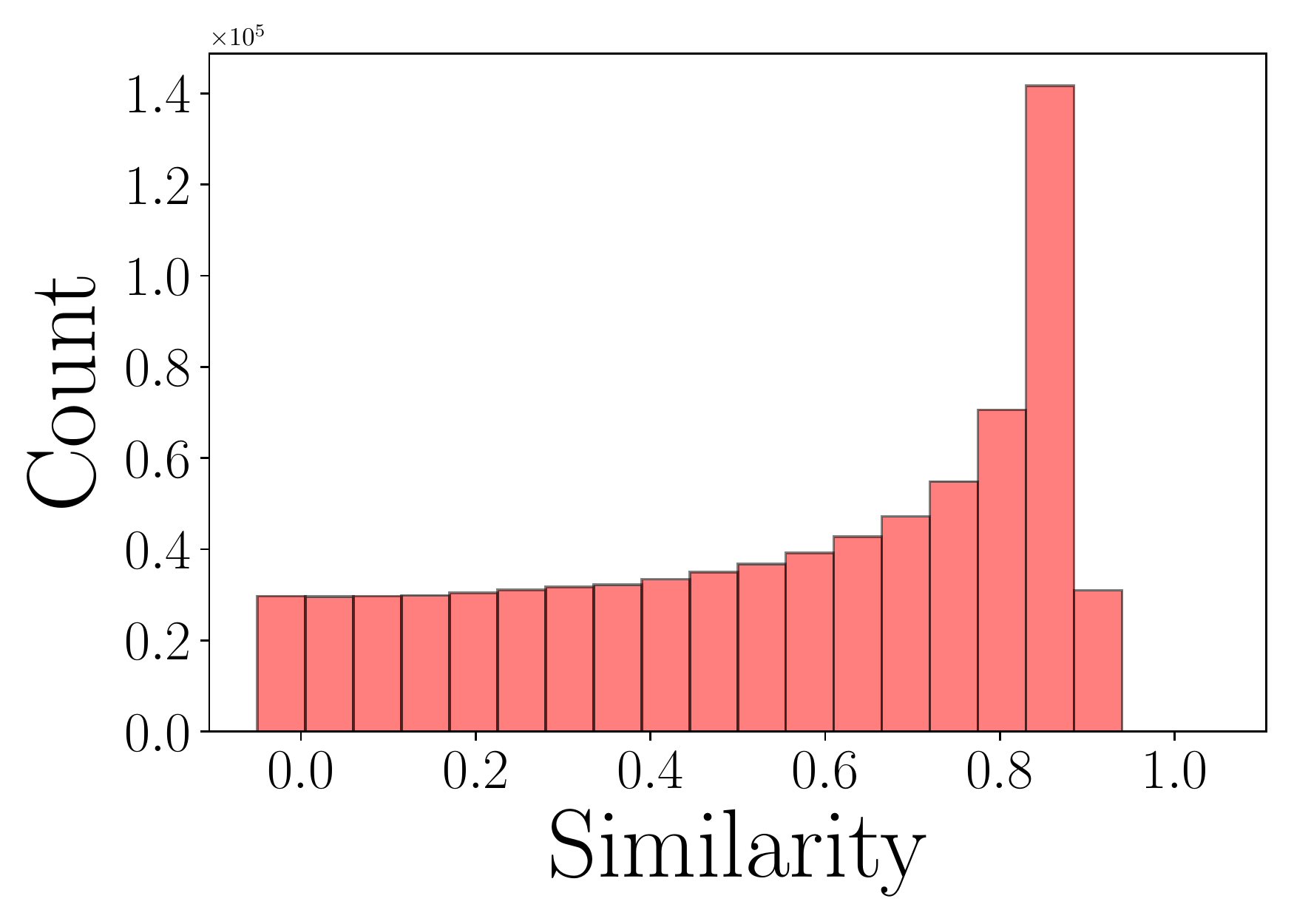}
        \caption{$\X_{1}${\scriptsize{(train)}} vs $\bar{\X}_{2}${\scriptsize{(test)}}}
    \end{subfigure}
    \begin{subfigure}[b]{0.24\textwidth}
        \centering
        \includegraphics[width=\textwidth]{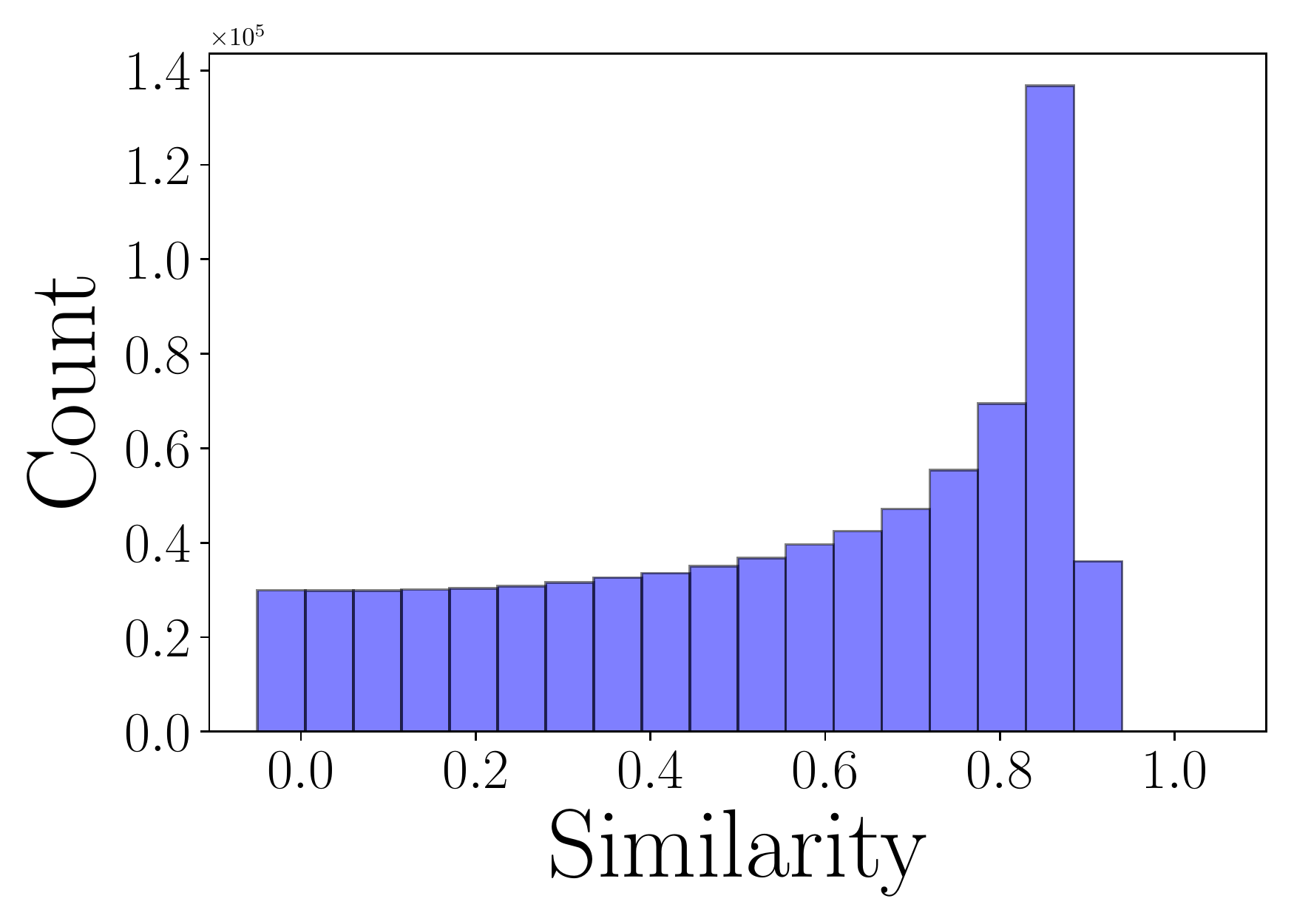}
        \caption{$\X_{2}${\scriptsize{(train)}} vs $\bar{\X}_{1}${\scriptsize{(test)}}}
    \end{subfigure}
    \begin{subfigure}[b]{0.24\textwidth}
        \centering
        \includegraphics[width=\textwidth]{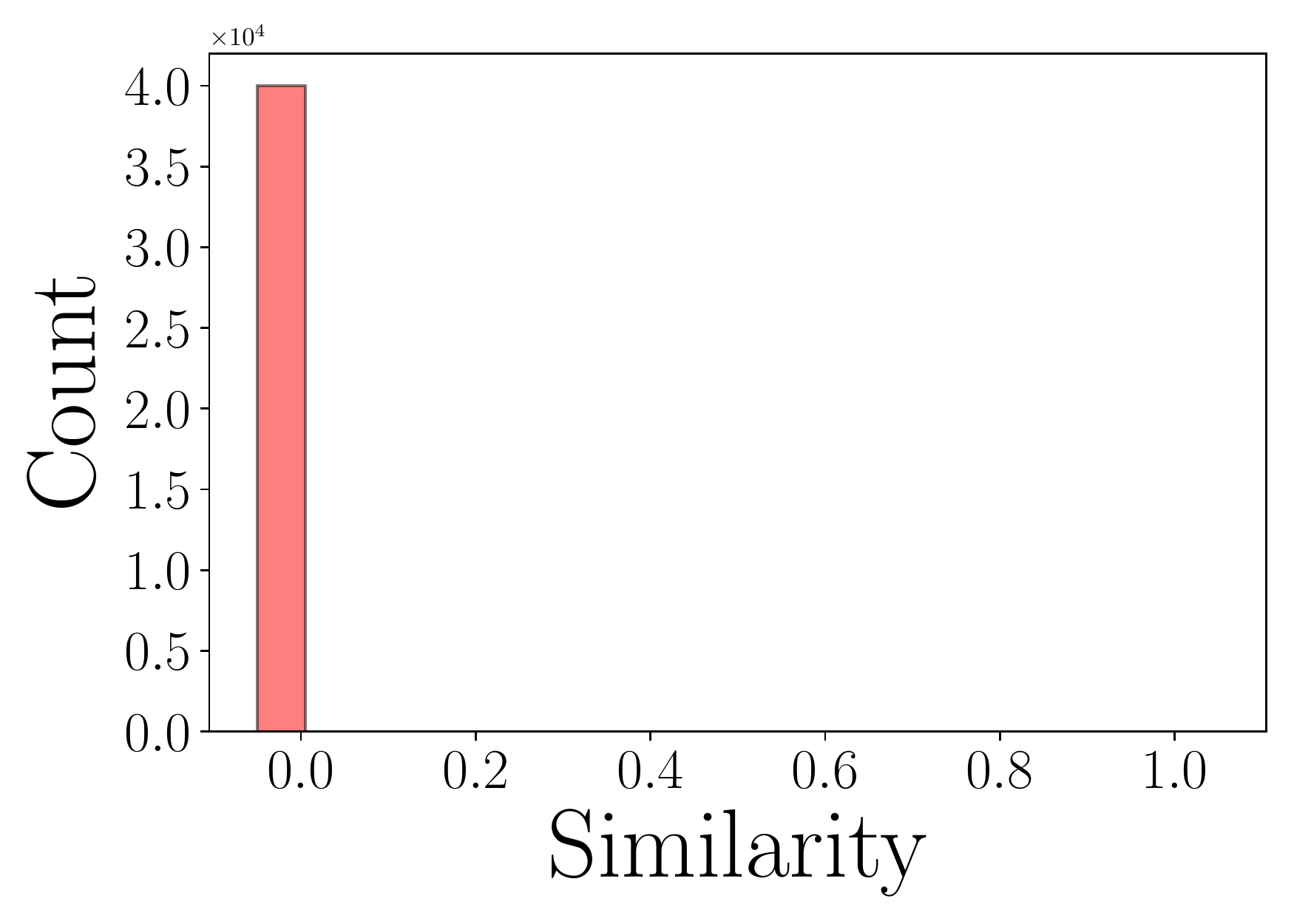}
        \caption{$\Z_{1}${\scriptsize{(train)}} vs. $\Z_{2}${\scriptsize{(test)}}}
    \end{subfigure}
    \begin{subfigure}[b]{0.24\textwidth}
        \centering
        \includegraphics[width=\textwidth]{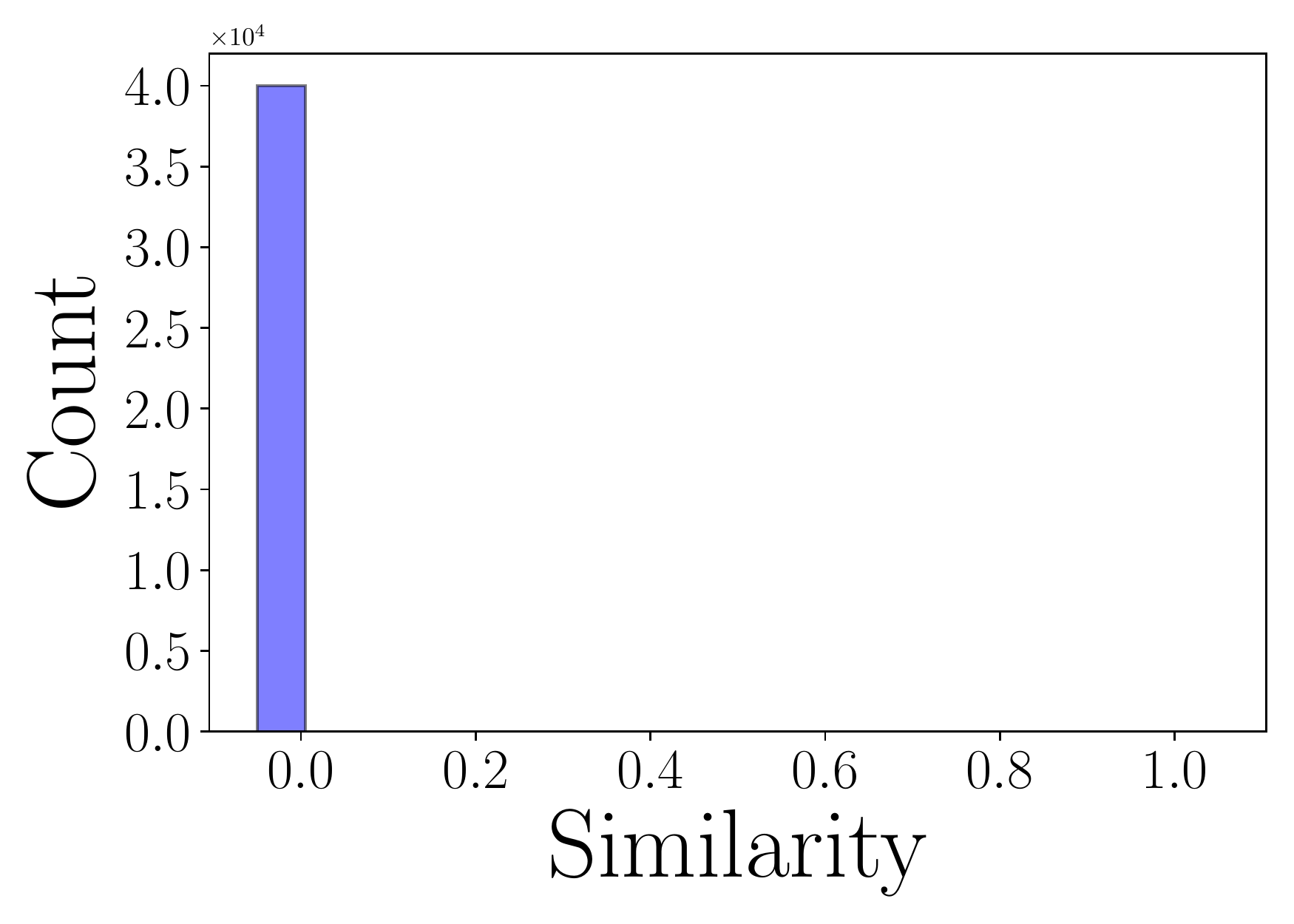}
        \caption{$\Z_{2}${\scriptsize{(train)}} vs $\Z_{1}${\scriptsize{(test)}}}
    \end{subfigure}
    \begin{subfigure}[b]{0.24\textwidth}
        \centering
        \includegraphics[width=\textwidth]{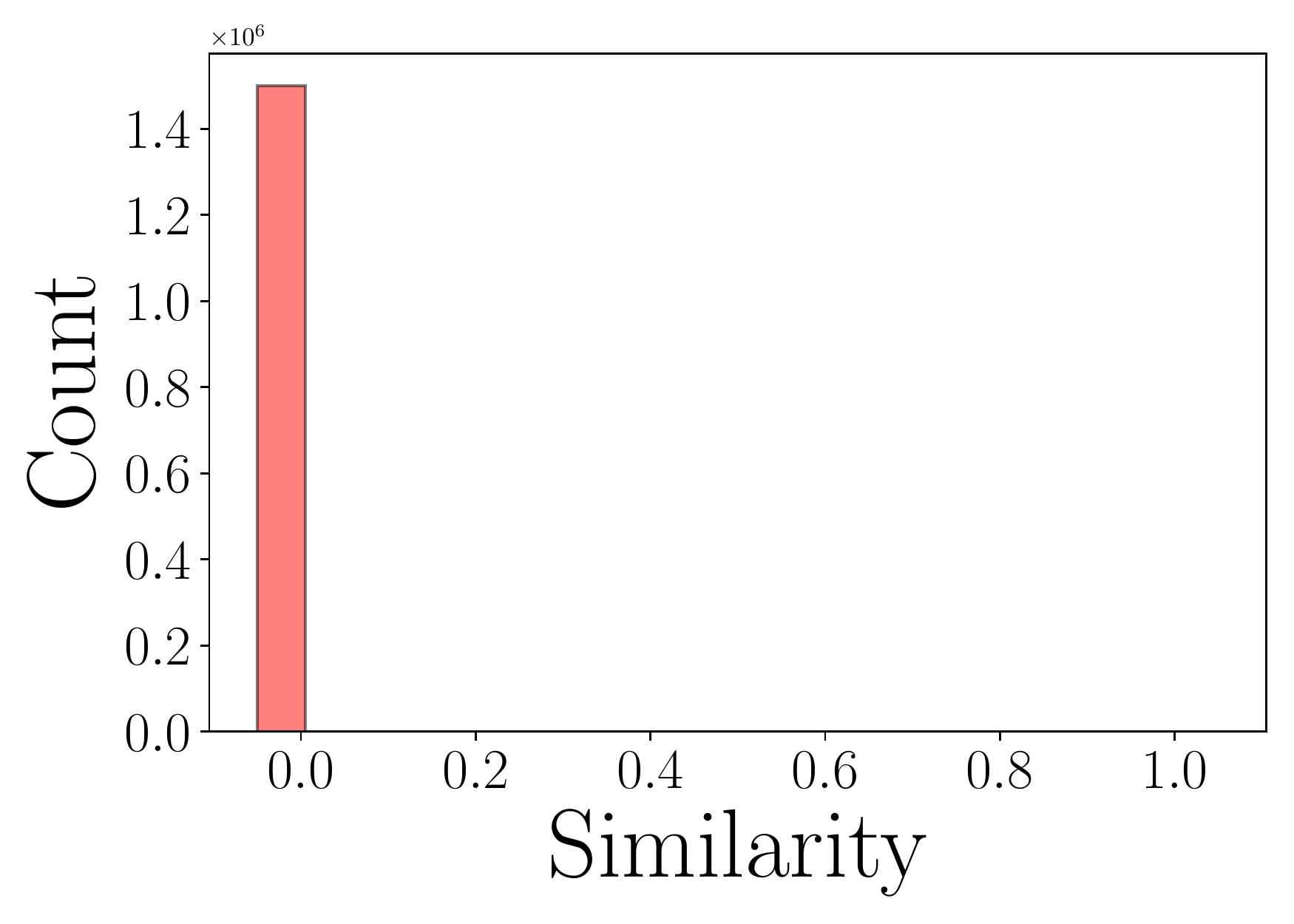}
        \caption{$\Z_{1}${\scriptsize{(train)}} vs $\bar{\Z}_{2}${\scriptsize{(test)}}}
    \end{subfigure}
    \begin{subfigure}[b]{0.24\textwidth}
        \centering
        \includegraphics[width=\textwidth]{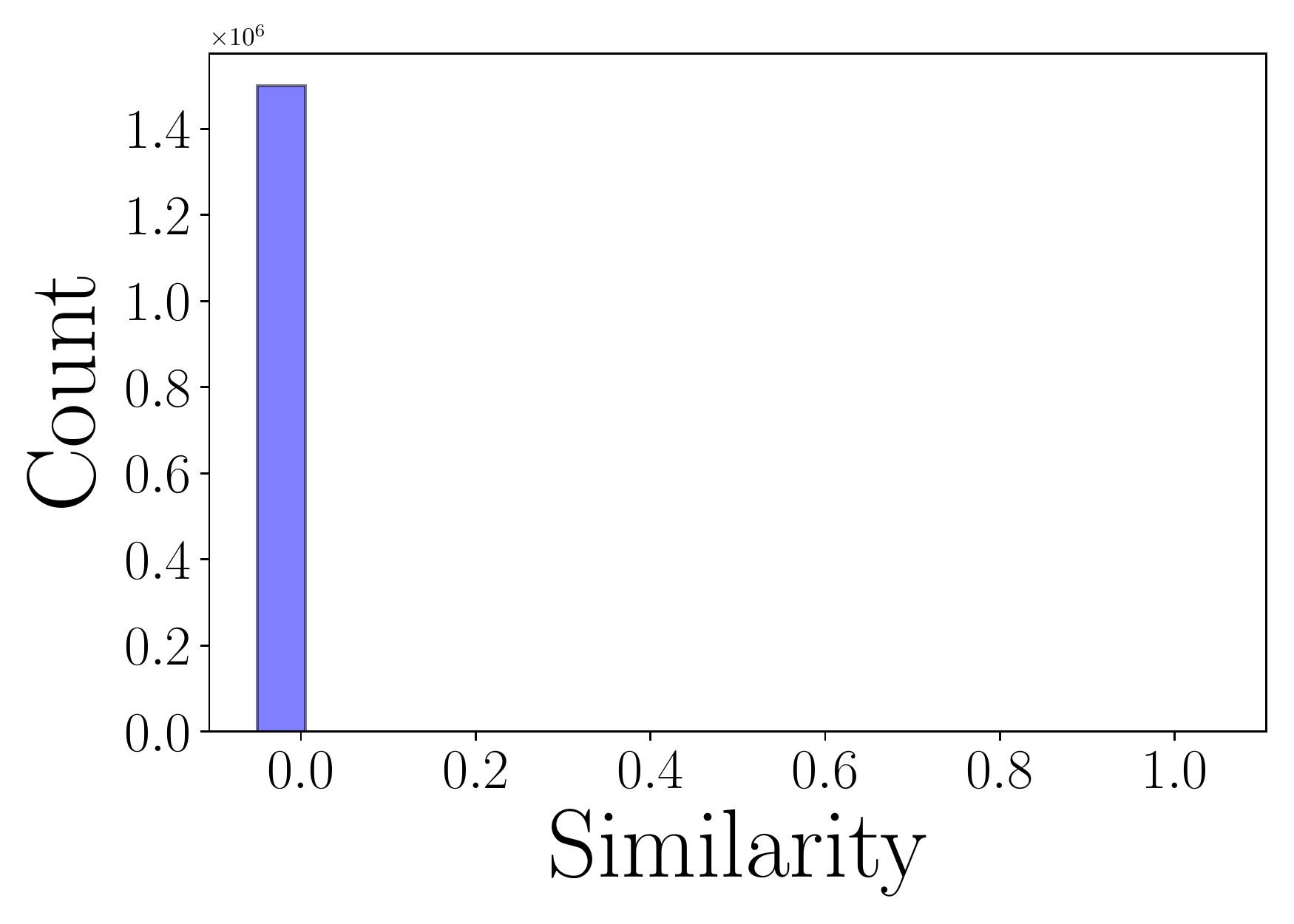}
        \caption{$\Z_{2}${\scriptsize{(train)}} vs $\bar{\Z}_{1}${\scriptsize{(test)}}}
    \end{subfigure}
    \caption{Histogram of cosine similarity between pairs sampled from different classes for learning 1D function. The histogram of cosine similarity between training data $\bm \X_{c}$ as well as representations $\bm {\Z}_{c}$ vs. testing (shifted) data $\bar{\X}_{c^{\prime}}$ as well as (shifted) representations $\bar{\Z}_{c^{\prime}}$, where we let $c$ denote the class index  and $c\neq c^{\prime}$.}
    \label{fig:appendix-cosine-hist-1d}
\end{figure}

We provide additional experiments for \textit{Learning Shift Invariant Features} in \textsection\ref{sec:experiments}. The code for sampling from $h_1(t) = \textsf{sin}(t) + \epsilon$ and $h_2(t) = \textsf{sign}(\textsf{sin}(t)) + \epsilon$ is described in Algorithm~\ref{alg:appendix-generate-1d}, and the pseudocode for sampling from $2$ classes $\{h_1, h_2\}$ is described as follows, we sample training and test signals using the same procedure. 
\begin{tcolorbox}
\begin{footnotesize}
\begin{verbatim}
t0 = np.random.uniform(low=0, high=10*np.pi, size=samples)
x = np.linspace(t0, t0+2*np.pi, time).T
noise1 = np.random.normal(0, 0.1, size=(samples, time))
X1 = np.sin(x) + noise1
noise2 = np.random.normal(0, 0.1, size=(samples, time))
X2 = np.sign(np.sin(x)) + noise2
data = np.vstack([X1, X2])
labels = np.hstack([np.ones(samples)*1,
                    np.ones(samples)*2]).astype(np.int32)
\end{verbatim}
\end{footnotesize}
\label{code:sampling-sinusoid}
\end{tcolorbox}
We also provide cosine similarities between samples in Figure~\ref{fig:appendix-sinusoids-heatmaps}. We visualize the cosine similarities for the input $\X_{\text{train}}, \X_{\text{test}}$ as well as the learned representations $\Z_{\text{train}}, \Z_{\text{test}}$. The cosine similarity between sample pairs selected from different classes are shown in  Figure~\ref{fig:appendix-cosine-hist-1d}. We can observe that the original data is not orthogonal w.r.t. different classes, and the the ReduNet is able to learn discriminative (orthogonal) representations.

\begin{algorithm}
    \caption{Pseudocode for sampling signals from $1D$ functions}
    \label{alg:sampling-sinusoid}
	\begin{algorithmic}[1]
	    \REQUIRE Number of samples $m$, number of classes $k$, number of features $n$, function $\{h_1, \ldots, h_k\}$.
	    \FOR{$j = 1, 2, \ldots, k$}
            \FOR{$i = 1, 2, \ldots, m$}
    	       \STATE $t_0 \sim \text{Uniform}[0, 10\pi]$;
    	       \STATE $\bm{t} = [t_0, t_0 + {2 \pi}/{n}, t_0 + ({2 \pi}/{n}) \cdot 2, t_0 + ({2 \pi}/{n}) \cdot 3, \ldots, t_0 + ({2 \pi}/{n}) \cdot (n-1)]$;
    	       \STATE $\bm x_{j}^i = h_j(\bm{t}) + \epsilon$ \emph{\# broadcast over vector $\bm{t}$};
    	    \ENDFOR
    	    \STATE $\X_{j}=[\bm x_{j}^1, \bm x_{j}^2, \ldots, \bm x_{j}^m]$;
        \ENDFOR
        \STATE $\X = [\X_1, \X_2, \ldots, \X_k]$;
        \STATE shuffle $\X$.
        \ENSURE outputs $\X$.
	\end{algorithmic}
	\label{alg:appendix-generate-1d}
\end{algorithm}

\begin{figure}[ht]
    \centering
    \begin{subfigure}[b]{0.4\textwidth}
        \centering
        \includegraphics[width=\textwidth]{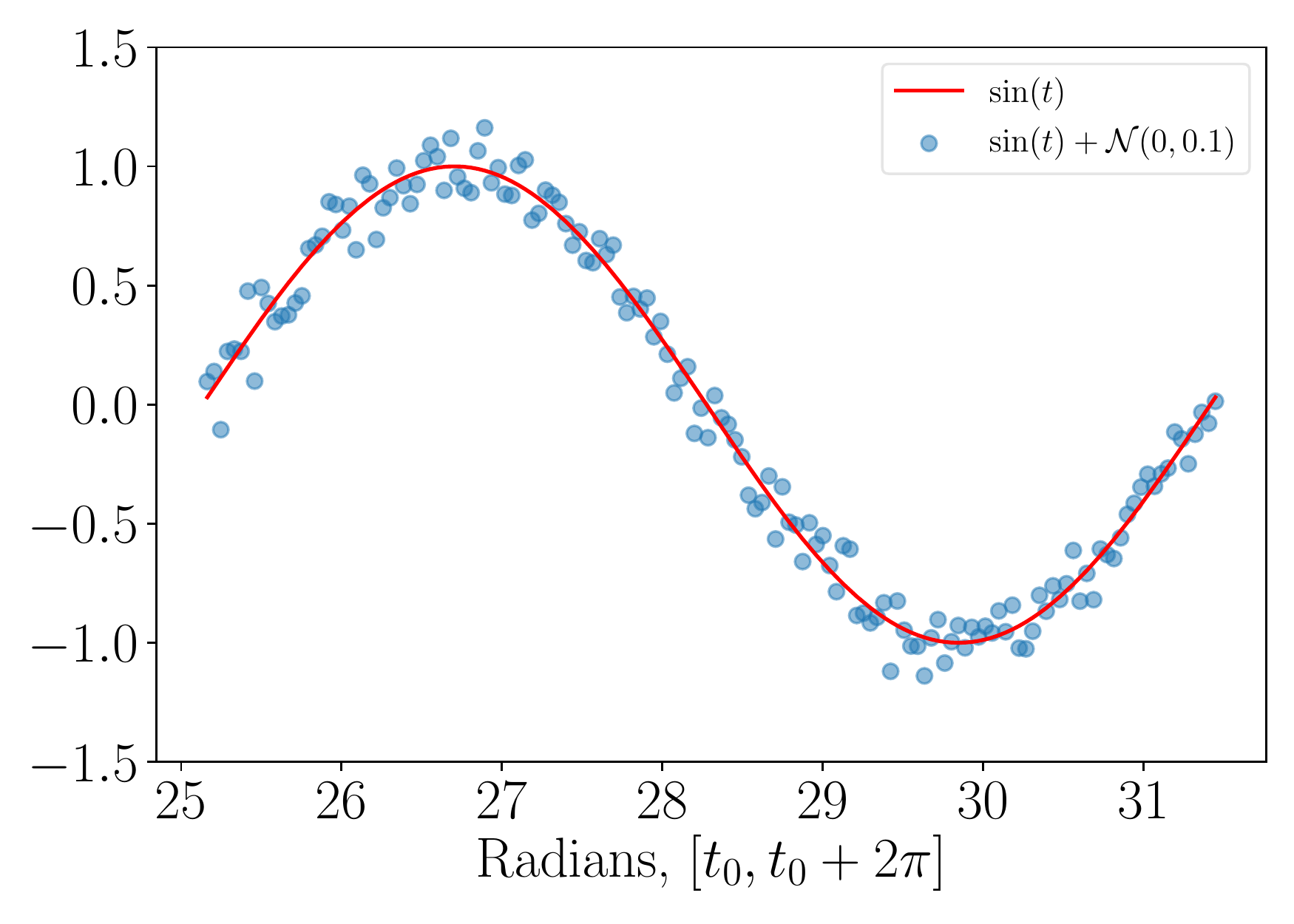}
    \end{subfigure}
    \begin{subfigure}[b]{0.4\textwidth}
        \centering
        \includegraphics[width=\textwidth]{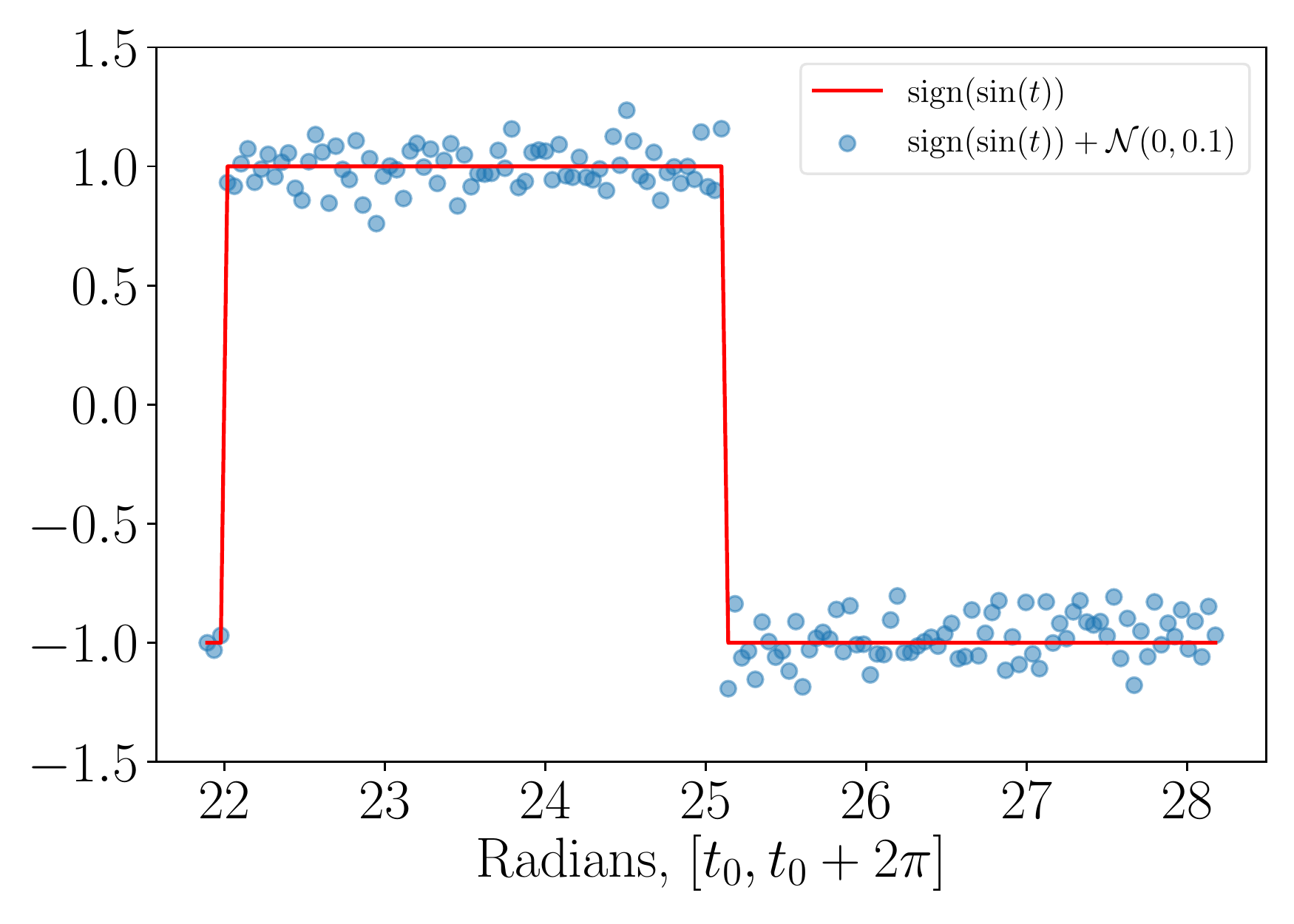}
    \end{subfigure}
    \caption{Visualization of signals in 1D. Blue dots represent the sampled signal used for training with dimension $n = 150$. Red curves represent the underlying 1D function (noiseless). \textbf{(Left)} One sample from class 1; \textbf{(Right)} One sample from class 2.}
    \label{fig:appendix-1d-function-visualize}
\end{figure}

\subsection{Additional Experiments on learning rotational invariance on MNIST}
\begin{figure}[ht]
    \centering
    \begin{subfigure}[b]{0.24\textwidth}
        \centering
        \includegraphics[width=\textwidth]{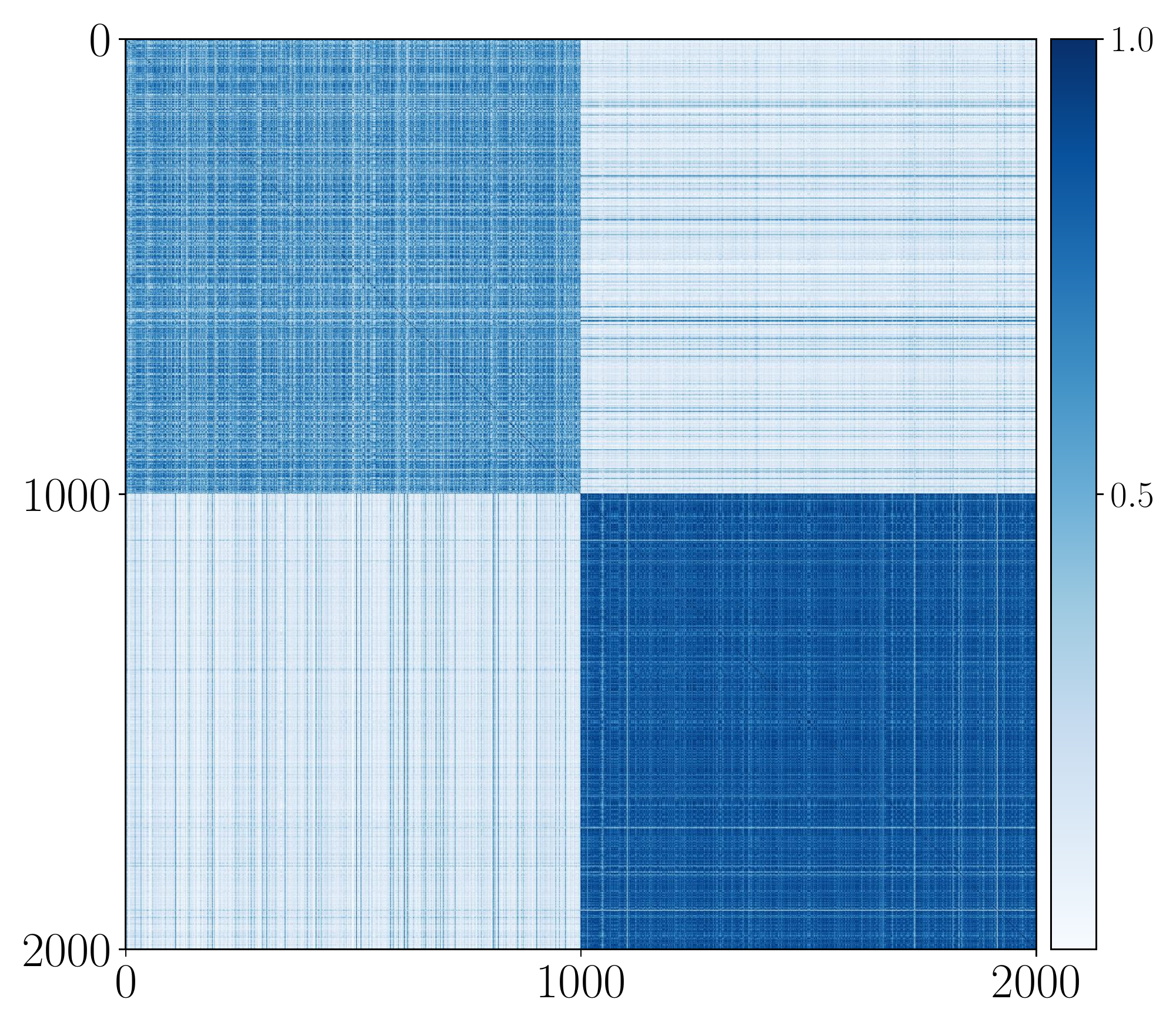}
        \caption{$\X_{\text{train}}$}
        \label{fig:mnist1d-heatmaps-a}
    \end{subfigure}
    \begin{subfigure}[b]{0.24\textwidth}
        \centering
        \includegraphics[width=\textwidth]{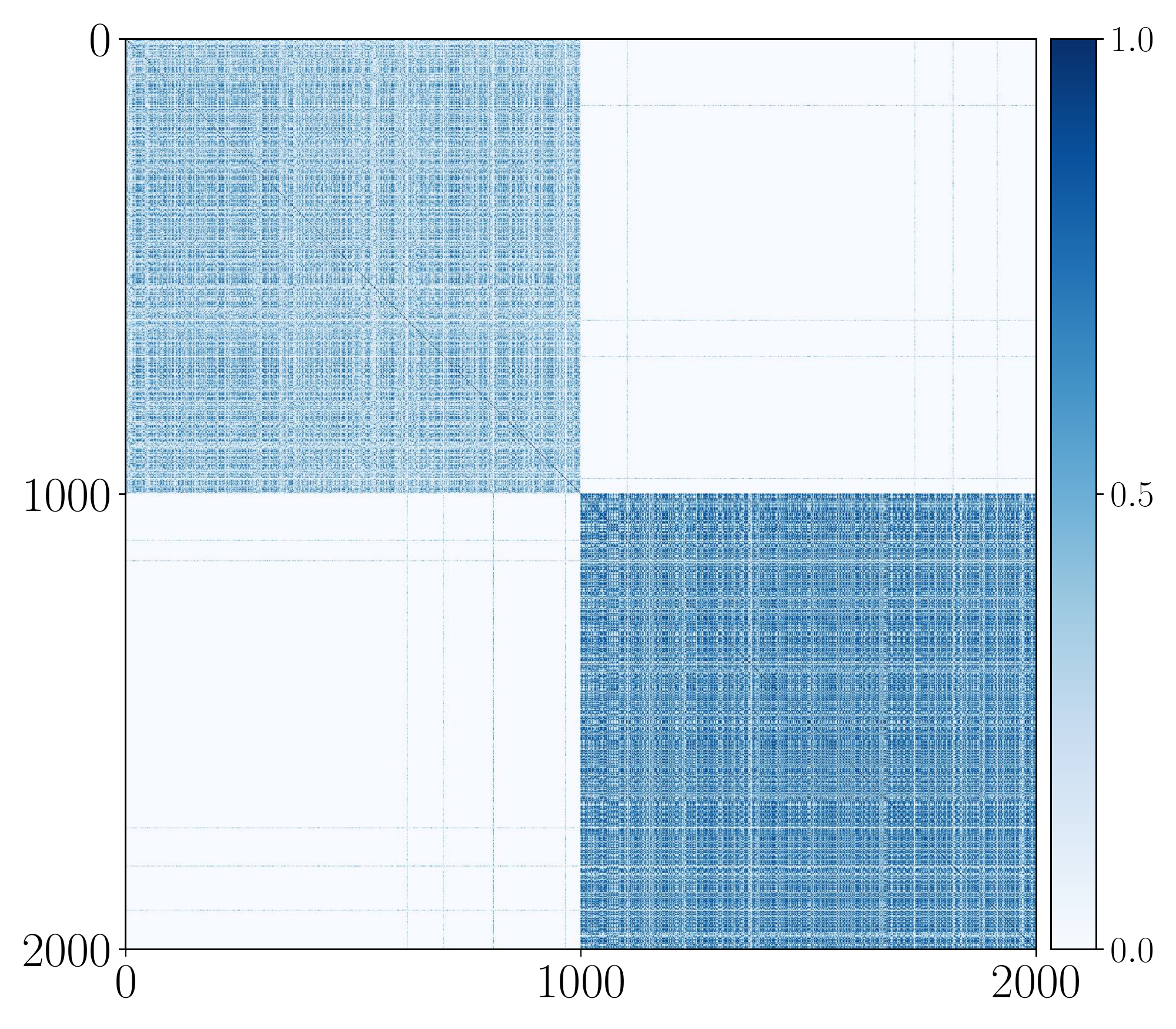}
        \caption{$\Z_{\text{train}}$}
        \label{fig:mnist1d-heatmaps-b}
    \end{subfigure}
    \begin{subfigure}[b]{0.24\textwidth}
        \centering
        \includegraphics[width=\textwidth]{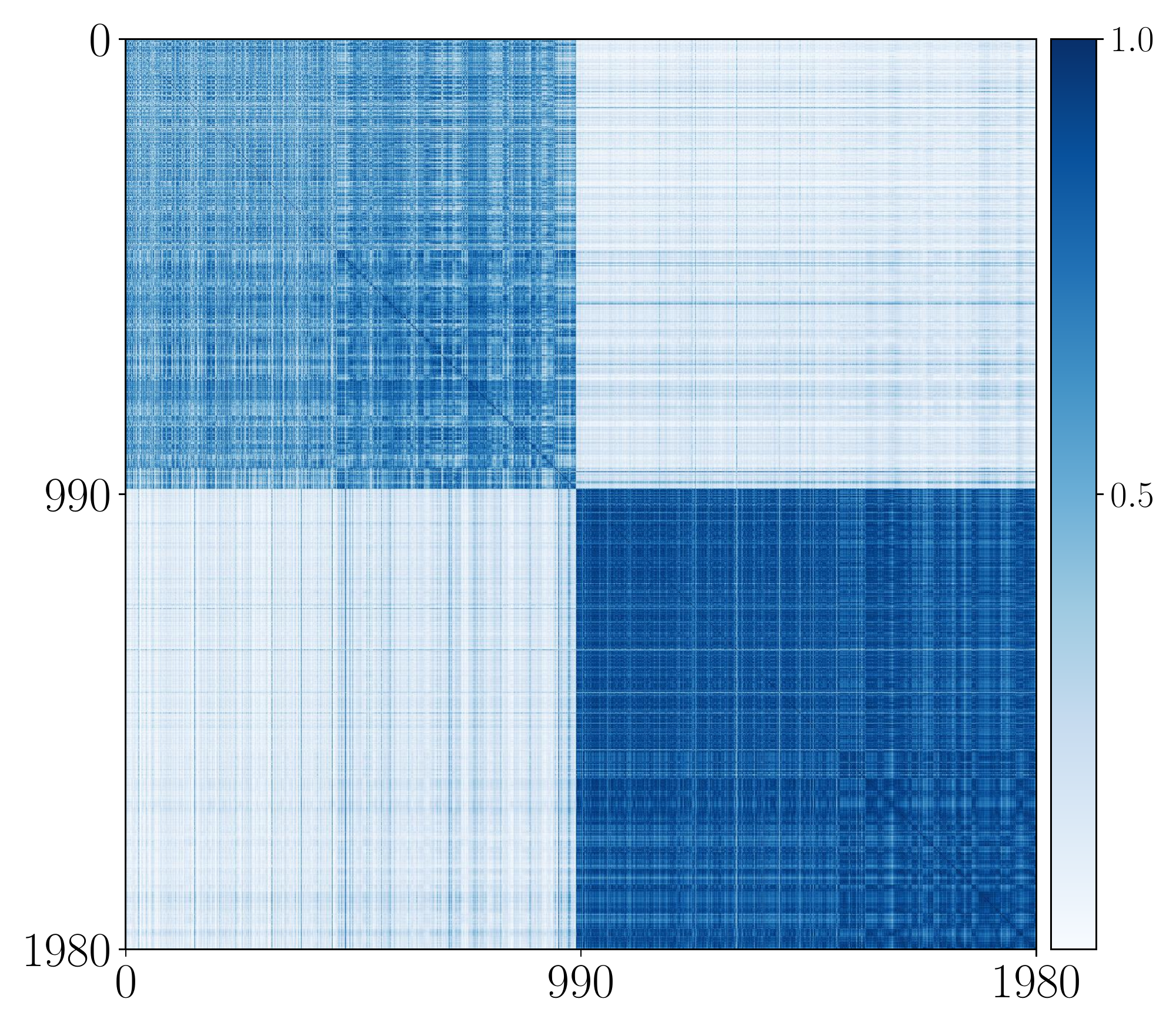}
        \caption{$\X_{\text{test}}$}
        \label{fig:mnist1d-heatmaps-c}
    \end{subfigure}
    \begin{subfigure}[b]{0.24\textwidth}
        \centering
        \includegraphics[width=\textwidth]{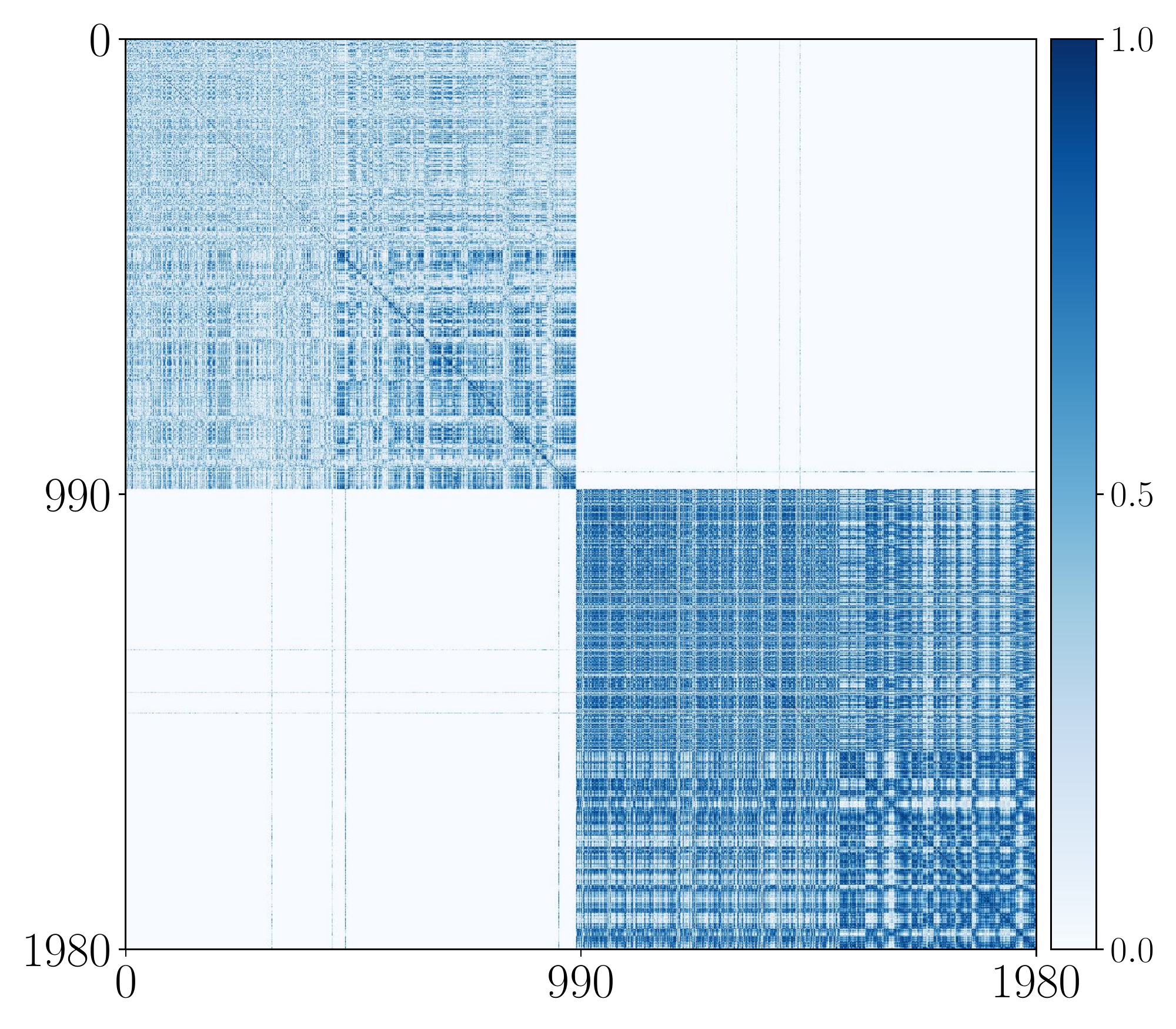}
        \caption{$\Z_{\text{test}}$}
        \label{fig:mnist1d-heatmaps-d}
    \end{subfigure}
    \caption{Cosine similarity (absolute value) of training/test data as well as traning/test representations for learning rotational invariant representations on MNIST.}
    \label{fig:appendix-mnist1d-heatmaps}
\end{figure}

\begin{figure}[ht]
    \centering
    \begin{subfigure}[b]{0.24\textwidth}
        \centering
        \includegraphics[width=\textwidth]{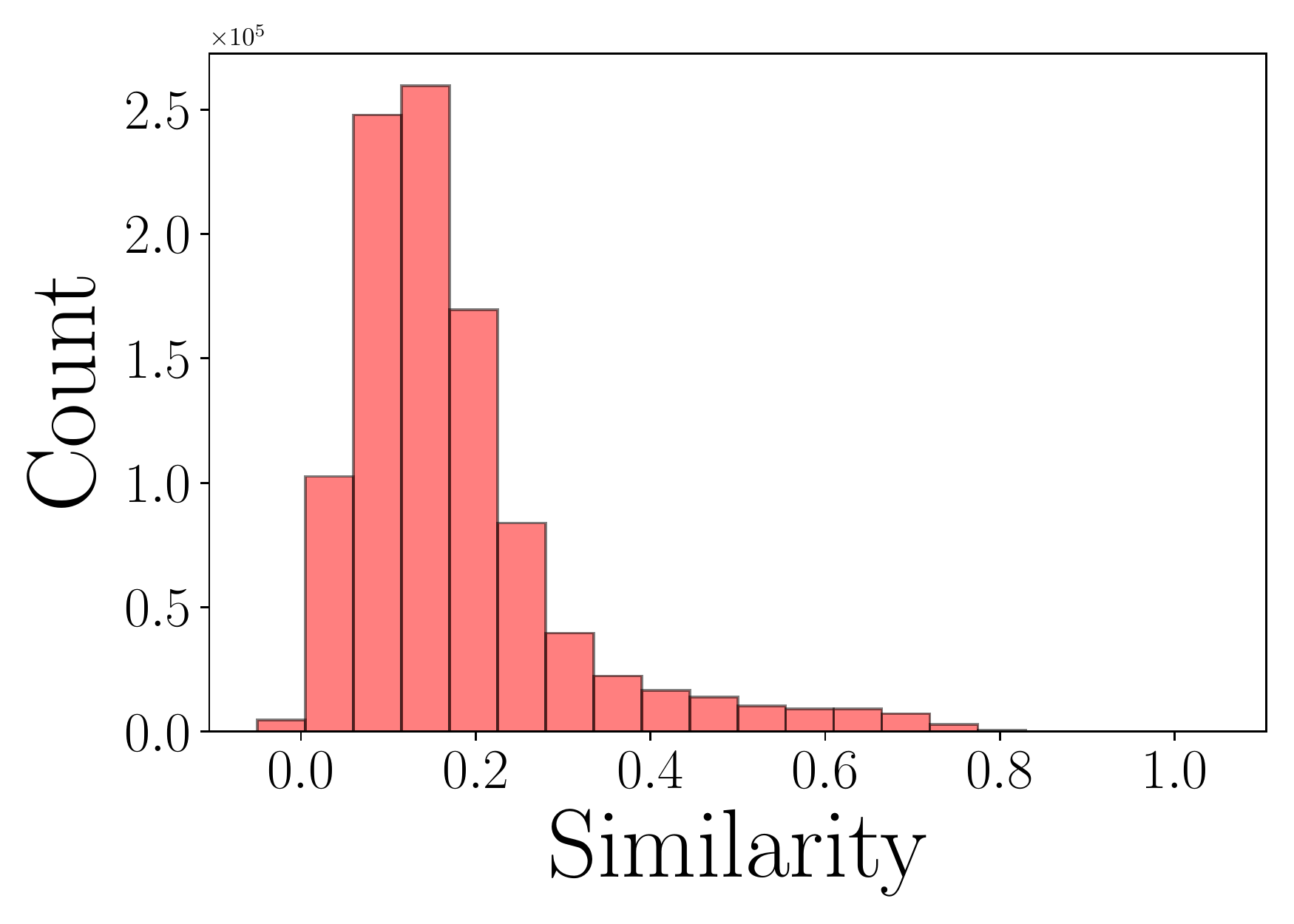}
        \caption{$\X_{1}${\scriptsize{(train)}} vs. $\X_{2}${\scriptsize{(test)}}}
    \end{subfigure}
    \begin{subfigure}[b]{0.24\textwidth}
        \centering
        \includegraphics[width=\textwidth]{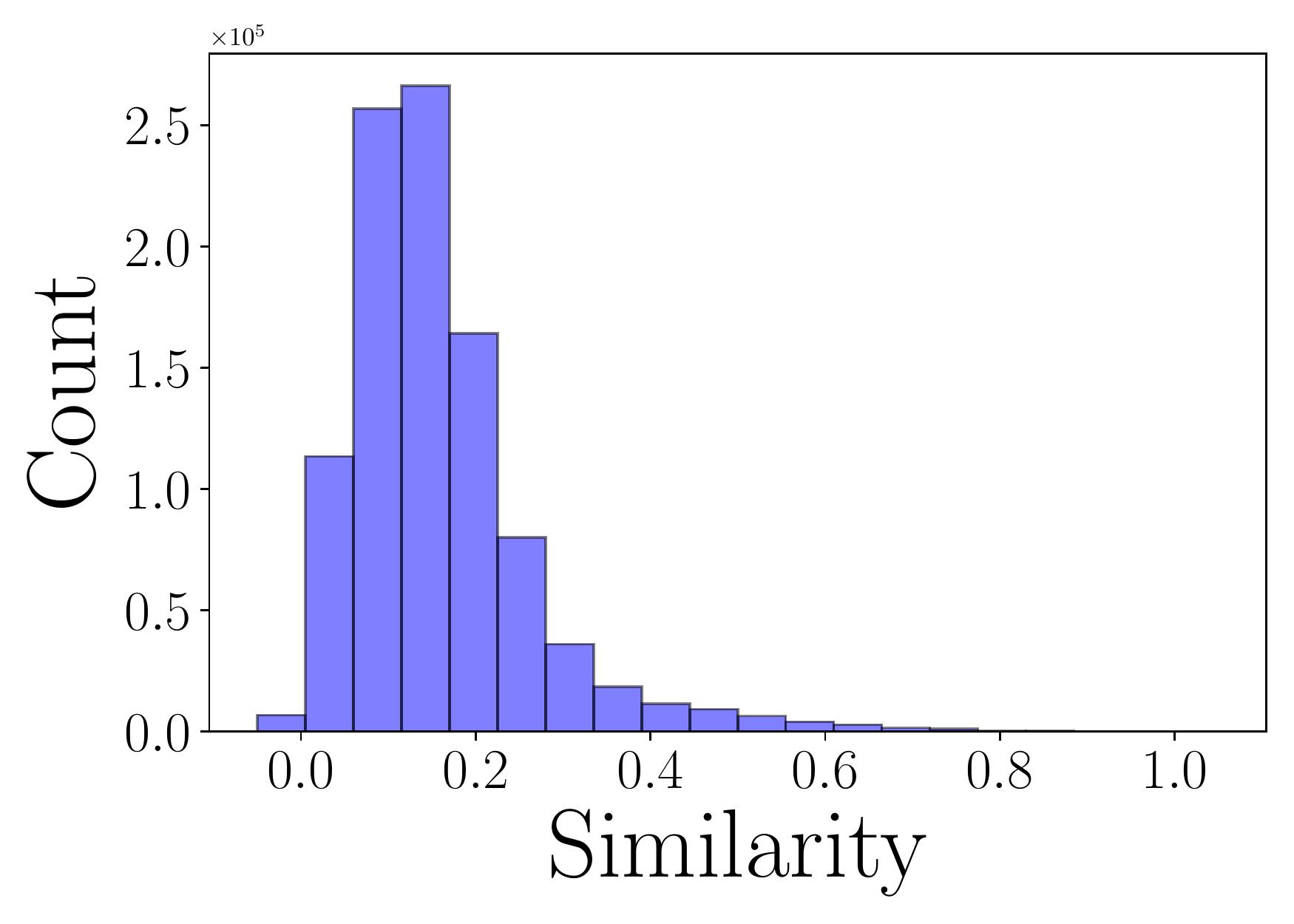}
        \caption{$\X_{2}${\scriptsize{(train)}} vs $\X_{1}${\scriptsize{(test)}}}
    \end{subfigure}
    \begin{subfigure}[b]{0.24\textwidth}
        \centering
        \includegraphics[width=\textwidth]{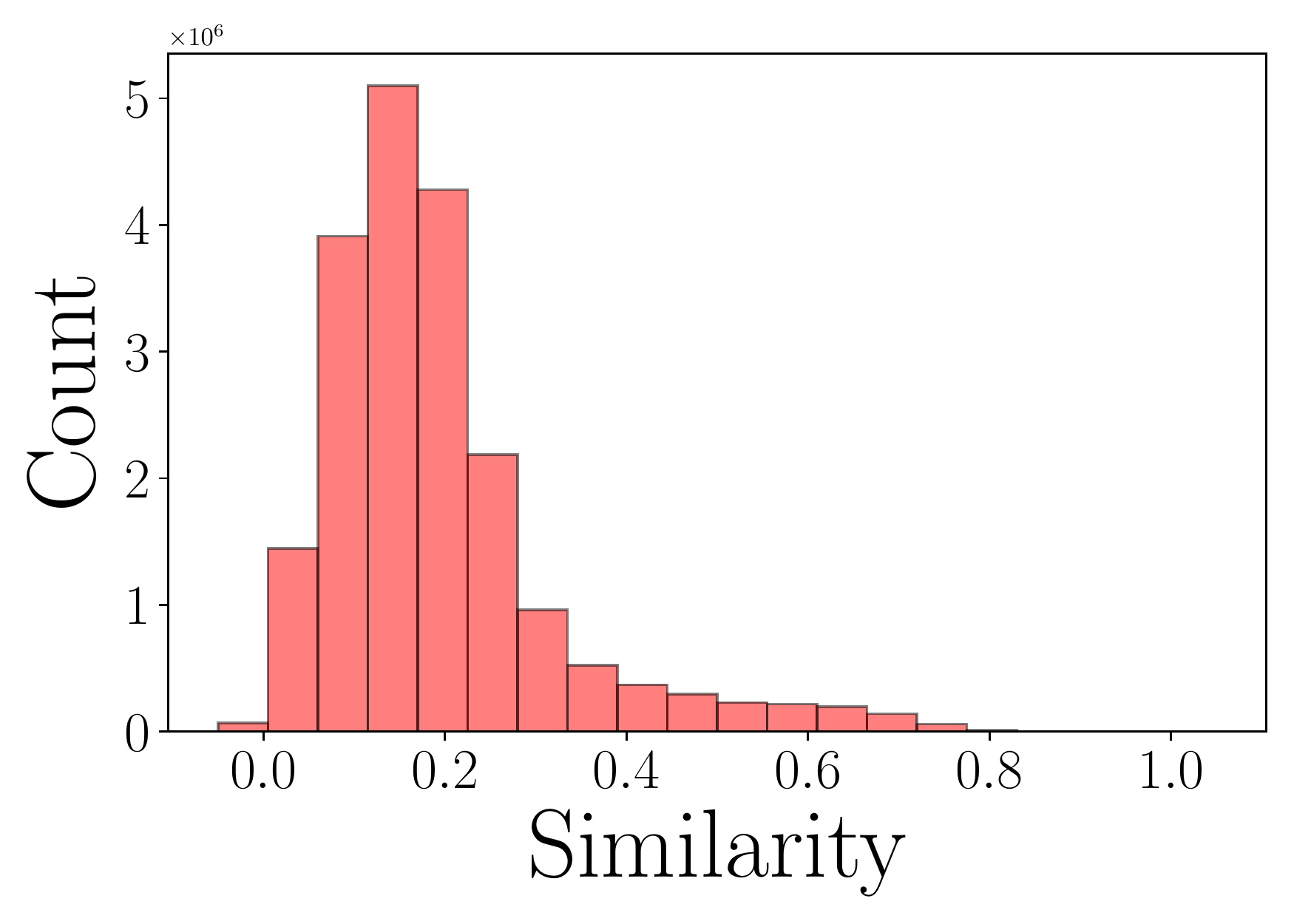}
        \caption{$\X_{1}${\scriptsize{(train)}} vs $\bar{\X}_{2}${\scriptsize{(test)}}}
    \end{subfigure}
    \begin{subfigure}[b]{0.24\textwidth}
        \centering
        \includegraphics[width=\textwidth]{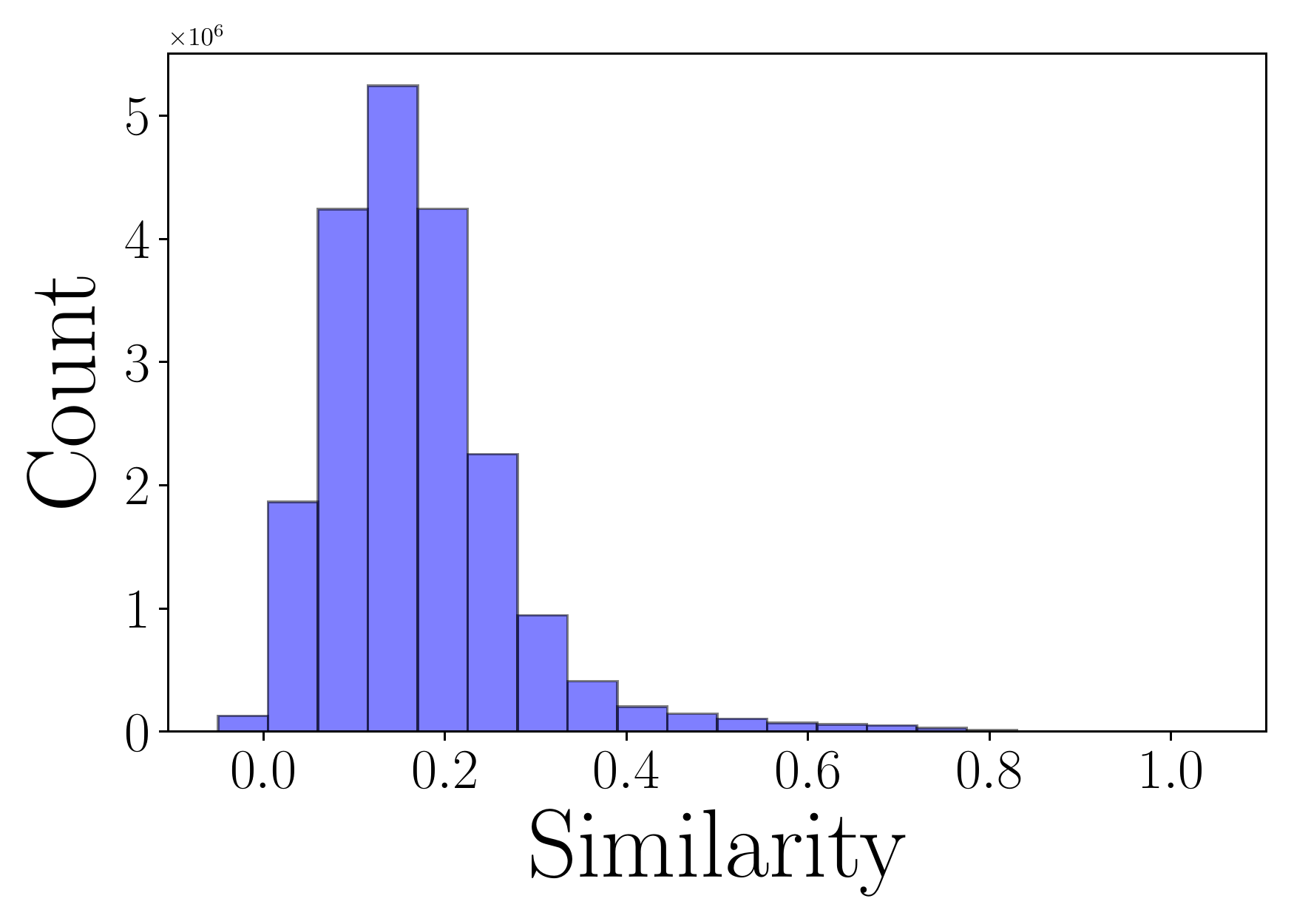}
        \caption{$\X_{2}${\scriptsize{(train)}} vs $\bar{\X}_{1}${\scriptsize{(test)}}}
    \end{subfigure}
    \begin{subfigure}[b]{0.24\textwidth}
        \centering
        \includegraphics[width=\textwidth]{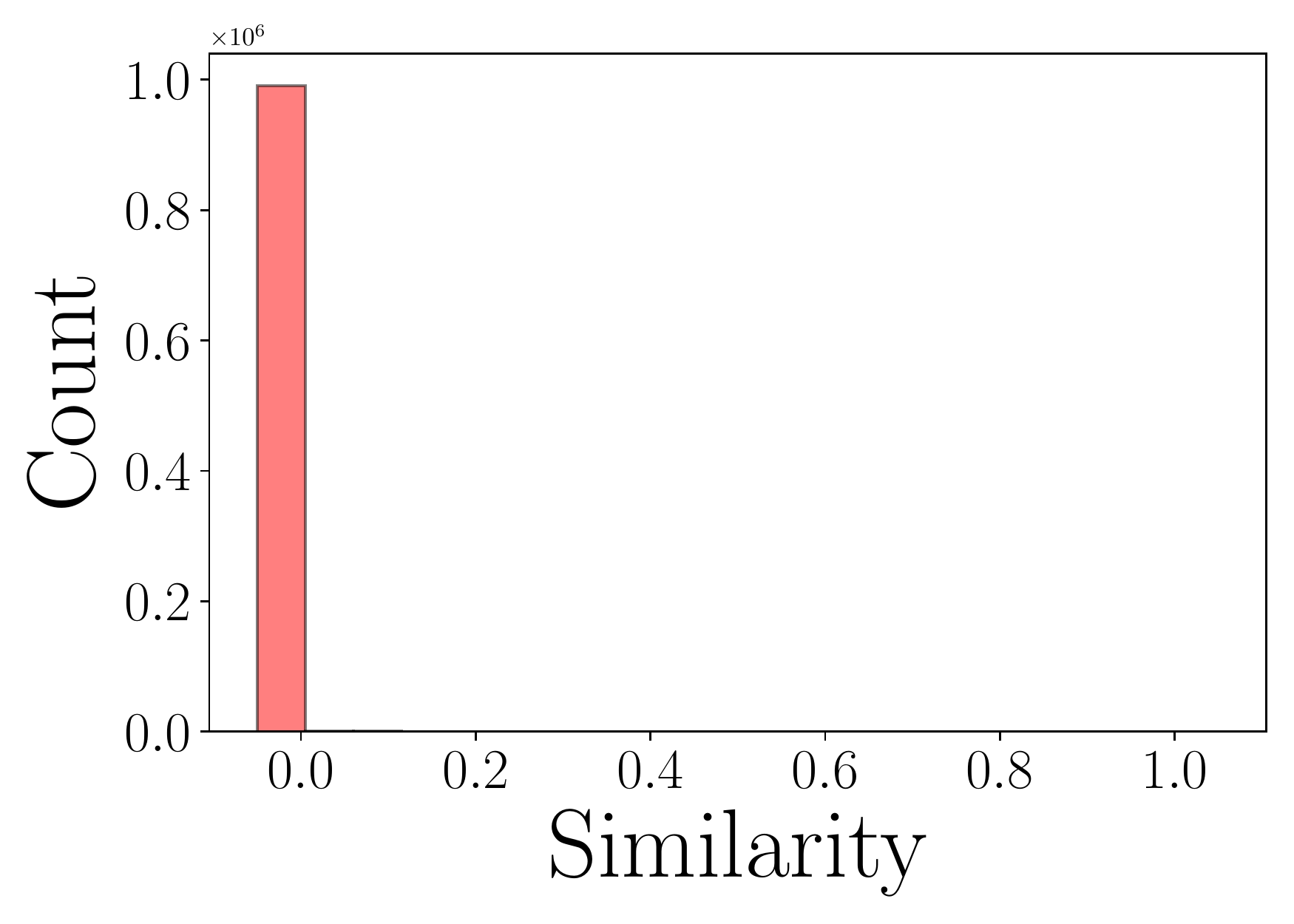}
        \caption{$\Z_{1}${\scriptsize{(train)}} vs. $\Z_{2}${\scriptsize{(test)}}}
    \end{subfigure}
    \begin{subfigure}[b]{0.24\textwidth}
        \centering
        \includegraphics[width=\textwidth]{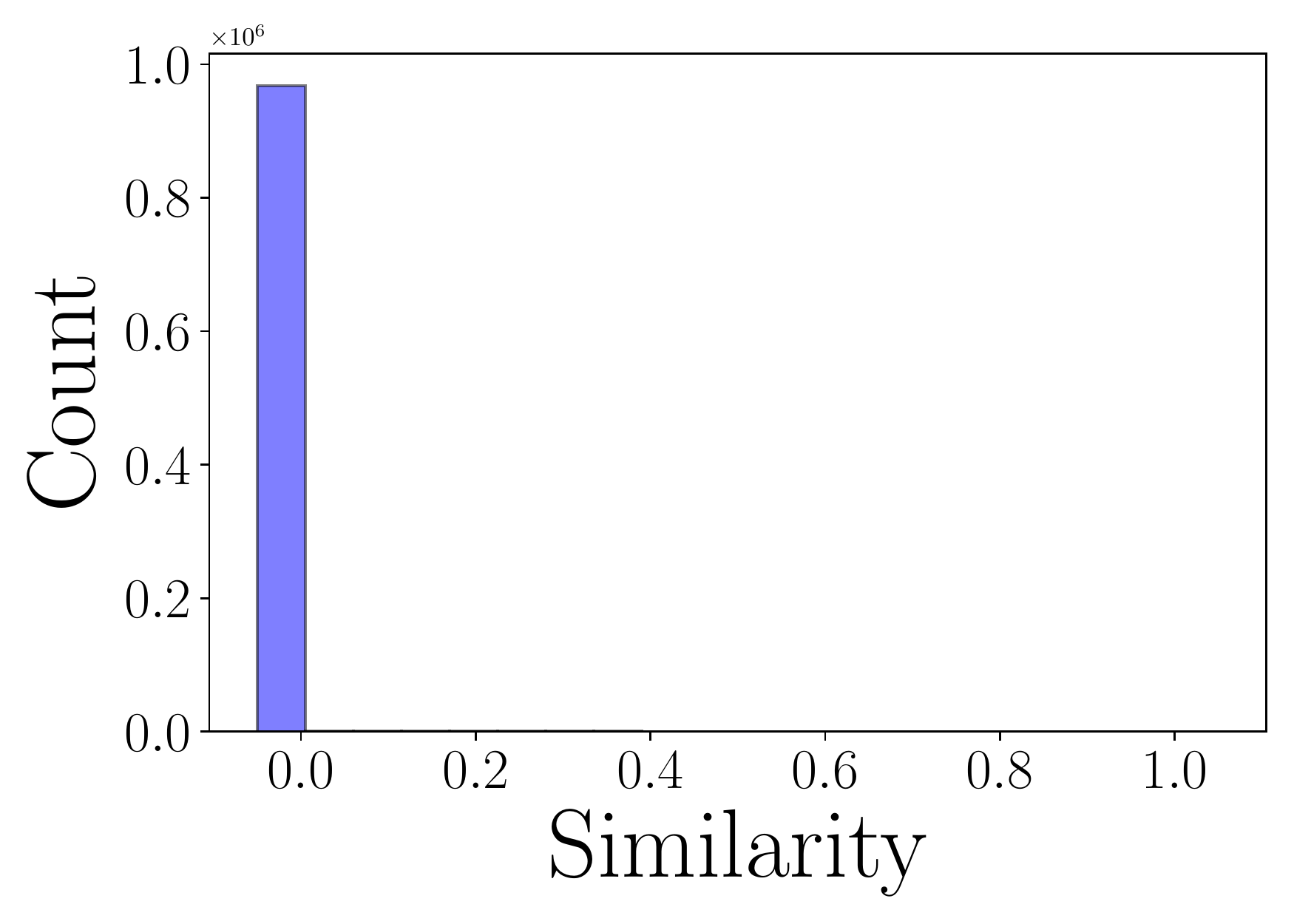}
        \caption{$\Z_{2}${\scriptsize{(train)}} vs $\Z_{1}${\scriptsize{(test)}}}
    \end{subfigure}
    \begin{subfigure}[b]{0.24\textwidth}
        \centering
        \includegraphics[width=\textwidth]{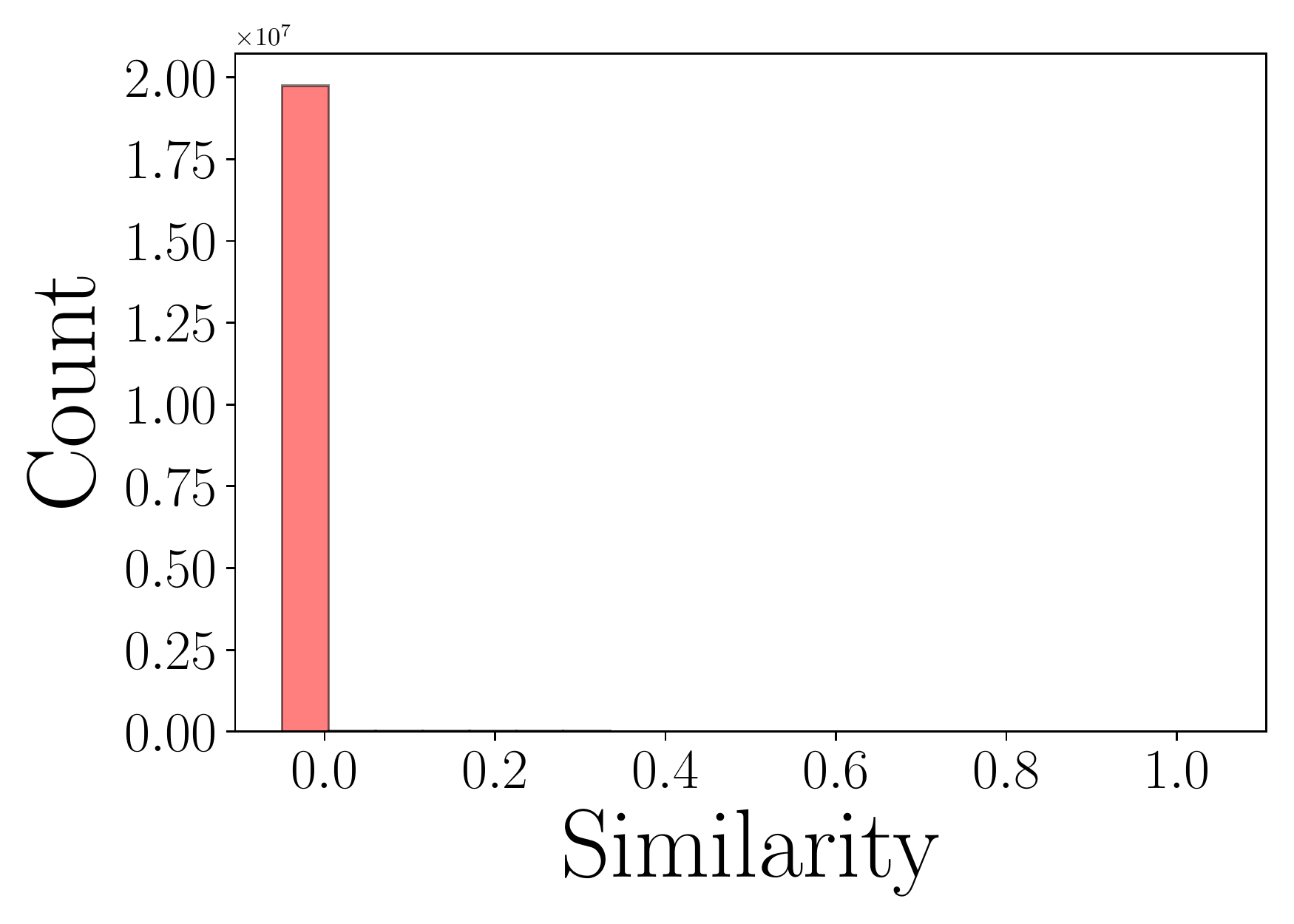}
        \caption{$\Z_{1}${\scriptsize{(train)}} vs $\bar{\Z}_{2}${\scriptsize{(test)}}}
    \end{subfigure}
    \begin{subfigure}[b]{0.24\textwidth}
        \centering
        \includegraphics[width=\textwidth]{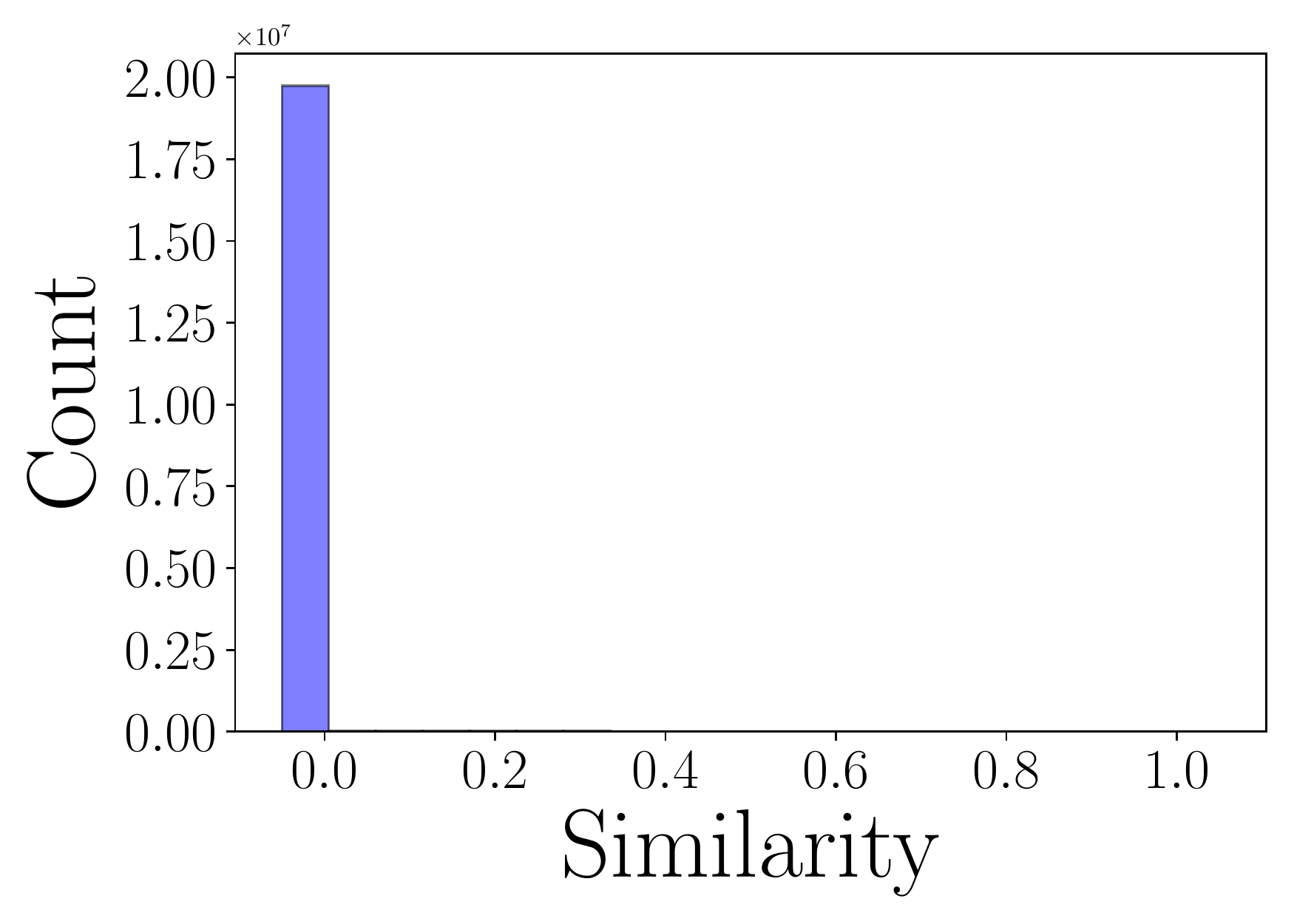}
        \caption{$\Z_{2}${\scriptsize{(train)}} vs $\bar{\Z}_{1}${\scriptsize{(test)}}}
    \end{subfigure}
    \caption{Histogram of cosine similarity between pairs sampled from different classes for learning rotational invariant representations on MNIST.  The histogram of cosine similarity between training data $\bm \X_{c}$ as well as representations $\bm {\Z}_{c}$ vs. testing (shifted) data $\bar{\X}_{c^{\prime}}$ as well as (shifted) representations $\bar{\Z}_{c^{\prime}}$, where we let $c$ denote the class index  and $c\neq c^{\prime}$.}
    \label{fig:appendix-cosine-hist-mnist}
\end{figure}

\begin{figure}
    \centering
    \begin{subfigure}[b]{0.49\textwidth}
        \includegraphics[width=\textwidth]{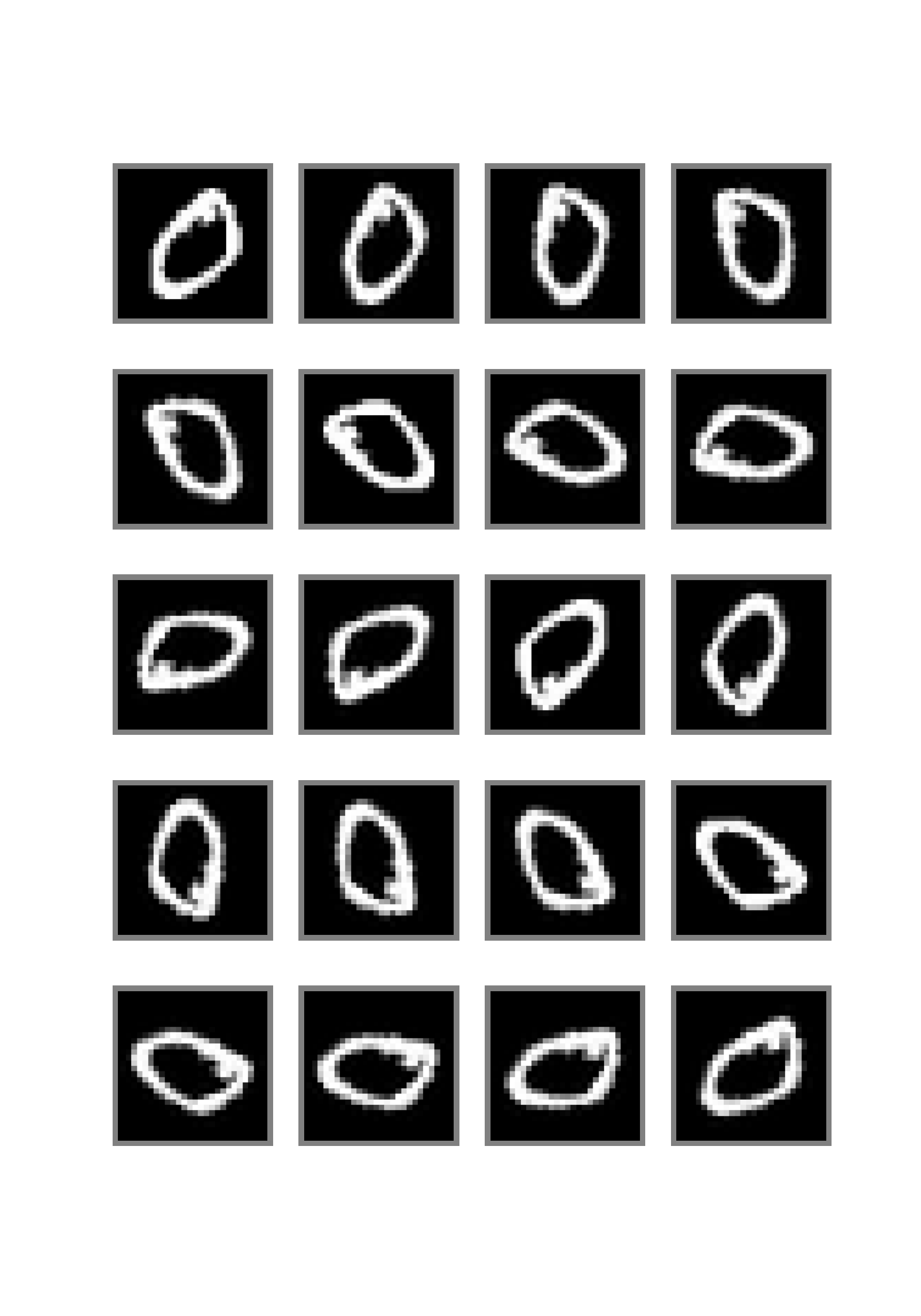}
    \end{subfigure}
    \begin{subfigure}[b]{0.49\textwidth}
        \includegraphics[width=\textwidth]{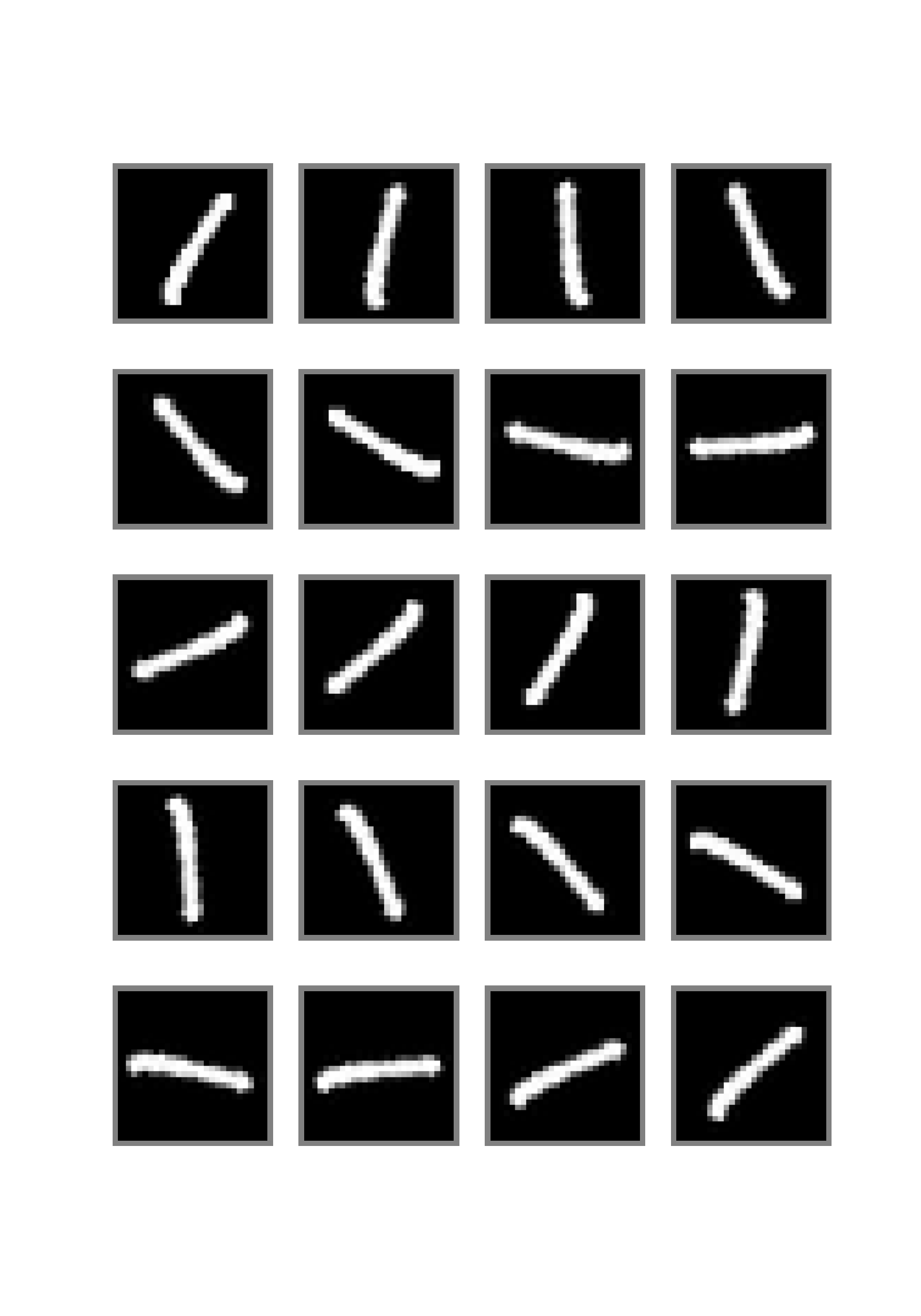}
    \end{subfigure}
    \caption{Examples of rotated images of MNIST digits for testing rotation invariance, each rotated by 18$^{\circ}$. (\textbf{Left}) digit `0'.  (\textbf{Right}) digit `1'. 
    }
    \label{fig:appendix-mnist-rotation-visualize}
\end{figure}

We provide additional experiments for \textit{learning rotational invariance on MNIST} in \textsection\ref{sec:experiments}. Examples of rotated images are shown in Figure~\ref{fig:appendix-mnist-rotation-visualize}.  We also provide cosine similarities between samples in Figure~\ref{fig:appendix-mnist1d-heatmaps}. We visualize the cosine similarities for the input $\X_{\text{train}}, \X_{\text{test}}$ as well as the learned representations $\Z_{\text{train}}, \Z_{\text{test}}$. The cosine similarity between sample pairs selected from different classes are shown in  Figure~\ref{fig:appendix-cosine-hist-mnist}. We can observe that the constructed ReduNet is able to learn discriminative (orthogonal) and invariant representations for MNIST digits.

